\documentclass[sigconf]{aamas} 

\usepackage{balance} 

\usepackage{hyperref}       
\usepackage{url}            
\usepackage{booktabs}       
\usepackage{amsfonts}       
\usepackage{nicefrac}       
\usepackage{microtype}      

\usepackage{amsmath}

\setcopyright{ifaamas}
\acmConference[AAMAS '21]{Proc.\@ of the 20th International Conference on Autonomous Agents and Multiagent Systems (AAMAS 2021)}{May 3--7, 2021}{Online}{U.~Endriss, A.~Now\'{e}, F.~Dignum, A.~Lomuscio (eds.)}
\copyrightyear{2021}
\acmYear{2021}
\acmDOI{}
\acmPrice{}
\acmISBN{}

\acmSubmissionID{417}

\usepackage{booktabs}
\usepackage{algorithm}
\usepackage[noend]{algpseudocode}
\usepackage{graphicx}
\usepackage{subcaption}

\title{State-Aware Variational Thompson Sampling for Deep Q-Networks}

\author{Siddharth Aravindan}
\affiliation{
  \institution{National University of Singapore}}
\email{siddharth.aravindan@comp.nus.edu.sg}

\author{Wee Sun Lee}
\affiliation{
  \institution{National University of Singapore}}
\email{leews@comp.nus.edu.sg}

\begin{abstract}
    
Thompson sampling is a well-known approach for balancing exploration and exploitation in reinforcement learning. It requires the posterior distribution of value-action functions to be maintained; this is generally intractable for tasks that have a high dimensional state-action space. We derive a variational Thompson sampling approximation for DQNs which uses a deep network whose parameters are perturbed by a learned variational noise distribution. We interpret the successful NoisyNets method \cite{fortunato2018noisy} as an approximation to the variational Thompson sampling method that we derive. Further, we propose State Aware Noisy Exploration (SANE) which seeks to improve on NoisyNets by allowing a non-uniform perturbation, where the amount of parameter perturbation is conditioned on the state of the agent. This is done with the help of an auxiliary perturbation module, whose output is state dependent and is learnt end to end with gradient descent. We hypothesize that such state-aware noisy exploration is particularly useful in problems where exploration in certain \textit{high risk} states may result in the agent failing badly. We demonstrate the effectiveness of the state-aware exploration method in the off-policy setting by augmenting DQNs with the auxiliary perturbation module.

\end{abstract}


\newcommand{\BibTeX}{\rm B\kern-.05em{\sc i\kern-.025em b}\kern-.08em\TeX}
\newcommand\norm[1]{\left\lVert#1\right\rVert}

\begin{document}


\pagestyle{fancy}
\fancyhead{}


\maketitle 


\section{Introduction}
Exploration is a vital ingredient in reinforcement learning algorithms that has largely contributed to its success in various applications \cite{khalil2017learning, mnih2015human, liang2017deep, gu2017deep}. Traditionally, deep reinforcement learning algorithms have used naive exploration strategies such as $\epsilon$-greedy, Boltzmann exploration or action-space noise injection to drive the agent towards unfamiliar situations. Although effective in simple tasks, such undirected exploration strategies do not perform well in tasks with high dimensional state-action spaces.

Theoretically, Bayesian approaches like Thompson sampling have been known to achieve an optimal  exploration-exploitation trade-off in multi-armed bandits \cite{agrawal2012analysis,agrawal2013further,kaufmann2012thompson} and also have been shown to provide near optimal regret bounds for Markov Decision Processes (MDPs) \cite{osband2017posterior,agrawal2017optimistic}. Practical usage of such methods, however, is generally intractable as they require the posterior distribution of value-action functions to be maintained. 

\begin{figure}
\centering
\begin{subfigure}[b]{0.4\textwidth}
\includegraphics[width=\textwidth]{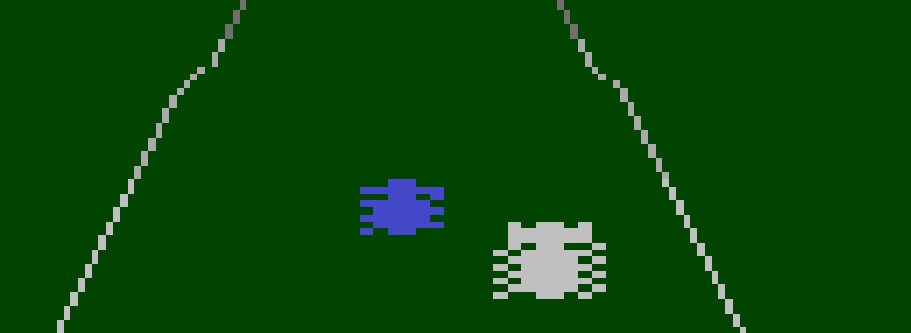}
\caption{A high risk state}
\end{subfigure}
\hspace{1.5 cm}
\begin{subfigure}[b]{0.4\textwidth}
\includegraphics[width=\textwidth]{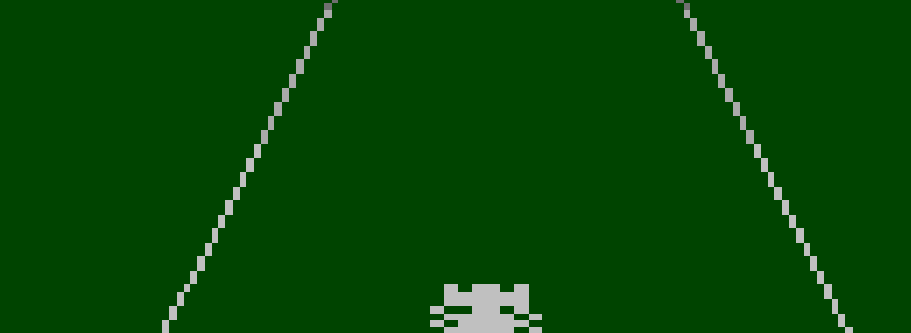}
\caption{A low risk state}
\end{subfigure}
\caption{The white car which is controlled by the agent, has to move forward while avoiding other cars. (a) In this state, any action other than moving straight will result in a crash, making it a high risk state. (b) This is a low risk state since exploring random actions will not lead to a crash.}
\label{fig:risk}
\end{figure} 
At the same time, in practical applications, perturbing the parameters of the model with Gaussian noise to induce exploratory behaviour has been shown to be more effective than $\epsilon$-greedy and other approaches that explore primarily by randomization of the action space  \cite{plappert2018parameter,fortunato2018noisy}. Furthermore, adding noise to model parameters is relatively easy and introduces minimal computational overhead. NoisyNets \cite{fortunato2018noisy}, in particular, has been known to achieve better scores on the full Atari suite than other exploration techniques\cite{Taiga2020On}. 

In this paper, we derive a variational Thompson sampling approximation for Deep Q-Networks (DQNs), where the model parameters are perturbed by a learned variational noise distribution. This enables us to interpret NoisyNets as an approximation of Thompson sampling, where minimizing the NoisyNet objective is equivalent to optimizing the variational objective with Gaussian prior and approximate posterior distributions. These Gaussian approximating distributions, however, apply perturbations uniformly across the agent's state space. We seek to improve this by approximating the Thompson sampling posterior with Gaussian distributions whose variance is dependent on the agent's state. 

To this end, we propose State Aware Noisy Exploration (SANE), an exploration strategy that induces exploration through state dependent parameter space perturbations. These perturbations are added with the help of an augmented state aware perturbation module, which is trained end-to-end along with the parameters of the main network by gradient descent. 

We hypothesize that adding such perturbations helps us mitigate the effects of high risk state exploration, while exploring effectively in low risk states. We define a high risk state as a state where a wrong action might result in adverse implications, resulting in an immediate failure or transition to states from which the agent is eventually bound to fail. Exploration in such states might result in trajectories similar to the ones experienced by the agent as a result of past failures, thus resulting in low information gain. Moreover, it may also prevent meaningful exploratory behaviour at subsequent states in the episode, that may have been possible had the agent taken the correct action at the state. A low risk state, on the other hand, is defined as a state where a random exploratory action does not have a significant impact on the outcome or the total reward accumulated by the agent within the same episode. A uniform perturbation scheme for the entire state space may thus be undesirable in cases where the agent might encounter high risk states and low risk states within the same task.  An instance of a high risk state and low risk state in Enduro, an Atari game, is shown in Figure \ref{fig:risk}. We try to induce uncertainty in actions, only in states where such uncertainty is needed through the addition of a state aware perturbation module.

To test our assumptions, we experimentally compare two SANE augmented Deep Q-Network (DQN) variants, namely the simple-SANE DQN and the Q-SANE DQN, with their NoisyNet counterparts \cite{fortunato2018noisy} on a suite of 11 Atari games. Eight of the games in the suite have been selected to have high risk and low risk states as illustrated in Figure \ref{fig:risk}, while the remaining three games do not exhibit such properties. We find that agents that incorporate SANE do better in most of the eight games.
An added advantage of SANE over NoisyNets is that it is more scalable to larger network models. The exploration mechanism in NoisyNets \cite{fortunato2018noisy} adds an extra learnable parameter for every weight to be perturbed by noise injection, thus tying the number of parameters in the exploration mechanism to the network architecture being perturbed. The noise-injection mechanism in SANE on the other hand, is a separate network module, independent of the architecture being perturbed. The architecture of this perturbation module can be modified to suit the task. This makes it more scalable to larger networks.
\section{Background}
\label{sec:background}
\subsection{Markov Decision Processes}
A popular approach towards solving sequential decision making tasks involves modelling them as MDPs. A MDP can be described as a 5-tuple, $(\mathcal{S}, \mathcal{A}, \mathcal{R}(.), \mathcal{T}(.),\gamma)$, where $\mathcal{S}$ and $\mathcal{A}$ denote the state space and action space of the task, $\mathcal{T}$ and $\mathcal{R}$ represent the state-transition and reward functions of the environment respectively and $\gamma$ is the discount factor of the MDP.
Solving a MDP entails learning an optimal policy that maximizes the expected cumulative discounted reward accrued during the course of an episode. Planning algorithms can be used to solve for optimal policies, when $\mathcal{T}$ and $\mathcal{R}$ are known. However, when these functions are unavailable, reinforcement learning methods help the agent \textit{learn} good policies.

\subsection{Deep Q-Networks}

A DQN \cite{mnih2015human} is a value based temporal difference learning algorithm, that estimates the action-value function by minimizing the temporal difference between two successive predictions. 
It uses a deep neural network as a function approximator to compute all action values of the optimal policy $Q^{\pi^*}(s,a)$, for a given a state $s$. A typical DQN comprises two separate networks, the Q network and the target network. The Q network aids the agent in interacting with the environment and collecting training samples to be added to the experience replay buffer, while the target network helps in calculating \textit{target} estimates of the action value function. The network parameters are learned by minimizing the loss $\mathcal{L}(\theta)$ given in Equation \ref{eq:dqn}, where $\theta$ and $\theta'$ are the parameters of the Q network and the target network respectively. The training instances $(s_i, a_i, r_i, s'_i)$ are sampled uniformly from the experience replay buffer, which contains the $k$ most recent transitions experienced by the agent. 

\begin{equation}
\mathcal{L}(\theta) = \mathbb{E} \big[\frac{1}{b}\sum\limits_{i=1}^{b}(Q(s_i,a_i; \theta) - (r_i + \gamma \max\limits_a Q(s'_i,a; \theta')))^2 \big]   
\label{eq:dqn}
\end{equation}

\subsection{Thompson Sampling}

 Thompson sampling \cite{thompson1933likelihood} works under the Bayesian framework to provide a well balanced exploration-exploitation trade-off. It begins with a prior distribution over the action-value and/or the environment and reward models. The posterior distribution over these models/values is updated based on the agent's interactions with the environment. A Thompson sampling agent tries to maximize its expected value by acting greedily with respect to a sample drawn from the posterior distribution.
 Thompson sampling has been known to achieve optimal and near optimal regret bounds in stochastic bandits \cite{agrawal2012analysis,agrawal2013further,kaufmann2012thompson} and MDPs \cite{osband2017posterior, agrawal2017optimistic} respectively. 
\section{Related Work}
\label{sec:rel_work}
Popularly used exploration strategies like $\epsilon$-greedy exploration, Boltzmann exploration and entropy regularization \cite{sutton2018reinforcement}, though effective, can be wasteful at times, as they do not consider the agent's uncertainty estimates about the state. In tabular settings, count based reinforcement learning algorithms such as UCRL \cite{jaksch2010near, auer2007logarithmic} handle this by maintaining state-action visit counts and incentivize agents with exploration bonuses to take actions that the agent is uncertain about. An alternative approach is suggested by posterior sampling algorithms like PSRL \cite{strens2000bayesian}, which maintain a posterior distribution over the possible environment models, and act optimally with respect to the model sampled from it. Both count based and posterior sampling algorithms have convergence guarantees in this setting and have been proven to achieve near optimal exploration-exploitation trade-off. Unfortunately, sampling from a posterior over environment models or maintaining visit counts  in most real world applications are computationally infeasible  due to the high dimensional state space and action space involved with these tasks. However, approximations of such methods that do well have been proposed in recent times. 

\citet{bellemare2016unifying}, generalizes count based techniques to non-tabular settings by using pseudo-counts obtained from a density model of the state space, while \cite{stadie2015incentivizing} follows a similar approach but uses a predictive model to derive the bonuses. \citet{ostrovski2017count} builds upon \cite{bellemare2016unifying} by improving upon the density models used for granting exploration bonuses. Additionally, surprise-based motivation \cite{achiam2017surprise} learns the transition dynamics of the task, and adds a reward bonus proportional to the Kullback–Leibler (KL) divergence between the true transition probabilities and the learned model to capture the agent's \textit{surprise} on experiencing a transition not conforming to its learned model. Such methods that add exploration bonuses  prove to be most effective in settings where the rewards are very sparse but are often complex to implement \cite{plappert2018parameter}.

Randomized least-squares value iteration (RLSVI) \cite{osband2016generalization} is an approximation of posterior sampling approaches to the function approximation regime. RLSVI draws samples from a distribution of linearly parameterized value functions, and acts according to the function sampled. \cite{osband2016deep} and \cite{osband2015bootstrapped} are similar in principle to \cite{osband2016generalization}; however, instead of explicitly maintaining a posterior distribution, samples are procured with the help of bootstrap re-sampling. Randomized Prior Functions \cite{osband2018randomized} adds \textit{untrainable prior networks} with the aim of capturing uncertainties not available from the agent's experience, while  \cite{azizzadenesheli2018efficient} tries to do away with duplicate networks by using Bayesian linear regression with Gaussian prior. Even though the action-value functions in these methods are no longer restricted to be linear, maintaining a bootstrap or computing a Bayesian linear regression makes these methods computationally expensive compared to others. 

Parameter perturbations which form another class of exploration techniques, have been known to enhance the exploration capabilities of agents in complex tasks ~\cite{xie2018nadpex, plappert2018parameter, florensa2017stochastic}. \citet{ruckstiess2010exploring} show that this type of policy perturbation in the parameter space outperforms action perturbation in policy gradient methods, where the policy is approximated with a linear function. However, \citet{ruckstiess2010exploring} evaluate this on tasks with low dimensional state spaces. 
When extended to high dimensional state spaces, black box parameter perturbations \cite{salimans2017evolution}, although proven effective, take a long time to learn good policies due to their non adaptive nature and inability to use gradient information. 
Gradient based methods that rely on adaptive scaling of the perturbations, drawn from spherical Gaussian distributions \cite{plappert2018parameter}, gradient based methods that learn the amount of noise to be added \cite{fortunato2018noisy} and gradient based methods that learn dropout policies for exploration \cite{xie2018nadpex} are known to be more sample efficient than black box techniques. NoisyNets \cite{fortunato2018noisy}, a method in this class, has been known to demonstrate consistent improvements over $\epsilon $-greedy across the Atari game suite unlike other count-based methods \cite{Taiga2020On}. Moreover, these methods are also often easier to implement and computationally less taxing than the other two classes of algorithms mentioned above.  

Our exploration strategy belongs to the class of methods that perturb parameters to effect exploration. Our method has commonalities with the parameter perturbing methods above as we sample perturbations from a spherical Gaussian distribution whose variance is learnt as a parameter of the network. However, the variance learnt, unlike NoisyNets \cite{fortunato2018noisy}, is conditioned on the current state of the agent. This enables it to sample perturbations from different Gaussian distributions to vary the amount of exploration when the states differ. Our networks also differ in the type of perturbations applied to the parameters. While \cite{fortunato2018noisy} obtains a noise sample from possibly different Gaussian distributions for each parameter, our network, like \cite{plappert2018parameter}, samples all perturbations from the same, but state aware, Gaussian distribution. Moreover, the noise injection mechanism in SANE is a separate network module that is subject to user design. This added flexibility might make it more scalable to larger network models when compared to NoisyNets, where this mechanism is tied to the network being perturbed.

\section{Variational Thompson Sampling}

\label{sec:ts}
Bayesian methods like Thompson Sampling use a posterior distribution $p(\theta | \mathcal{D})$ to sample the weights of the neural network, given  $\mathcal{D}$, the experience collected by the agent. $p(\theta | \mathcal{D})$ is generally intractable to compute and is usually approximated with a variational distribution $q(\theta)$. 
Let $D = (X,Y)$ be the dataset on which the agent is trained, with $X$ being the set of inputs, and $Y$ being the target labels.
Variational methods minimize the KL divergence between $q(\theta)$ and $p(\theta|D)$ to make $q(\theta)$ a better approximation. Appendix A shows that minimizing $KL(q(\theta), p(\theta|D))$ is equivalent to maximizing the Evidence Lower Bound (ELBO), given by Equation \ref{eq:KL1}.

\begin{align}
ELBO=\int q(\theta) \log p(Y|X,\theta)d\theta -KL( q(\theta), p(\theta))
\label{eq:KL1}
\end{align}

For a dataset with $N$ datapoints, and under the i.i.d assumption, we have : 
\begin{align*}
 \log p(Y|X,\theta) =\log \prod\limits_{i=1}^{N} p(y_i | x_i, \theta) = \sum\limits_{i=1}^{N}\log p(y_i | x_i, \theta)
\end{align*}

So, the objective to maximize is : 
\begin{align*}
& \max \int q(\theta) \sum\limits_{i=1}^{N}\log p(y_i | x_i, \theta)d\theta -KL( q(\theta), p(\theta)) \\ 
&= \max \left[    \left(\sum\limits_{i=1}^{N} \int q(\theta) \log p(y_i | x_i, \theta)d\theta \right) -KL( q(\theta), p(\theta))\right]
\end{align*}

In DQNs, the inputs $x_i$ are state-action tuples, and the its corresponding target $y_i$ is an estimate of $Q(s,a)$. Traditionally, DQNs are trained by minimizing the squared error, which assumes a Gaussian error distribution around the target value. Assuming the same, we define $\log p(y_i | x_i, \theta)$ in Equation \ref{eq:log_prob}, where $y_i$ is the approximate target Q value of $(s_t^i, a_t^i)$ given by $T_{i} = r^i_t + \max\limits_a \gamma Q(s^i_{t+1}, a; \theta')$, $\sigma^2_e$ is the variance of the error distribution  and $C(\sigma_e) = -\log \sqrt{(2\pi)}
\sigma_e$. 

\begin{align}
    \log p(y_i | x_i, \theta) = \frac{-(Q(s_t^i,a_t^i;\theta) - T_{i})^2}{2\sigma_e^2} + C(\sigma_e)  
    \label{eq:log_prob}
\end{align}

\begin{align}
    \nonumber ELBO =& \left(\sum\limits_{i=1}^{N} \int q(\theta) \left[\frac{-(Q(s^i_t,a^i_t;\theta) - T_{i})^2}{2\sigma_e^2} \right]d\theta \right) \\
    \nonumber & -KL( q(\theta), p(\theta)) +  NC(\sigma_e) \\
    \nonumber =& C_1 \left(\sum\limits_{i=1}^{N} \int q(\theta) \left[-(Q(s^i_t,a^i_t;\theta) - T_{i})^2 \right]d\theta \right) \\
    & -KL( q(\theta), p(\theta)) +  NC(\sigma_e)
    \label{eq:elbo_for_sane}
\end{align}

We approximate the integral for each example with a Monte Carlo estimate by sampling a $\hat{\theta_i} \sim q(\theta)$, giving 
\begin{align*}
     ELBO \approx& C_1 \left(\sum\limits_{i=1}^{N}-(Q(s^i_t,a^i_t;\hat{\theta_i}) - T_{i})^2 \right) 
     -KL( q(\theta), p(\theta)) +  NC(\sigma_e)
\end{align*}

 As $C(\sigma_e)$ is a constant with respect to $\theta$, maximizing the ELBO is approximately the same as optimizing the following objective.

\begin{align}
     \max \left[C_1 \left(\sum\limits_{i=1}^{N}-(Q(s^i_t,a^i_t;\hat{\theta_i}) - T_{i})^2 \right) 
     -KL( q(\theta), p(\theta))\right]
    \label{eq:final_elbo}
\end{align}

\subsection{Variational View of NoisyNet DQNs}
The network architecture of NoisyNet DQNs usually comprises a series of convolutional layers followed by some fully connected layers. The parameters of the convolutional layers are not perturbed, while every parameter of the fully connected layers is perturbed by a separate Gaussian noise whose variance is learned along with the other parameters of the network. 

For the unperturbed parameters of the convolutional layers, we consider $q(\theta_{c}) = \mathcal{N}(\mu_{c}, \epsilon I)$. The parameters of any neural network are usually used in the floating point format. We choose a value of $\epsilon$  that is close enough to zero, such that adding any noise sampled from these distributions does not change the value of the weight as represented in this format with high probability. For the parameters of the fully connected layers,  we take $q(\theta_{fc}) = \mathcal{N}(\mu_{fc}, \Sigma_{fc})$  where $\Sigma$  is a diagonal matrix with $\Sigma_{fc}^{ii}$ equal to the learned variance for the parameter $\theta_i$. We take the prior $p(\theta) = \mathcal{N}(0, I)$ for all the parameters of the network.

With this choice of $p(\theta)$ and $q(\theta)$, the value of $KL(q(\theta), p(\theta))$ can be computed as shown in Equation \ref{eq:final_kl}, where $k_1$ and $k_2$ are the number of parameters in the the convolutional and fully connected layers respectively. Note that $k_1$, $k_2$ and $\epsilon$ are constants given the network architecture. 

 \begin{align}
    \nonumber KL(q(\theta), p(\theta)) =&  \frac{1}{2}\left[-k_1 \log(\epsilon)  + k_1\epsilon + \norm{\mu_{c}}_2^2 -k_1 \right]  \\
    +&\frac{1}{2} \left[ -\log |\Sigma_{fc}| + tr(\Sigma_{fc}) + \norm{\mu_{fc}}_2^2  -k_2\right]
    \label{eq:final_kl}
\end{align}

As NoisyNet DQN agents are usually trained on several million interactions, we assume that the  KL term is dominated by the log likelihood term in the ELBO. Thus, maximizing the objective in Equation (\ref{eq:final_elbo}) can be approximated by optimizing the  following objective :  
\begin{align}
     \max \left(\sum\limits_{i=1}^{N}-(Q(s^i_t,a^i_t;\hat{\theta_i}) - T_{i})^2  \right)
    \label{eq:final_elbo_nn}
\end{align}
which is the objective that NoisyNet DQN agents optimize. In NoisyNets, every sample $\hat{\theta_i} \sim \mathcal{N}(\mu, \Sigma)$ is obtained by a simple reparameterization of the network parameters : $\hat{\theta_i} = \mu + \Sigma \epsilon$, where $\epsilon \sim \mathcal{N}(0,I)$. This reparameterization helps NoisyNet DQNs to learn through a sampled $\hat{\theta_i}$.

\subsection{State Aware Approximating Distributions}
It can be seen that the approximate posterior distribution $q(\theta)$ is state agnostic, i.e., it applies perturbations uniformly across the state space, irrespective of whether the state is \textit{high risk} or \textit{low risk}. We thus postulate that $q(\theta|s)$ is potentially a better variational approximator . $q(\theta)$ is  a special case of a state aware variational approximator  where $q(\theta|s)$ is the same for all $s$.
A reasonable ELBO estimate for such an approximate distribution  would be to extend the ELBO in Equation \ref{eq:elbo_for_sane} to accommodate $q(\theta|s)$ as shown in \ref{eq:elbo_sane}. 
\begin{align}
\nonumber ELBO =&C_1 \left(\sum\limits_{i=1}^{N} \int q(\theta|s^i_t) \left[-(Q(s^i_t,a^i_t;\theta) - T_{i})^2 \right]d\theta \right) \\
    & - \frac{1}{N} \sum\limits_{i=1}^N KL( q(\theta|s^i_t), p(\theta)) +  NC(\sigma_e)
\label{eq:elbo_sane}
\end{align}
Approximating the integral for each example with a Monte Carlo estimate by sampling a $\hat{\theta_i} \sim q(\theta|s^i_t)$, maximizing the ELBO is equivalent to solving  \ref{eq:final_elbo_sane}.
\begin{align}
     \max \left[\sum\limits_{i=1}^{N} \left(-C_1(Q(s^i_t,a^i_t;\hat{\theta_i}) - T_{i})^2  
     -\frac{1}{N}KL( q(\theta|s^i_t), p(\theta))\right)\right]
    \label{eq:final_elbo_sane}
\end{align}
We assume that the KL term will eventually be dominated by the log likelihood term in the ELBO, given a sufficiently large dataset. 
This posterior approximation leads us to the formulation of SANE DQNs as described in the following sections.
\section{State Aware Noisy Exploration}
\label{sec:sae}
\begin{figure*}
\centering
\includegraphics[width=0.98\textwidth]{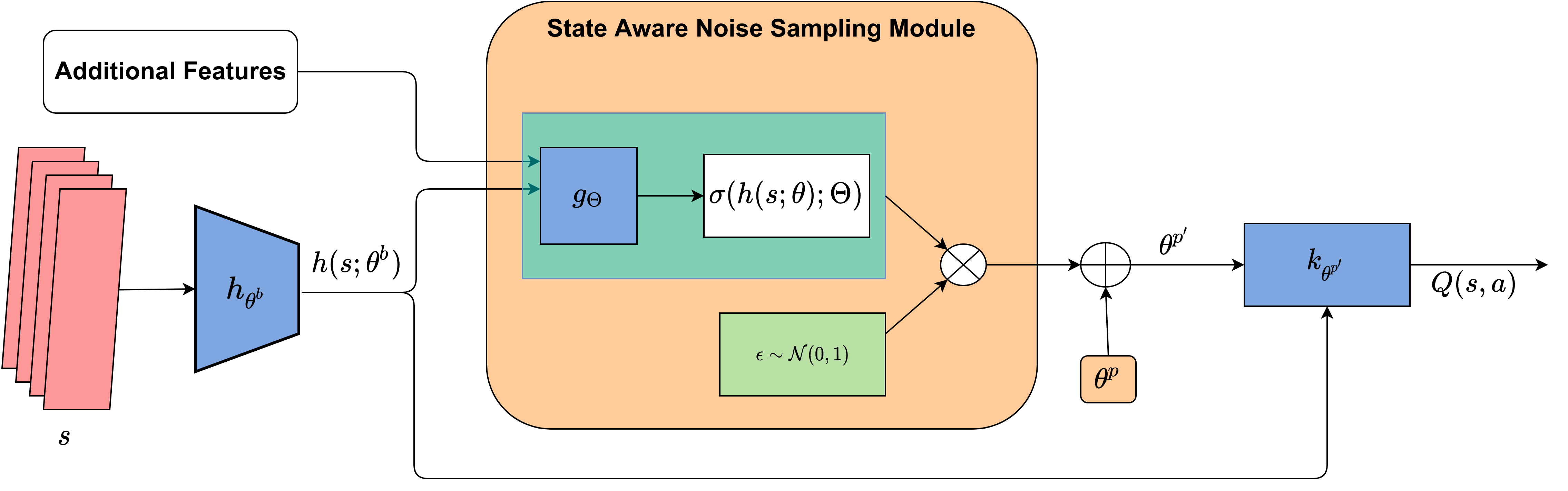}
\caption{A high level view of a State Aware Noisy Exploring Network.}
\label{fig:SAE_module}
\end{figure*} 
State Aware Noisy Exploration (SANE), is a parameter perturbation based exploration strategy which induces exploratory behaviour in the agent by adding noise to the parameters of the network. The noise samples are drawn from the Gaussian distribution  $\mathcal{N}(0,\sigma^2(h(s; \theta); \Theta))$, where $\sigma(h(s; \theta); \Theta)$ is computed as a function of a hidden representation, $h(s; \theta)$, of the state $s$ of the agent by an auxiliary neural network module, i.e., $\sigma(h(s; \theta); \Theta) = g_{\Theta}(h(s; \theta))$, where  $\Theta$ and $\theta$ refer to the parameters of the auxiliary perturbation network ($g$) and the parameters of the main network respectively. 

\subsection{State Aware Noise Sampling}
To procure state aware noise samples, we first need to compute $\sigma(h(s;\theta); \Theta)$, the state dependent standard deviation of the Normal distribution from which the perturbations are sampled. As stated above, we do this by adding an auxiliary neural network module.  $\sigma(h(s; \theta); \Theta)$ is then used to generate perturbations $\epsilon \sim \mathcal{N}(0,\sigma^2(h(s; \theta); \Theta))$ for every network parameter using noise samples from the standard Normal distribution, $\hat{\epsilon} \sim \mathcal{N}(0,1)$, in tandem with a simple reparameterization of the sampling network \cite{salimans2017evolution, plappert2018parameter, fortunato2018noisy} as shown in Equation \ref{eq:reparameterization}.
\begin{equation}
    \epsilon = \sigma(h(s;\theta);\Theta) \hat{\epsilon}; \;\; \hat{\epsilon} \sim \mathcal{N}(0,1)
    \label{eq:reparameterization}
\end{equation}
State aware perturbations can be added to all types of layers in the network. The standard baseline architectures used by popular deep reinforcement learning algorithms for tasks such as playing Atari games mainly consist of several convolutional layers followed by multiple fully connected layers. We pass the output of the last convolutional layer as the hidden representation $h(s;\theta)$ to compute the state aware standard deviation, $\sigma(h(s; \theta); \Theta)$, where $\theta$ is the set of parameters of the convolutional layers. Perturbations using $\sigma(h(s; \theta); \Theta)$  are then applied to the following fully connected layers. 

Our mechanism of introducing perturbations is similar to Noisy DQNs \cite{fortunato2018noisy} and adaptive parameter space noise \cite{plappert2018parameter}. Given a vector $x \in \mathbb{R}^l$ as input to a fully connected layer with $m$ outputs, an unperturbed layer computes a matrix transformation of the form $y=Wx+b$, where $W$ and $b$ are the parameters associated with the layer, and $y \in \mathbb{R}^m$. We modify such layers with state-aware perturbations, by adding noise elements sampled from $\mathcal{N}(0,\sigma^2(h(s; \theta); \Theta))$ (Equation \ref{eq:reparameterization}). This results in the perturbed fully connected layer computing a transform equivalent to Equation \ref{eq:state_aware_layer}, where $\widetilde{W} = W+\sigma(h(s; \theta); \Theta)\epsilon_w$, $\widetilde{b} = b+\sigma(h(s; \theta); \Theta)\epsilon_b$, and $\epsilon_w \in \mathbb{R}^{m \times l}$, $\epsilon_b \in \mathbb{R}^m$ are vectors whose elements are samples from a standard normal distribution. 
\begin{equation}
    y = \widetilde{W}x + \widetilde{b}
    \label{eq:state_aware_layer}
\end{equation}
A high level view of a neural network with the augmented state aware perturbation module is shown in Figure \ref{fig:SAE_module}. We partition $\theta$ into $\theta^b$ and $\theta^p$, where $\theta^b$ is the set of parameters used to generate the hidden state representation $h(s;\theta^b)$ using the neural network $h$ and $\theta^p$ are the parameters to which noise is to be added. Given the hidden state representation, perturbation module $g_\Theta$, is used to compute the state dependent standard deviation $\sigma(h(s; \theta^b); \Theta)$, which is used to perturb the parameters $\theta^p$ of the network $k$. $k$ then computes action-values for all actions. Additional features that may aid in exploration such as state visit counts or uncertainty estimates can also be appended to  $h(s; \theta^b)$ before being passed as input to $g_\Theta$.

\citet{fortunato2018noisy} suggests two alternatives to generate $\epsilon_w$ and $\epsilon_b$. The more computationally expensive alternative, Independent Gaussian noise, requires the sampling of each element of $\epsilon_w$ and $\epsilon_b$ independently, resulting in a sampling of $lm + m$ quantities per layer. Factored Gaussian noise, on the other hand, samples two vectors $\hat{\epsilon_l}$ and $\hat{\epsilon_m}$ of sizes $l$ and $m$ respectively. These vectors are then put through a real valued function $z(x) = sgn(x) \sqrt{x}$ before an outer product is taken to generate $\epsilon_w$ and $\epsilon_b$ (Equation \ref{eq:apply_function}), which are the required noise samples for the layer. Readers are referred to \cite{fortunato2018noisy} for more details on these two noise sampling techniques. Being less computationally taxing and not having any notable impact on the performance \cite{fortunato2018noisy}, we select Factored Gaussian noise as our method for sampling perturbations.  
\begin{equation}
    \epsilon_w = z(\hat{\epsilon_l}) \otimes z(\hat{\epsilon_m}), \;\; \epsilon_b = z(\hat{\epsilon_m}) 
    \label{eq:apply_function}
\end{equation}
\subsection{Network Parameters and Loss Function}
The set of learnable parameters for a SANE network, is a union of the set of parameters of the main network, $\theta$, and the set of parameters of the auxiliary network perturbation module, $\Theta$. Moreover, in place of minimizing the loss over the original set of parameters, $\mathbb{E}\left[\mathcal{L}(\theta) \right]$, the SANE network minimizes the function $\mathbb{E}\left[\mathcal{L}([\theta, \Theta])\right]$, which is the loss corresponding to the network parameterized by the perturbed weights of the network. Furthermore, with both the main network and the perturbation module being differentiable entities, using the reparameterization trick to sample the perturbations allows the joint optimization of both $\theta$ and $\Theta$ via backpropagation.

\subsection{State Aware Deep Q Learning}
We follow an approach similar to \cite{fortunato2018noisy} to add state aware noisy exploration to DQNs \cite{mnih2015human}. In our implementation of a SANE DQN for Atari games, $\theta^b$ and $\theta^p$ correspond to the set of parameters in the convolutional layers and the fully connected layers respectively.  The Q network and the target network have their own copies of the network and perturbation module parameters. 

The DQN learns by minimizing the following loss, where $\theta, \theta'$ represent the network parameters of the Q-network and the target network and $\Theta, \Theta'$ represent the perturbation module parameters of the Q-network and the target network respectively. The training samples $(s_i, a_i, r_i, s'_i)$ are drawn uniformly from the replay buffer.
\begin{align*}
    \mathcal{L}(\theta, \Theta) =\mathbb{E} \big[\frac{1}{b}\sum\limits_{i=1}^{b}(Q(s_i,a_i; \theta, \Theta) - (r_i + \gamma \max\limits_a Q(s'_i,a; \theta', \Theta')))^2 \big]  
\end{align*}
\begin{figure*}[t!]
\begin{subfigure}[b]{0.11\textwidth}
\includegraphics[width= \linewidth,height=3cm]{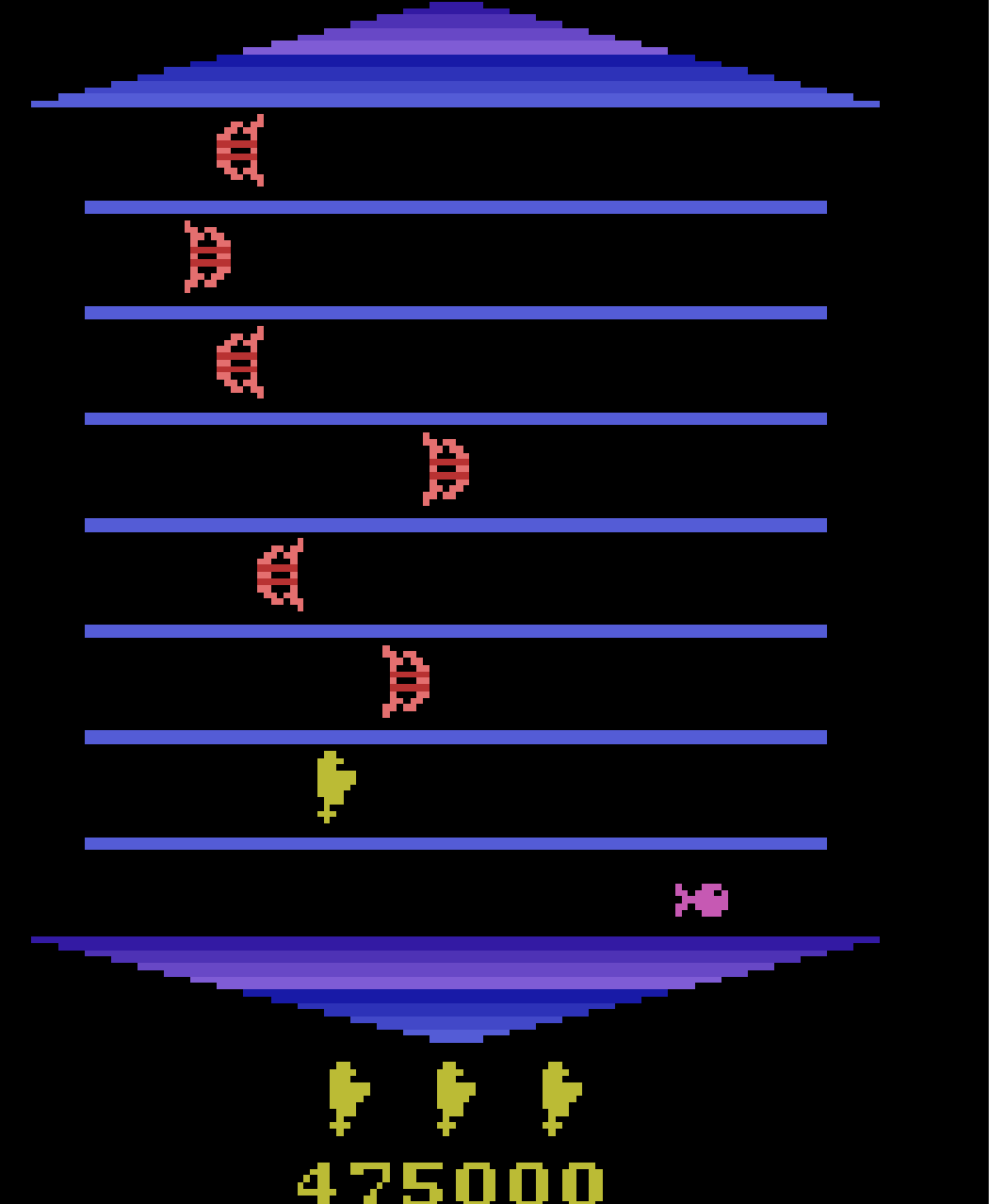}
\caption{}
\label{fig:asterix_h}
\end{subfigure}
~
\begin{subfigure}[b]{0.11\textwidth}
\includegraphics[width= \linewidth,height=3cm]{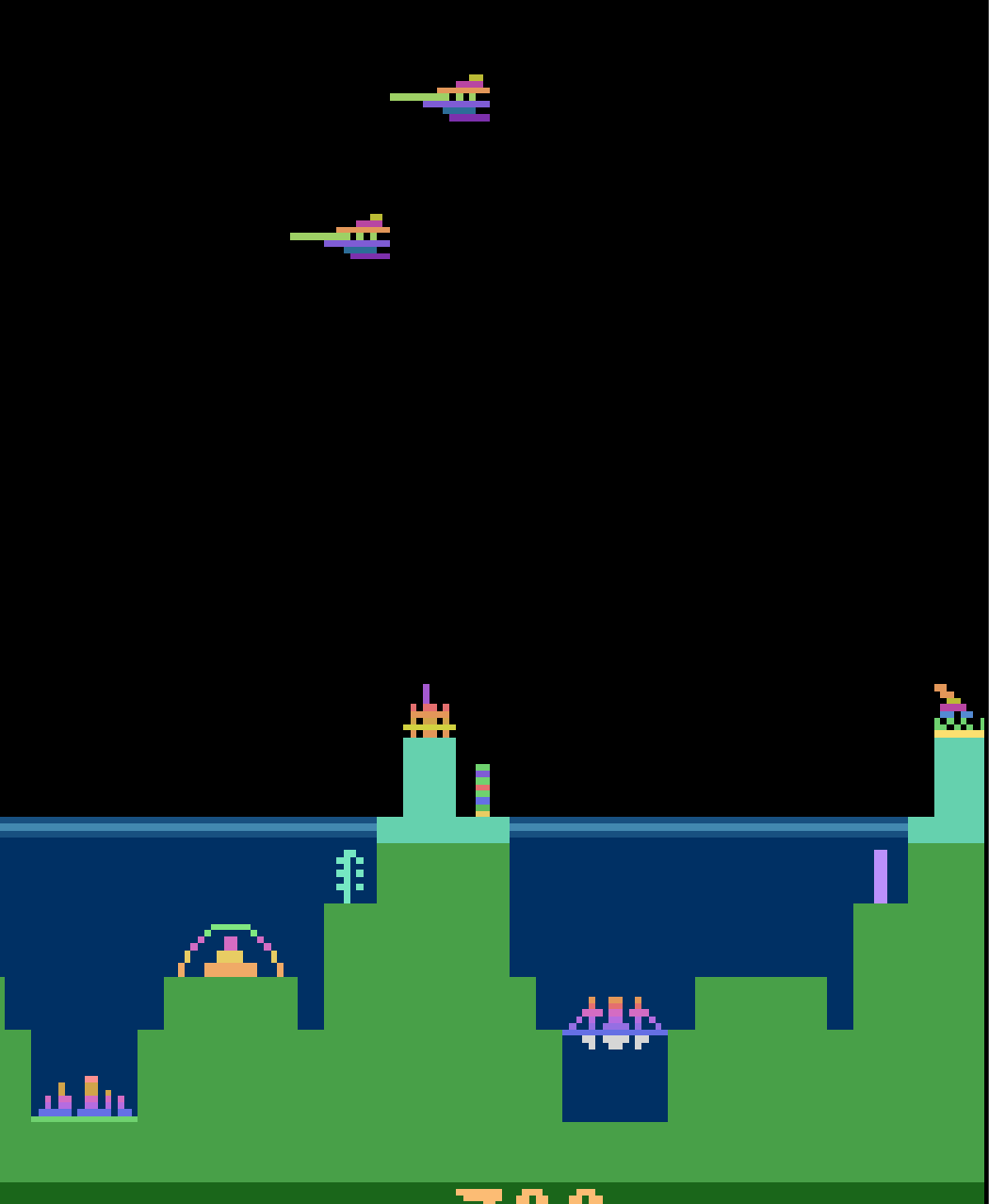}
\caption{}
\label{fig:atlantis_h}
\end{subfigure}
~
\begin{subfigure}[b]{0.11\textwidth}
\includegraphics[width= \linewidth,height=3cm]{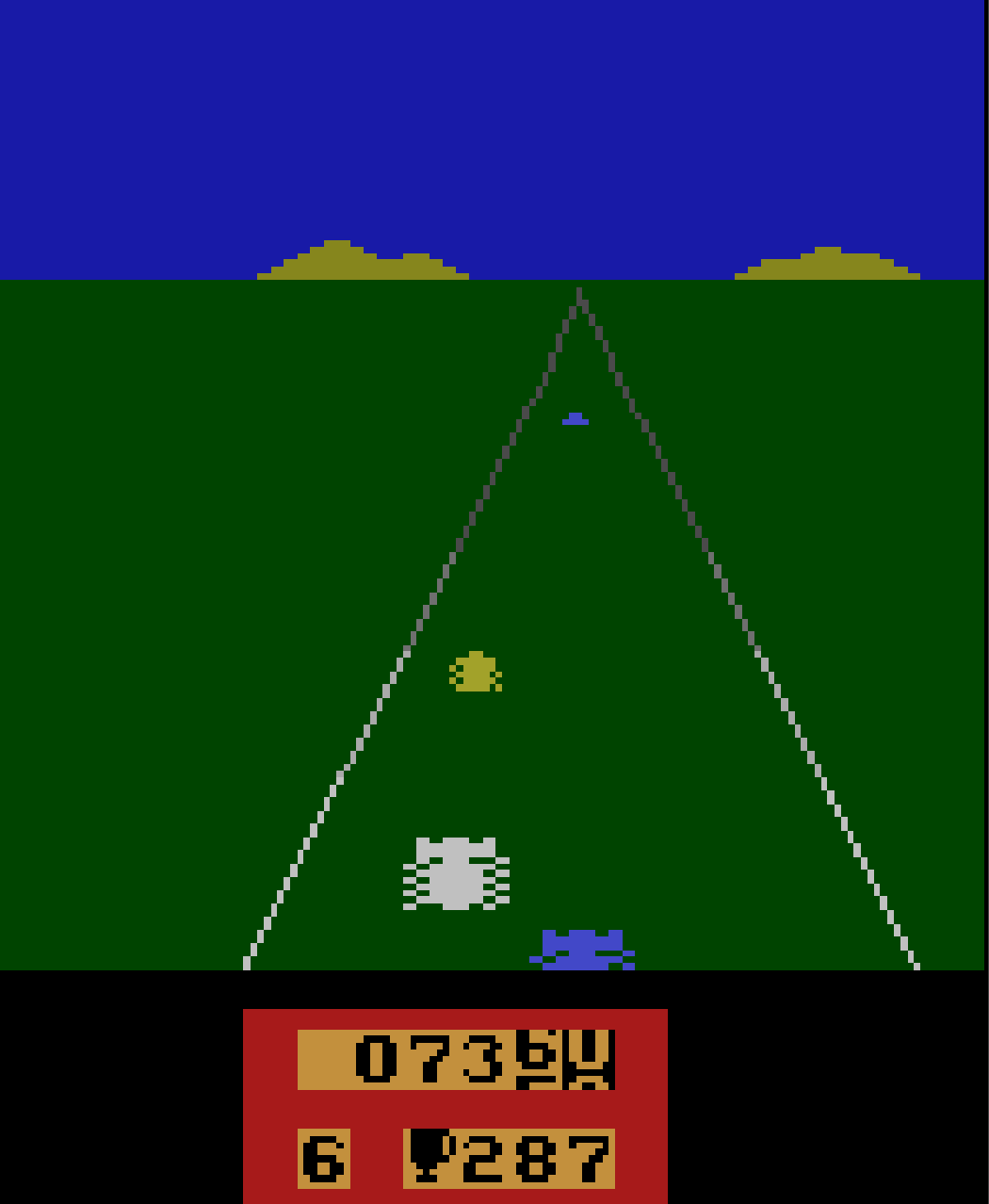}
\caption{}
\label{fig:enduro_h}
\end{subfigure}
~
\begin{subfigure}[b]{0.11\textwidth}
\includegraphics[width= \linewidth,height=3cm]{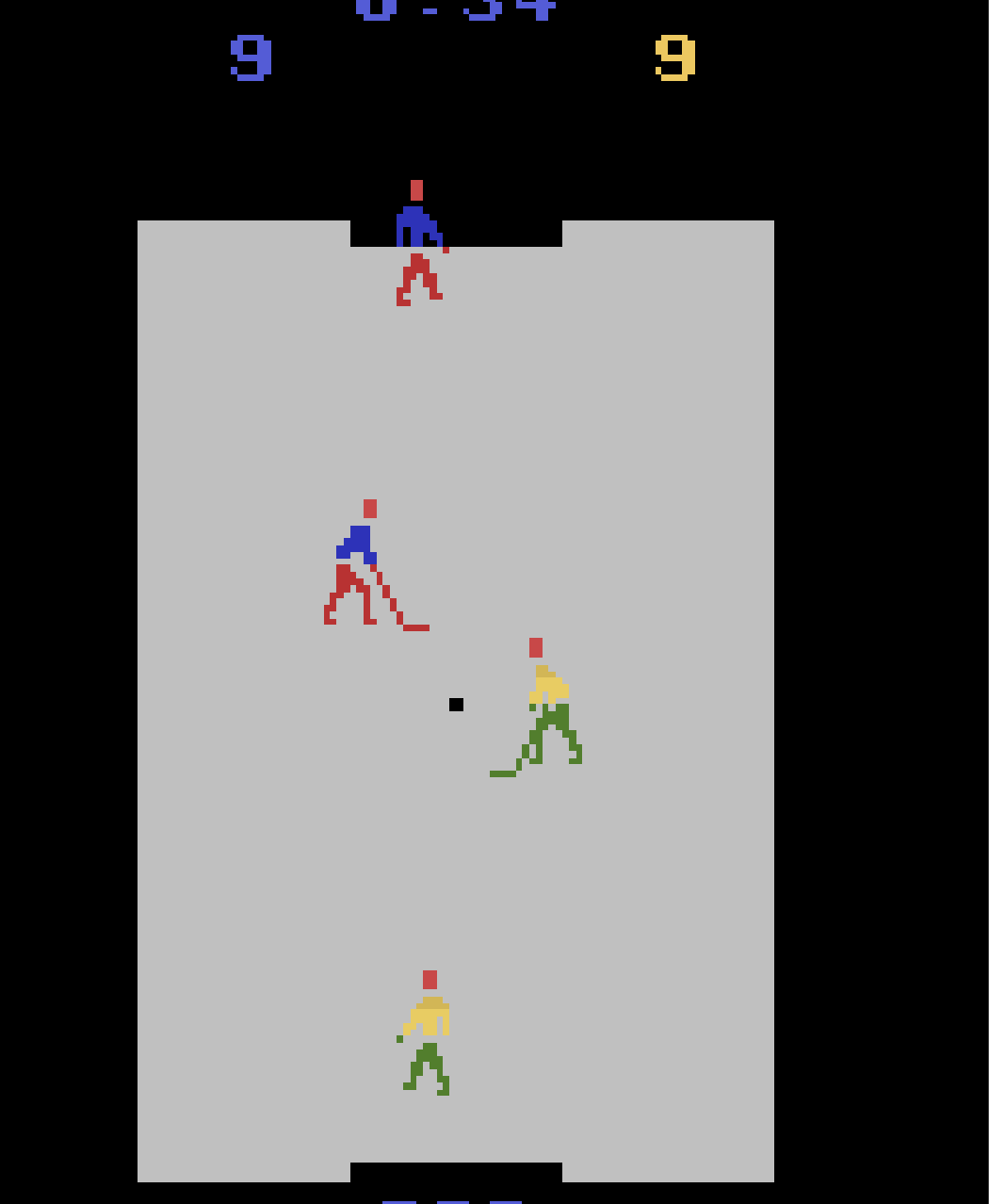}
\caption{}
\label{fig:icehockey_h}
\end{subfigure}
 ~ 
\begin{subfigure}[b]{0.11\textwidth}
\includegraphics[width= \linewidth,height=3cm]{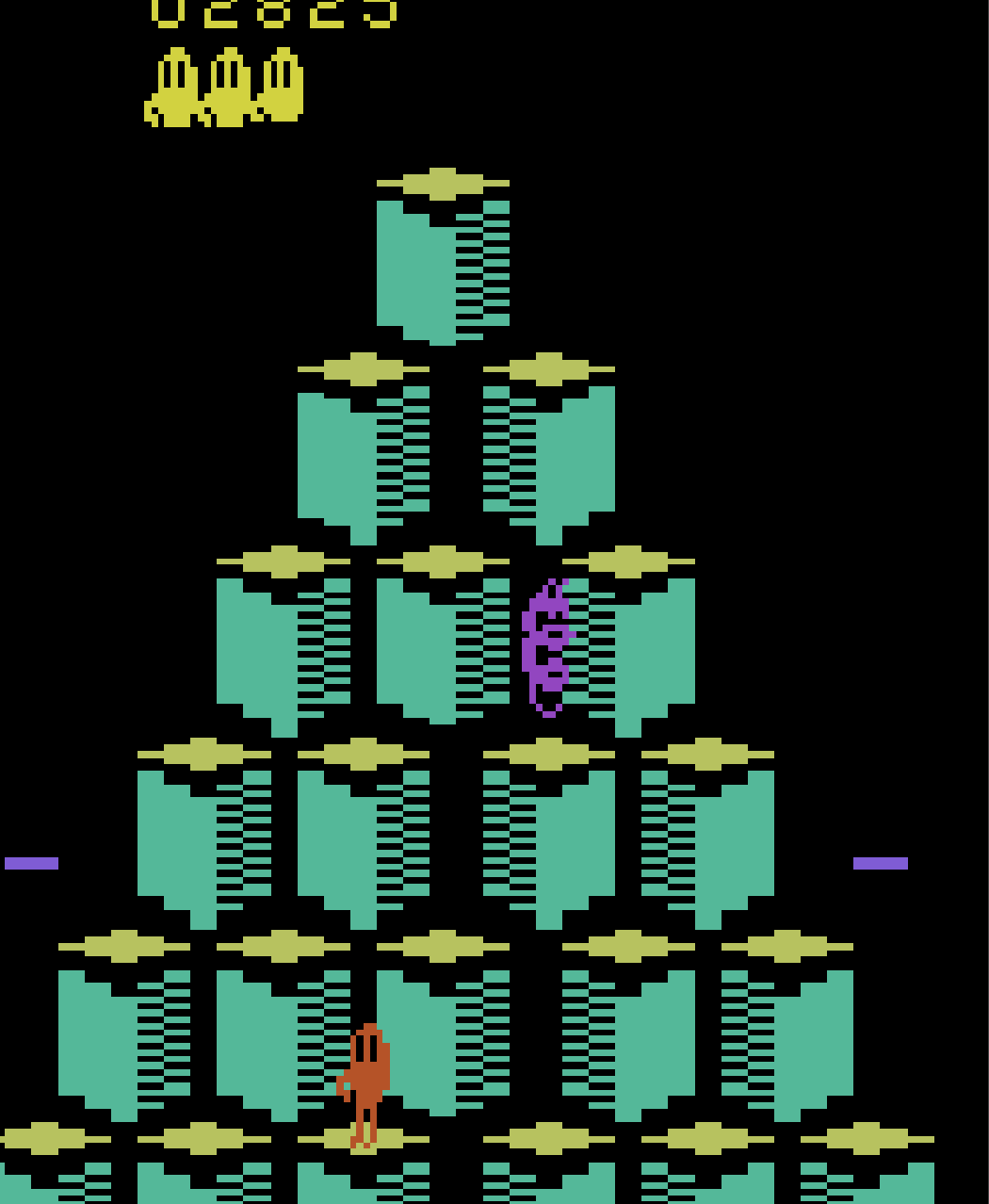}
\caption{}
\label{fig:qbert_h}
\end{subfigure}
~
\begin{subfigure}[b]{0.11\textwidth}
\includegraphics[width= \linewidth,height=3cm]{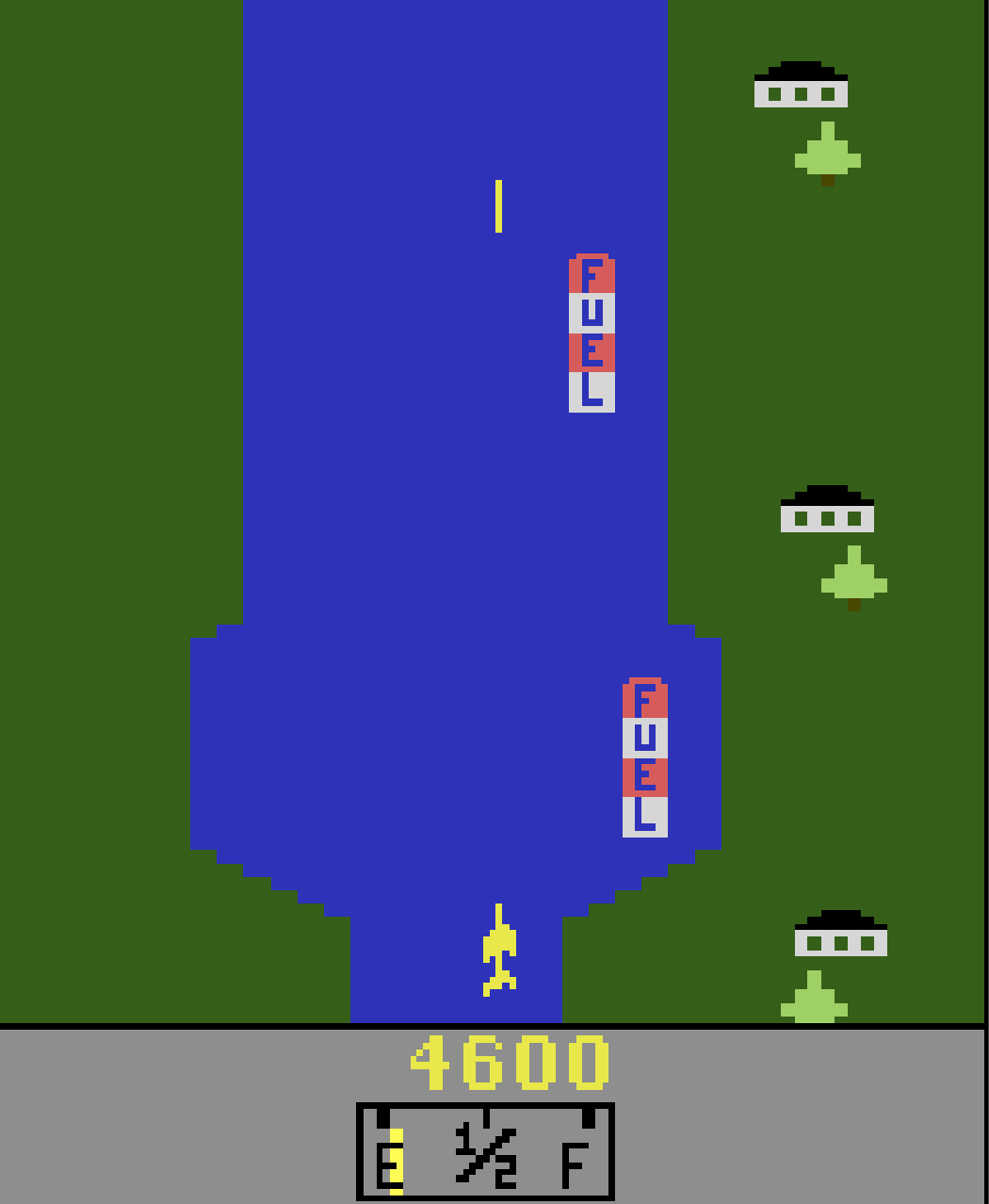}
\caption{}
\label{fig:riverraid_h}
\end{subfigure}
~
\begin{subfigure}[b]{0.11\textwidth}
\includegraphics[width= \linewidth,height=3cm]{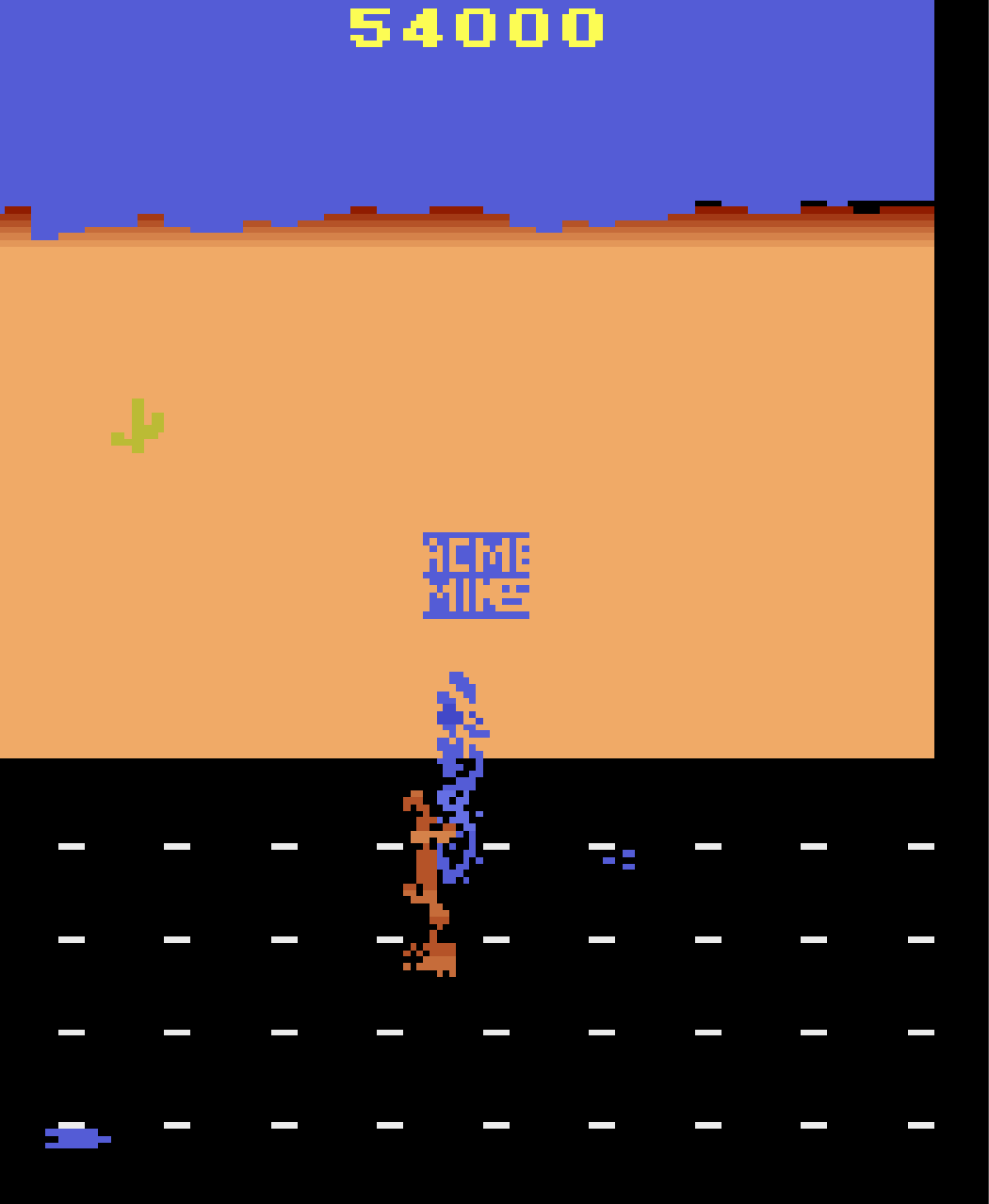}
\caption{}
\label{fig:roadrunner_h}
\end{subfigure}
~
\begin{subfigure}[b]{0.11\textwidth}
\includegraphics[width= \linewidth,height=3cm]{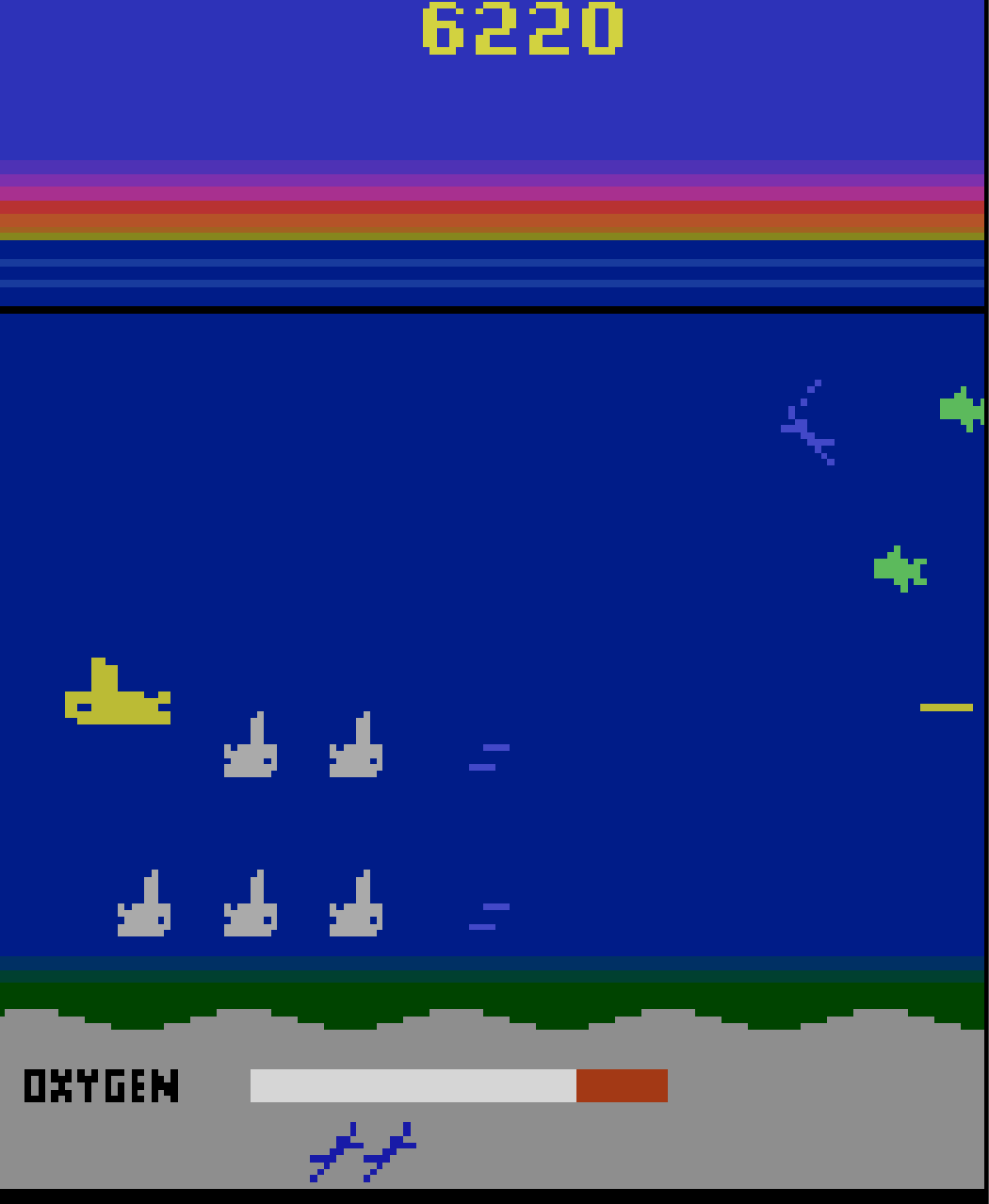}
\caption{}
\label{fig:seaquest_h}
\end{subfigure}
 
\caption{High risk states learnt by Q-SANE in the 8 game sub-suite}
\label{fig:all_states_high_risk}
\end{figure*}

\begin{figure*}[t!]
\begin{subfigure}[b]{0.11\textwidth}
\includegraphics[width= \linewidth,height=3cm]{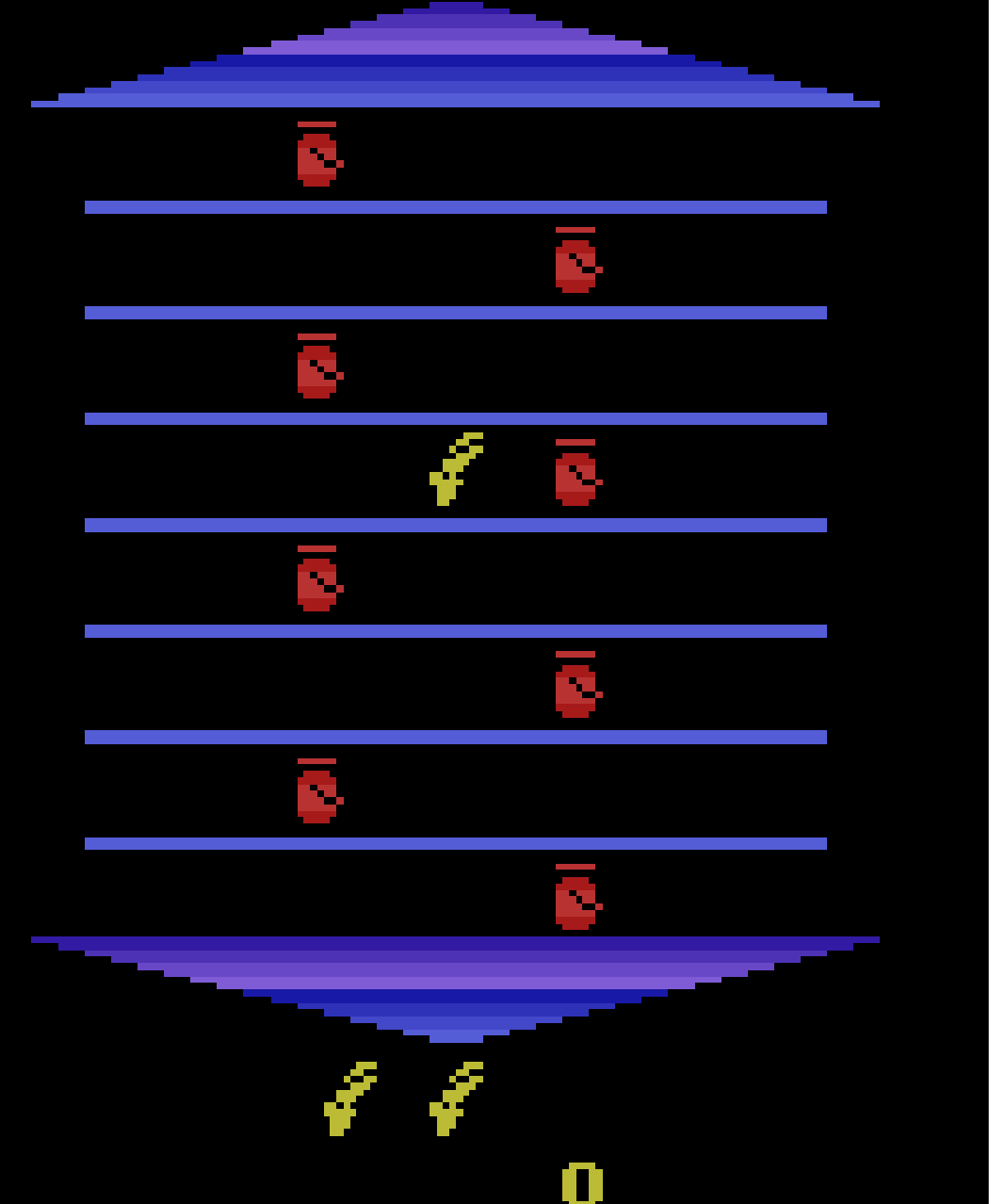}
\caption{}
\label{fig:asterix_l}
\end{subfigure}
~
\begin{subfigure}[b]{0.11\textwidth}
\includegraphics[width= \linewidth,height=3cm]{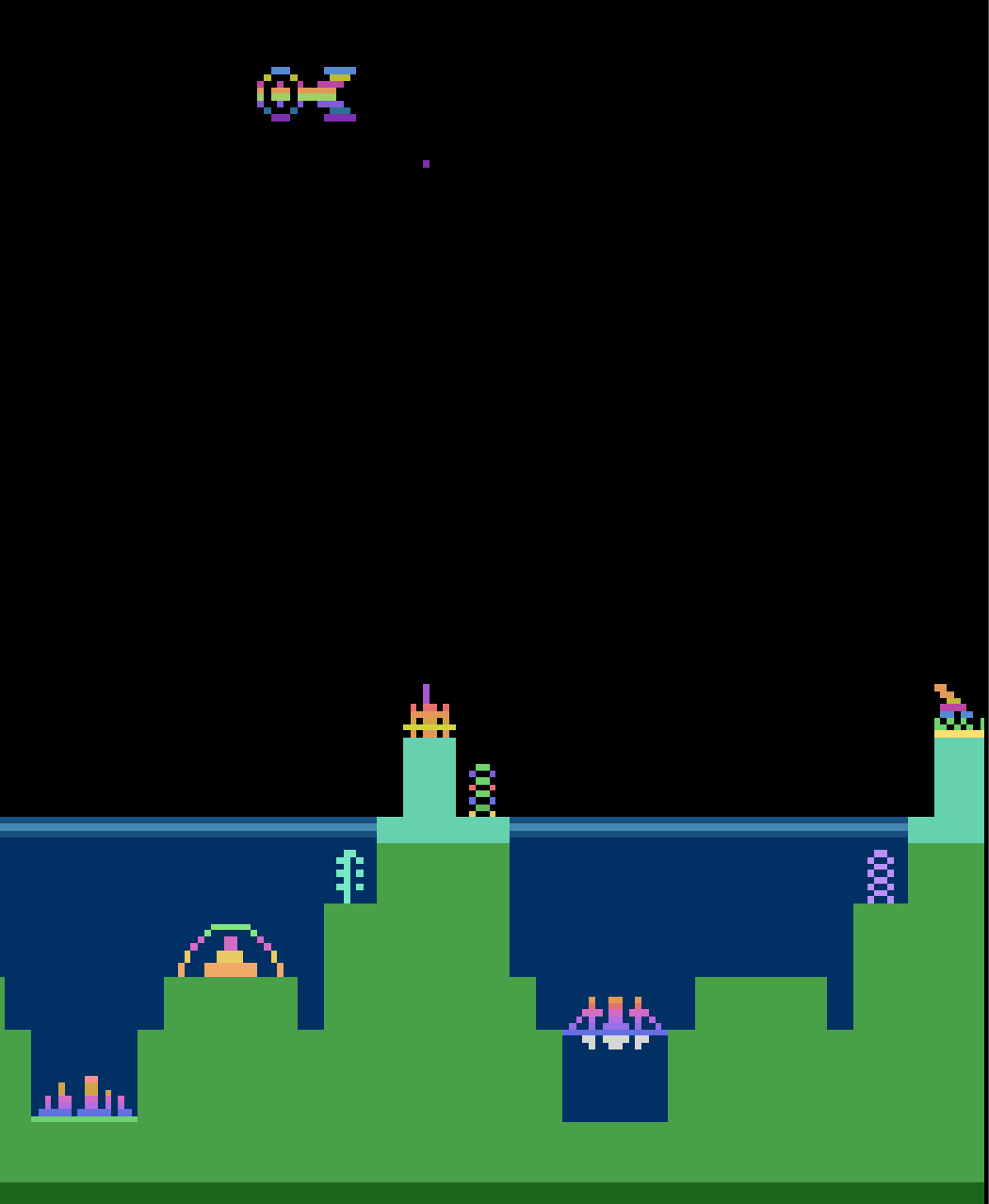}
\caption{}
\label{fig:atlantis_l}

\end{subfigure}
~
\begin{subfigure}[b]{0.11\textwidth}
\includegraphics[width= \linewidth,height=3cm]{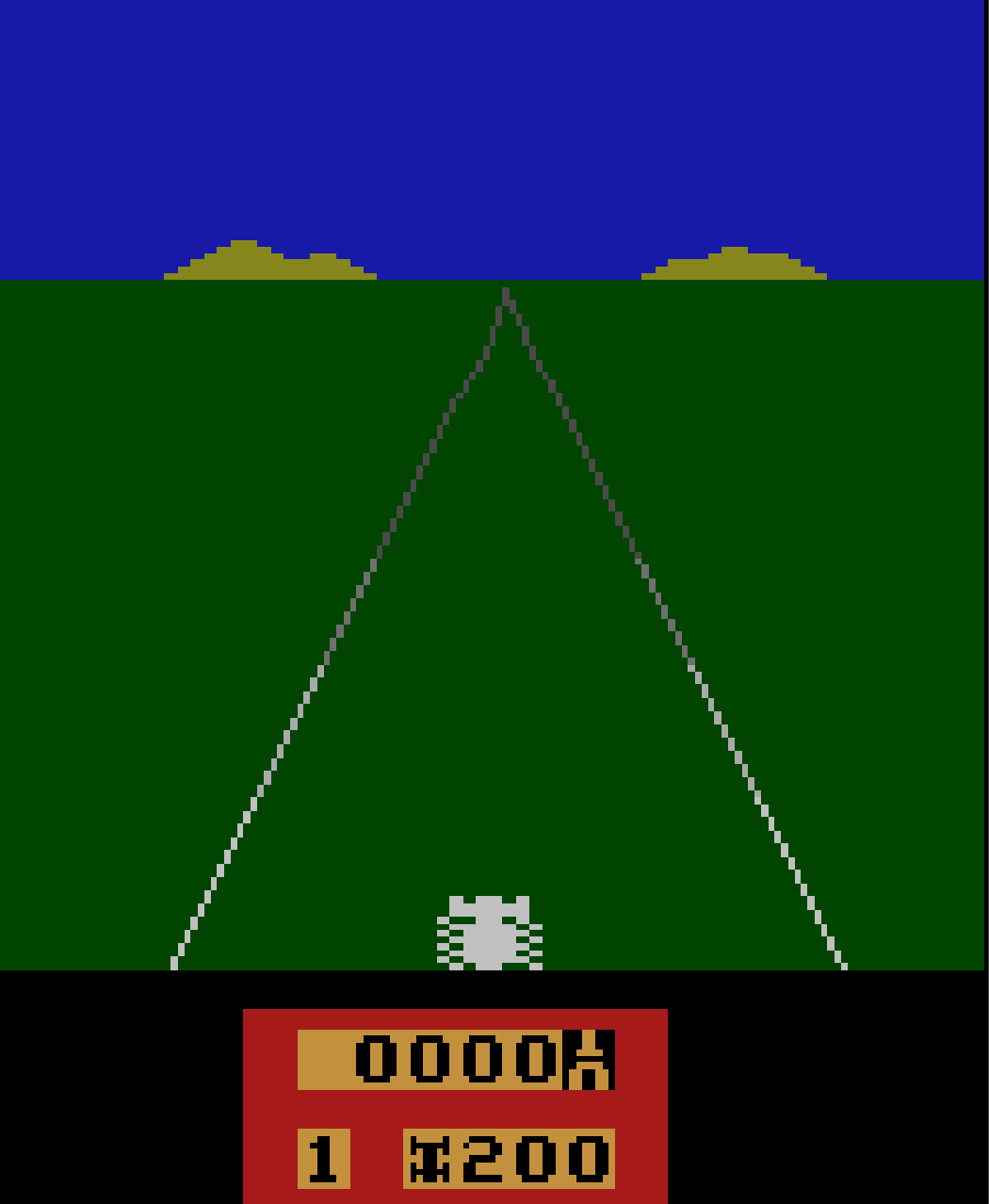}
\caption{}
\label{fig:enduro_l}

\end{subfigure}
~
\begin{subfigure}[b]{0.11\textwidth}
\includegraphics[width= \linewidth,height=3cm]{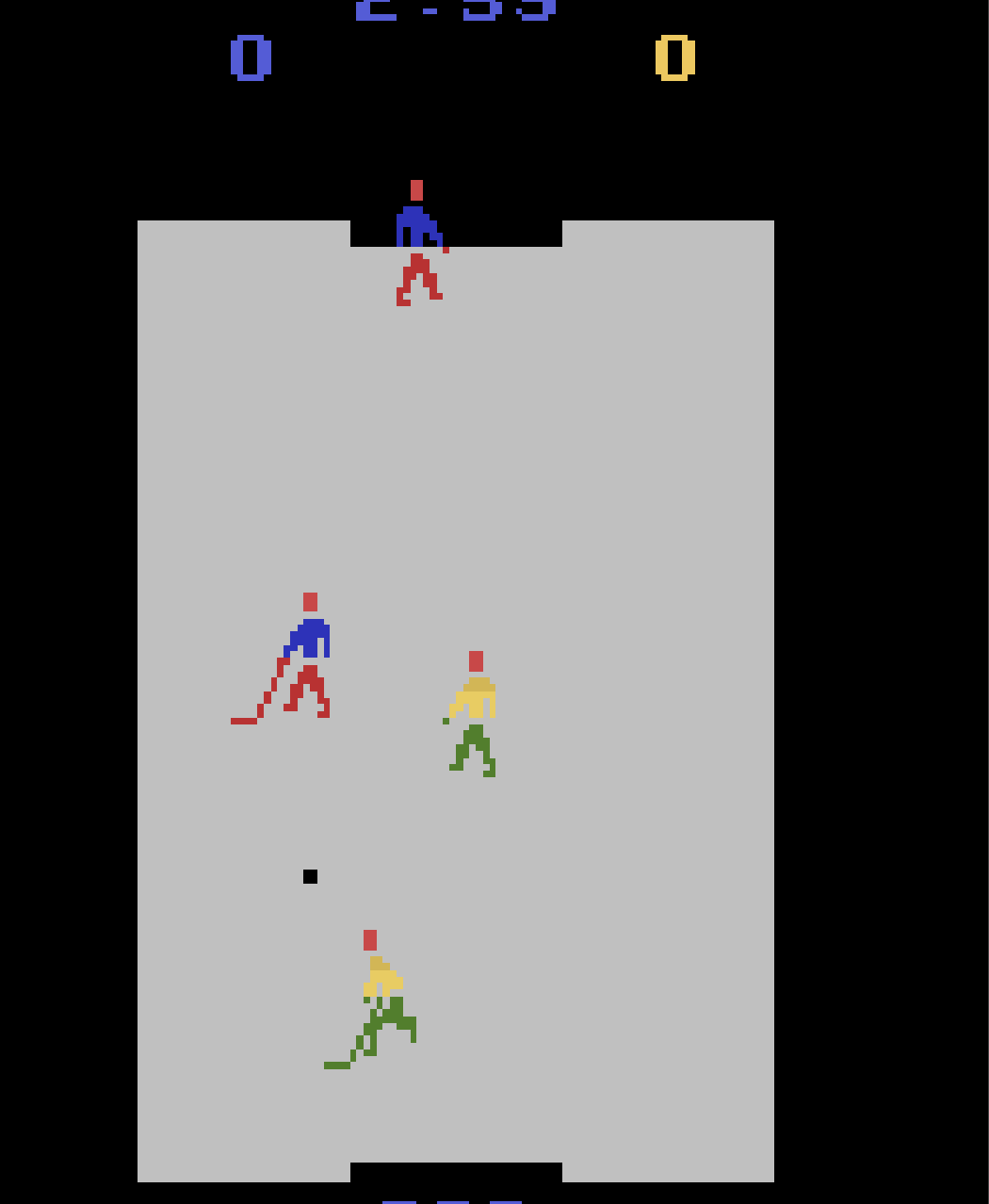}
\caption{}
\label{fig:icehockey_l}

\end{subfigure}
 ~ 
\begin{subfigure}[b]{0.11\textwidth}
\includegraphics[width= \linewidth,height=3cm]{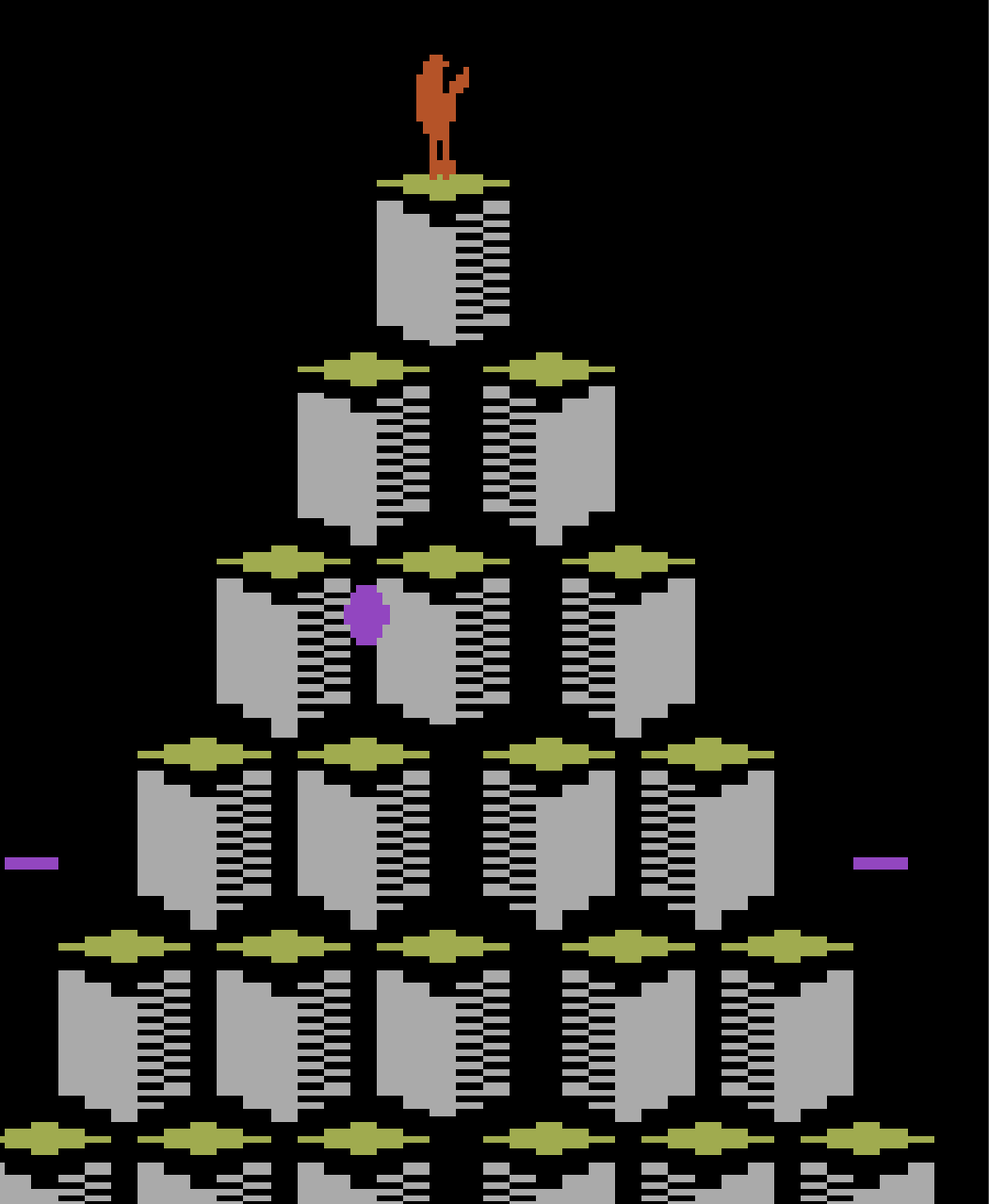}
\caption{}
\label{fig:qbert_l}

\end{subfigure}
~
\begin{subfigure}[b]{0.11\textwidth}
\includegraphics[width= \linewidth,height=3cm]{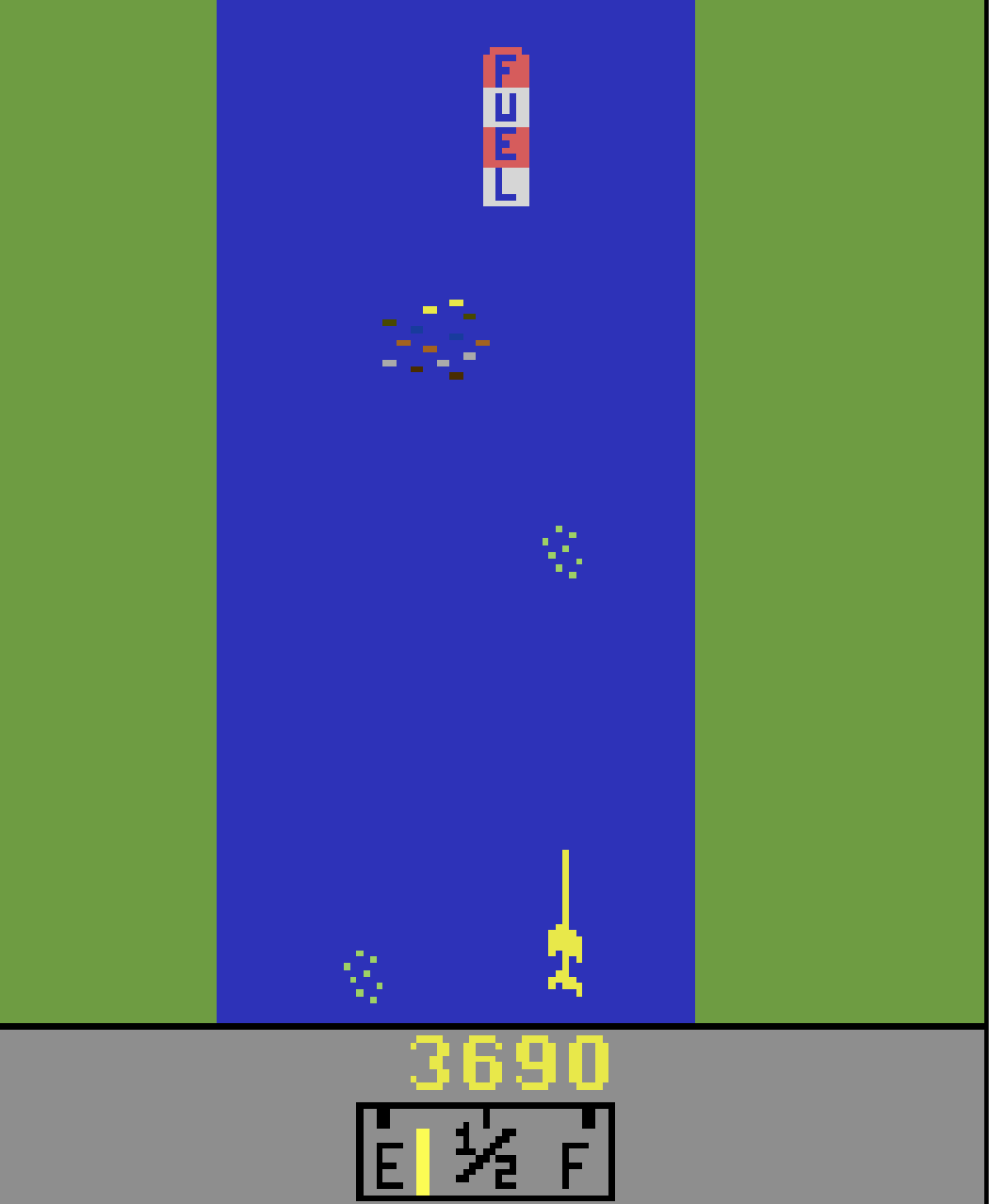}
\caption{}
\label{fig:riverraid_l}

\end{subfigure}
~
\begin{subfigure}[b]{0.11\textwidth}
\includegraphics[width= \linewidth,height=3cm]{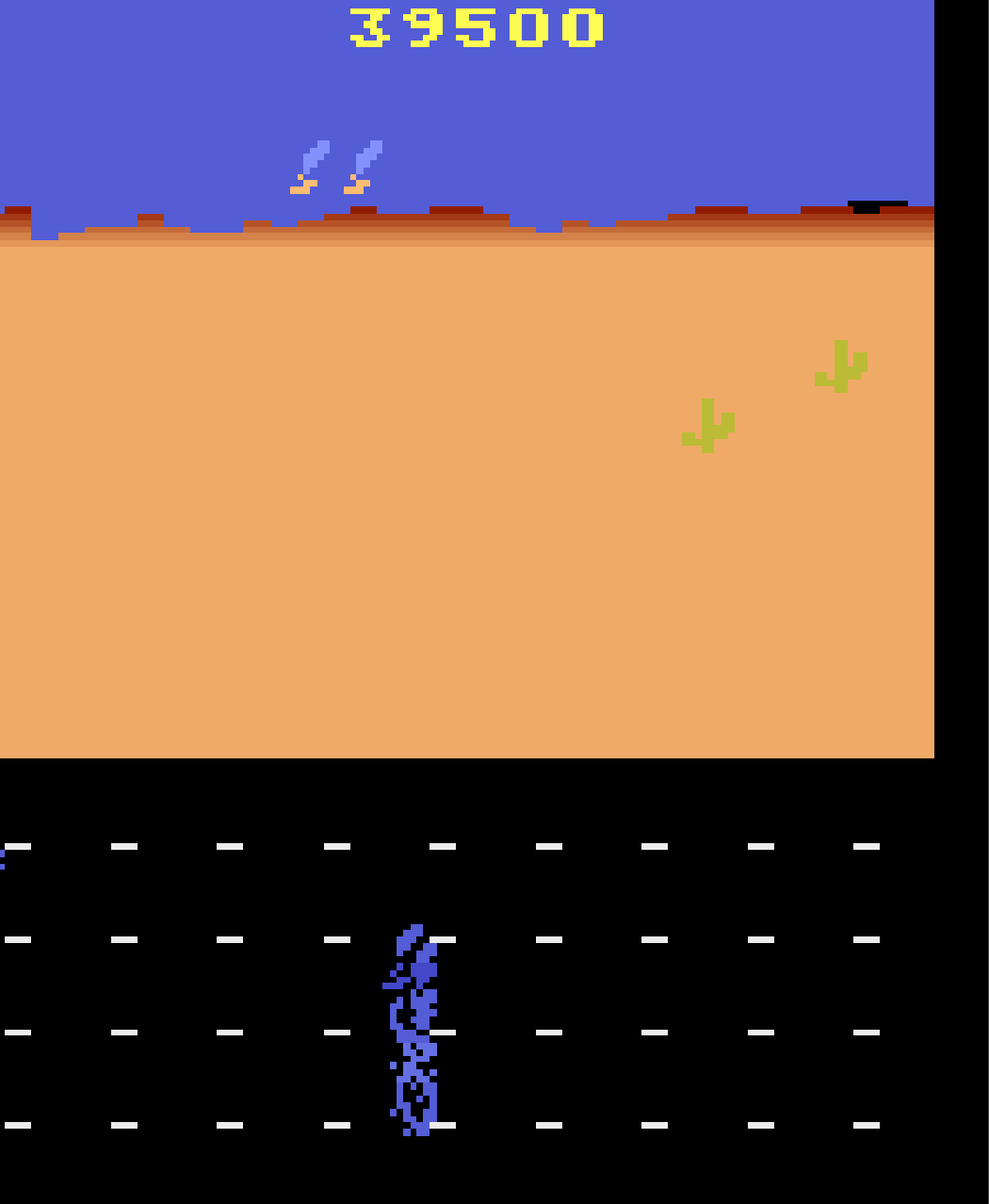}
\caption{}
\label{fig:roadrunner_l}

\end{subfigure}
~
\begin{subfigure}[b]{0.11\textwidth}
\includegraphics[width= \linewidth,height=3cm]{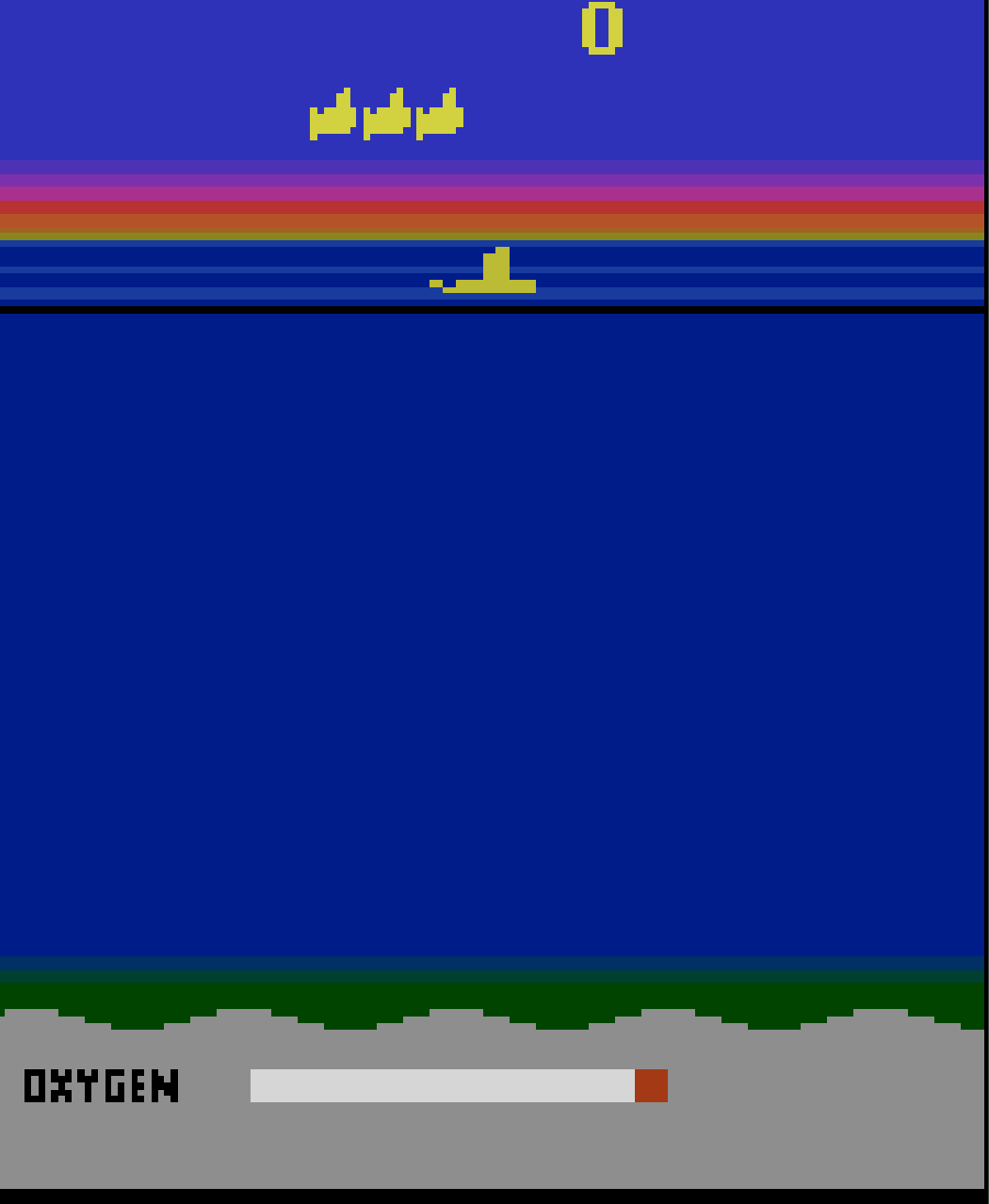}
\caption{}
\label{fig:seaquest_l}

\end{subfigure}
 
\caption{Low risk states learnt by Q-SANE in the 8 game sub-suite}
\label{fig:all_states_low_risk}
\end{figure*} 
\subsection{Variational View of SANE DQNs}
In SANE DQNs, we allow the network to use a different posterior approximation $q(\theta|s)$ for different states but restrict the perturbations that is added to all parameters to be sampled by the same distribution given a state $s$.
Similar to NoisyNets, for the unperturbed parameters of the convolutional layers and the perturbation module, we consider $q(\theta^{b}|s) = \mathcal{N}(\mu_{b}, \epsilon I),q(\Theta|s) = \mathcal{N}(\mu_{\Theta}, \epsilon I) $ and for the parameters of the fully connected layers,  we take $q(\theta^{p}|s) = \mathcal{N}(\mu^{p}, \sigma^2(h(s; \theta^b); \Theta)) \forall s$. We take the prior $p(\theta) = \mathcal{N}(0, I)$ for all the parameters of the network. It follows that the objective to maximize is the same as objective (\ref{eq:final_elbo_nn}), but where the parameters $\hat{\theta_i}$ are drawn from  $q(\theta|s^i_t)$.

In our experiments, we compare two different SANE DQN variants, namely, the simple-SANE DQN and the Q-SANE DQN. Both these SANE DQNs have the same network structure as shown in Figure \ref{fig:SAE_module}. Q-SANE DQNs and simple-SANE DQNs differ in the additional features that are added to the perturbation module. Simple-SANE DQNs add no additional features to aid the perturbation module. On the other hand, Q-SANE DQNs use the non-noisy Q-values of the state as additional features. The non-noisy Q-values are computed via a forward pass of the neural network with no perturbations applied to $\theta^{p}$. Adding Q-values as features to the perturbation module can be useful, as a state where all the action values take similar values could be an indication of a low risk state and vice versa.

\section{Experiments}
\label{sec:exp}

We conduct our experiments on a suite of 11 Atari games. This suite contains 8 games that exhibit both high and low risk states (see Figures \ref{fig:risk}, \ref{fig:all_states_high_risk} and \ref{fig:all_states_low_risk}) that we expect would benefit from state aware exploratory behaviour. We expect SANE not to have any notable benefit in the other 3 games. 

\subsection{Atari Test Suite}

The 11 game Atari test suite has an 8 game sub-suite, consisting of the games Asterix, Atlantis, Enduro, IceHockey, Qbert, Riverraid, RoadRunner and Seaquest. High risk and low risk states of these games (in order) are shown in Figures \ref{fig:all_states_high_risk} and \ref{fig:all_states_low_risk} respectively. The games in this sub-suite have properties that benefit agents when trained with SANE exploration. Most high risk states in these games, occur when the agent is either at risk of being hit (or captured) by an enemy or at risk of going out of bounds of the play area. Figures \ref{fig:asterix_h}, \ref{fig:atlantis_h}, \ref{fig:enduro_h}, \ref{fig:qbert_h}, \ref{fig:roadrunner_h} and \ref{fig:seaquest_h} illustrate this property of high risk states. Low risk states, on the other hand, correspond to those states where the agent has a lot of freedom to move around without risking loss of life. Most of the states in Figure \ref{fig:all_states_low_risk} demonstrate this. 

Additionally, there maybe other complex instances of high risk states. For instance, in Riverraid,  states where the agent is about to run out of fuel can be considered high risk states (Figure \ref{fig:riverraid_h}).  Moreover, sometimes the riskiness of a state can be hard to identify. This is illustrated by the high and low risk states of IceHockey shown in Figures \ref{fig:icehockey_h} and \ref{fig:icehockey_l} respectively. In games like IceHockey, high risk states are determined by the positions of the players and the puck. Figure \ref{fig:icehockey_h} is a high risk state as the puck's possession is being contested by both teams, while \ref{fig:icehockey_l} is low risk as the opponent is certain to gain the puck's possession in the near future. 
 
We also include 3 games, namely, FishingDerby, Boxing and Bowling, in our suite to check the sanity of SANE agents in games where we expect SANE exploration not to have any additional benefits over NoisyNets. 
  
\subsection{Parameter Initialization}

\begin{figure*}[t!]
\begin{subfigure}[b]{0.23\textwidth}
\includegraphics[width=\linewidth, height=3cm]{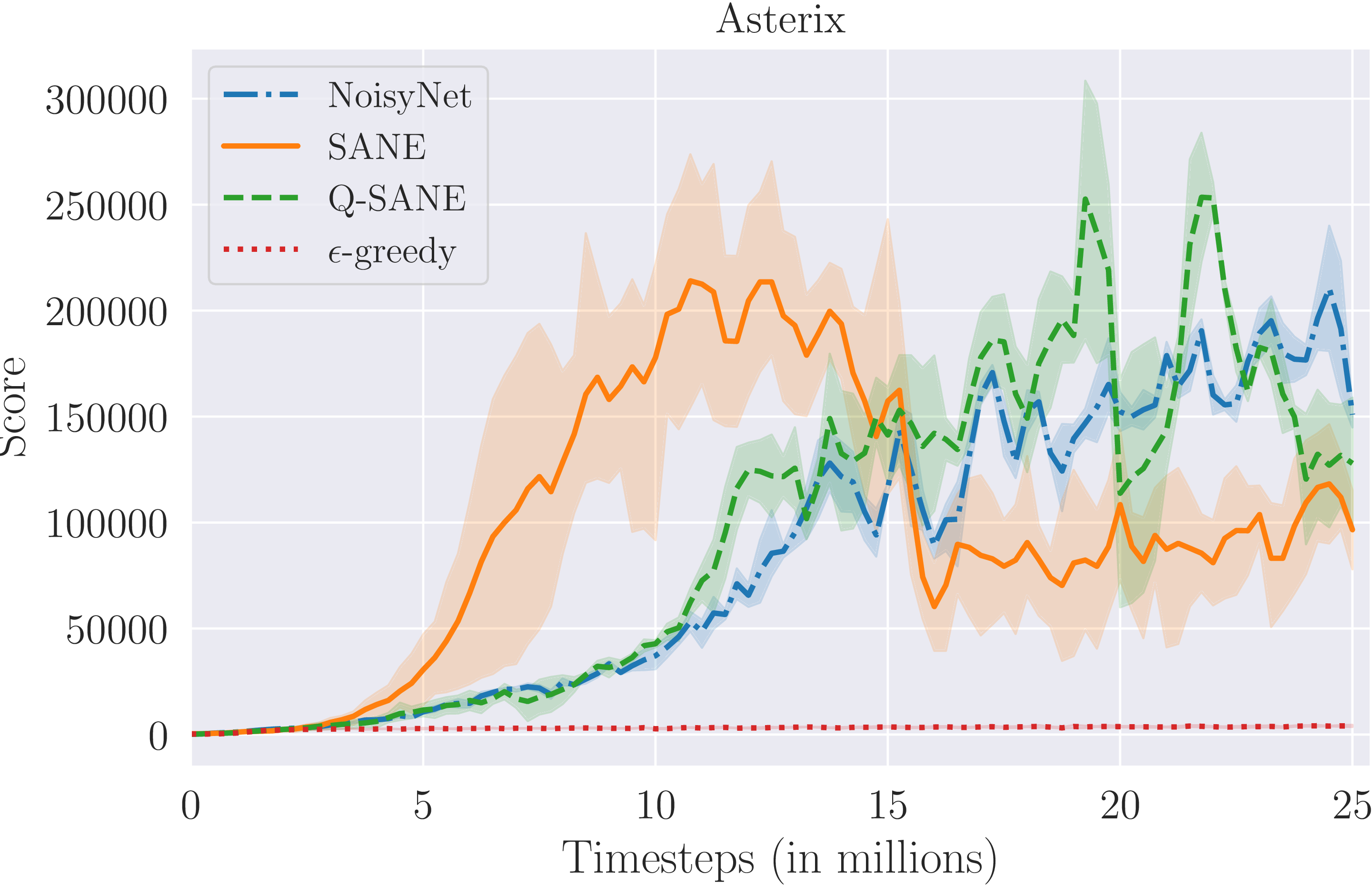}
\end{subfigure}
~
\begin{subfigure}[b]{0.23\textwidth}
\includegraphics[width=\linewidth,height=3cm]{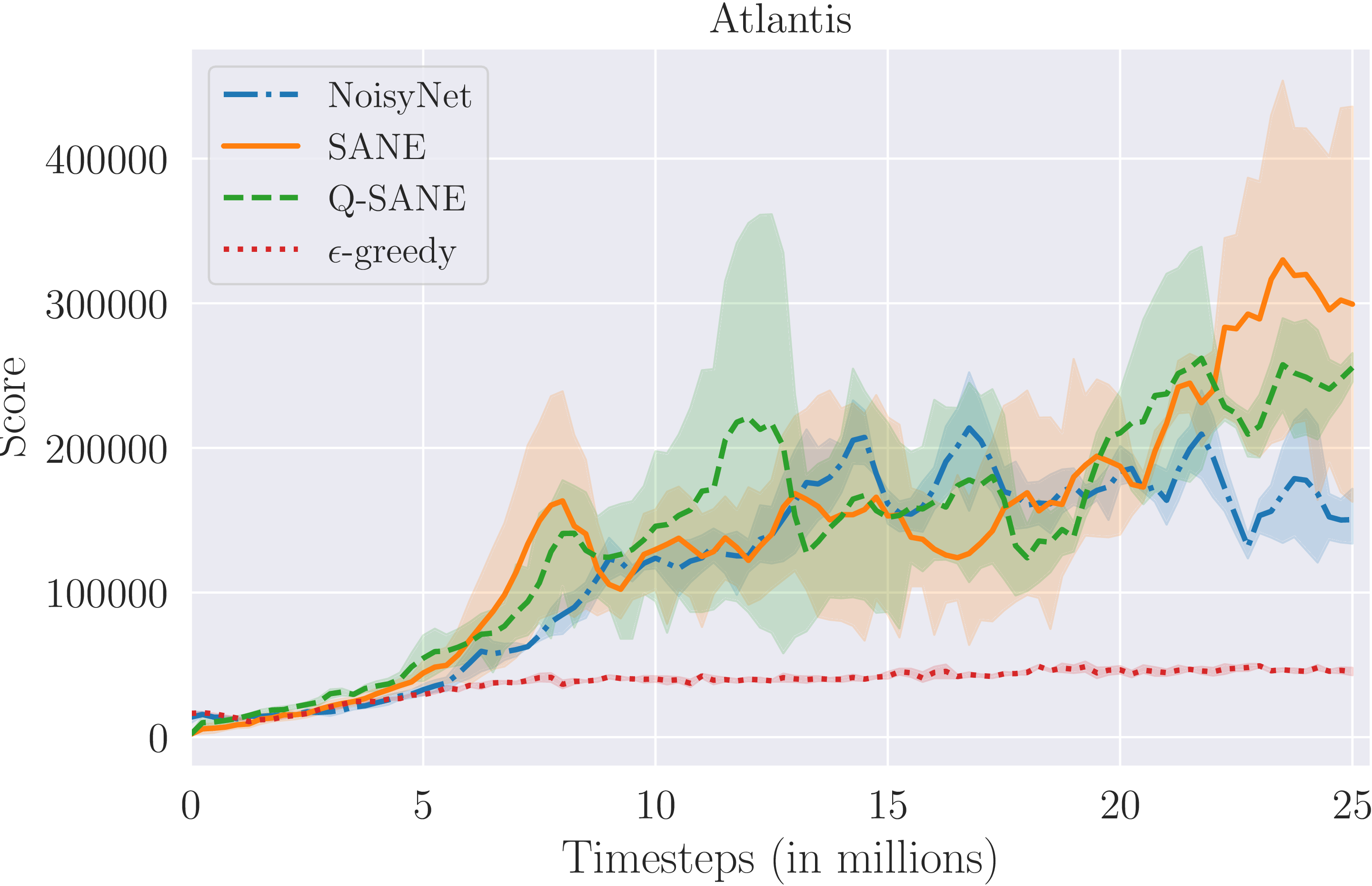}
\end{subfigure}
~
\begin{subfigure}[b]{0.23\textwidth}
\includegraphics[width=\linewidth,height=3cm]{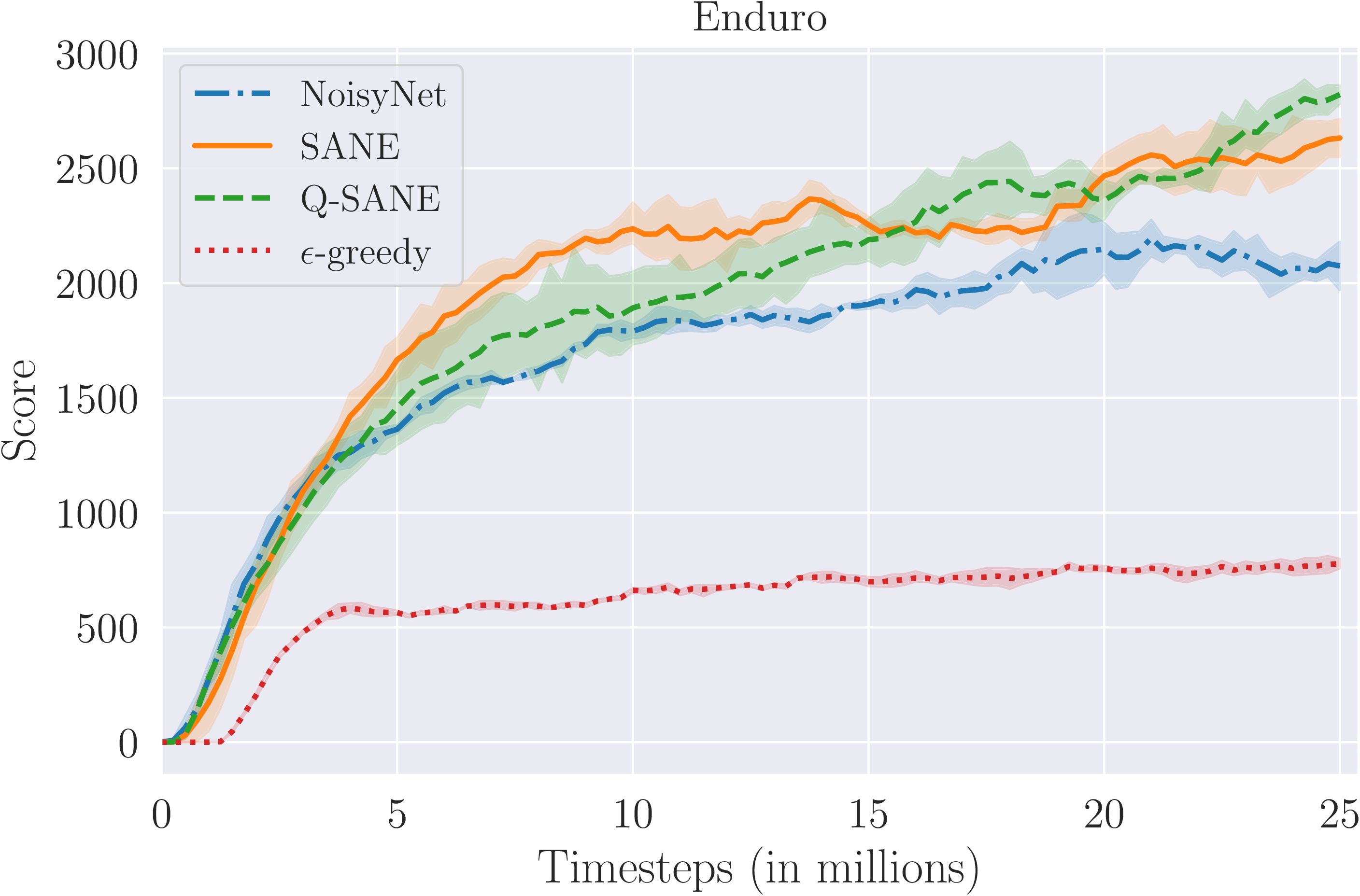}
\end{subfigure}
~
\begin{subfigure}[b]{0.23\textwidth}
\includegraphics[width=\linewidth,height=3cm]{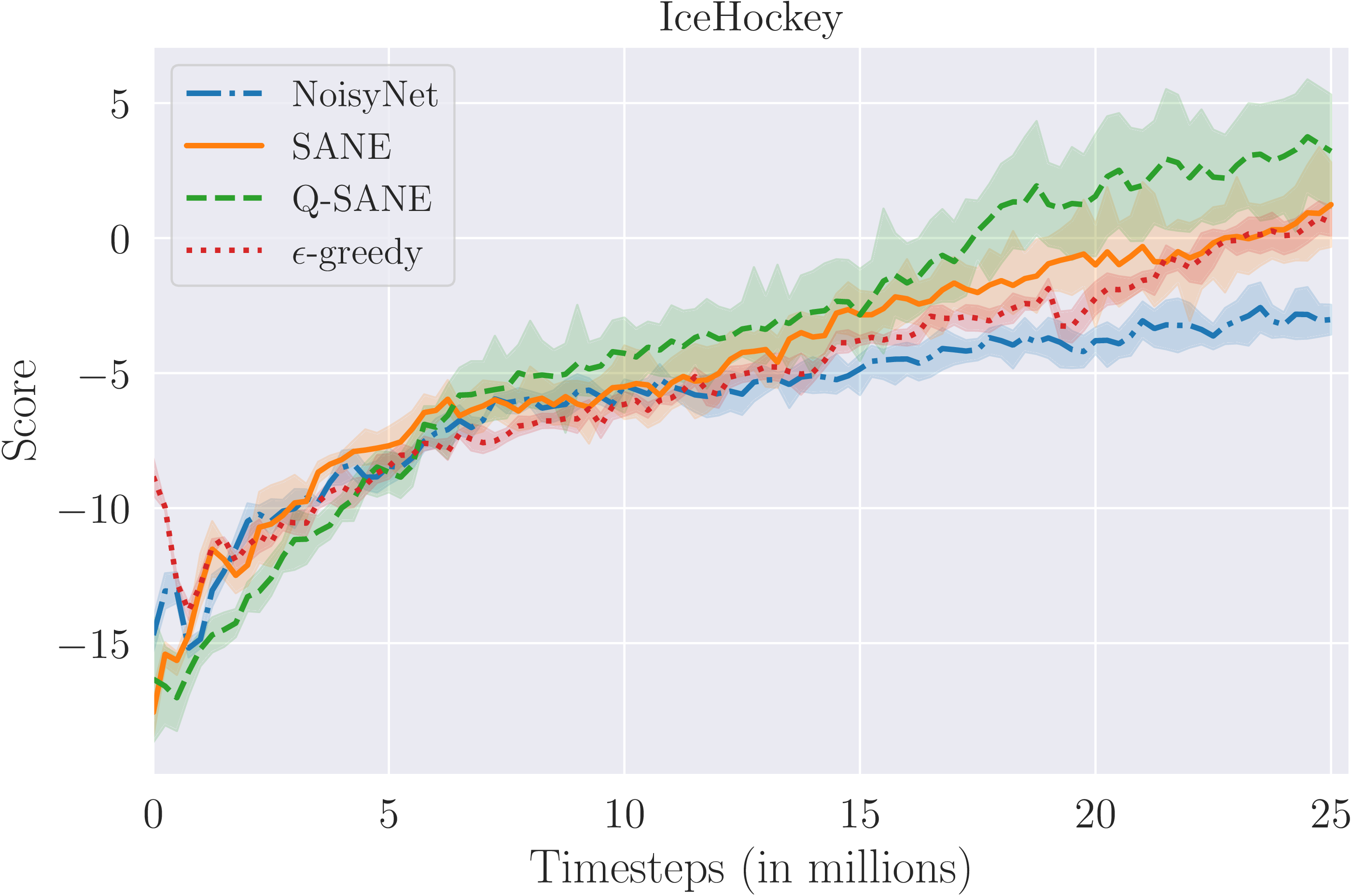}
\end{subfigure}
 \newline 

\begin{subfigure}[b]{0.23\textwidth}
\includegraphics[width=\linewidth,height=3cm]{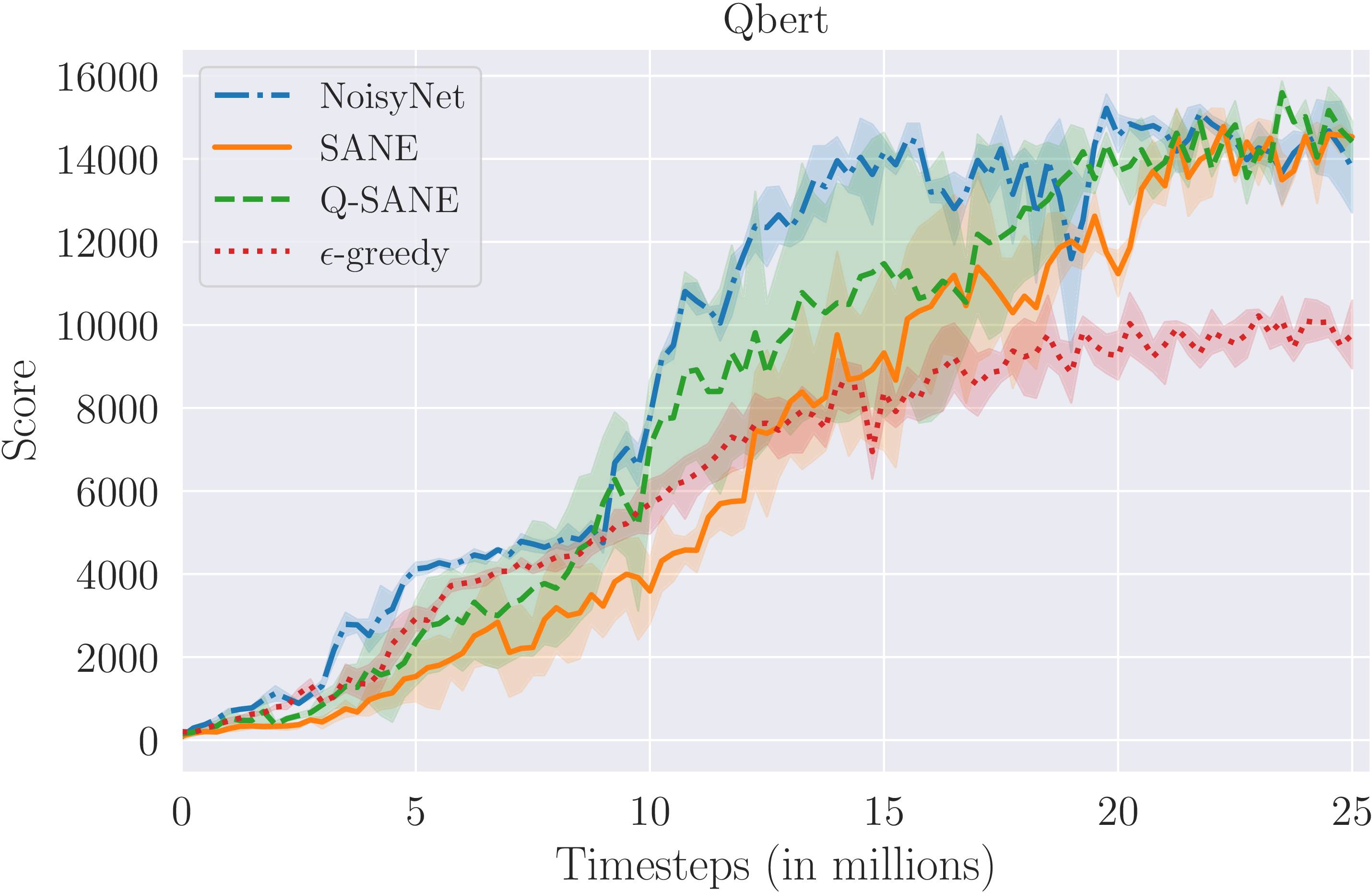}
\end{subfigure}
~
\begin{subfigure}[b]{0.23\textwidth}
\includegraphics[width=\linewidth,height=3cm]{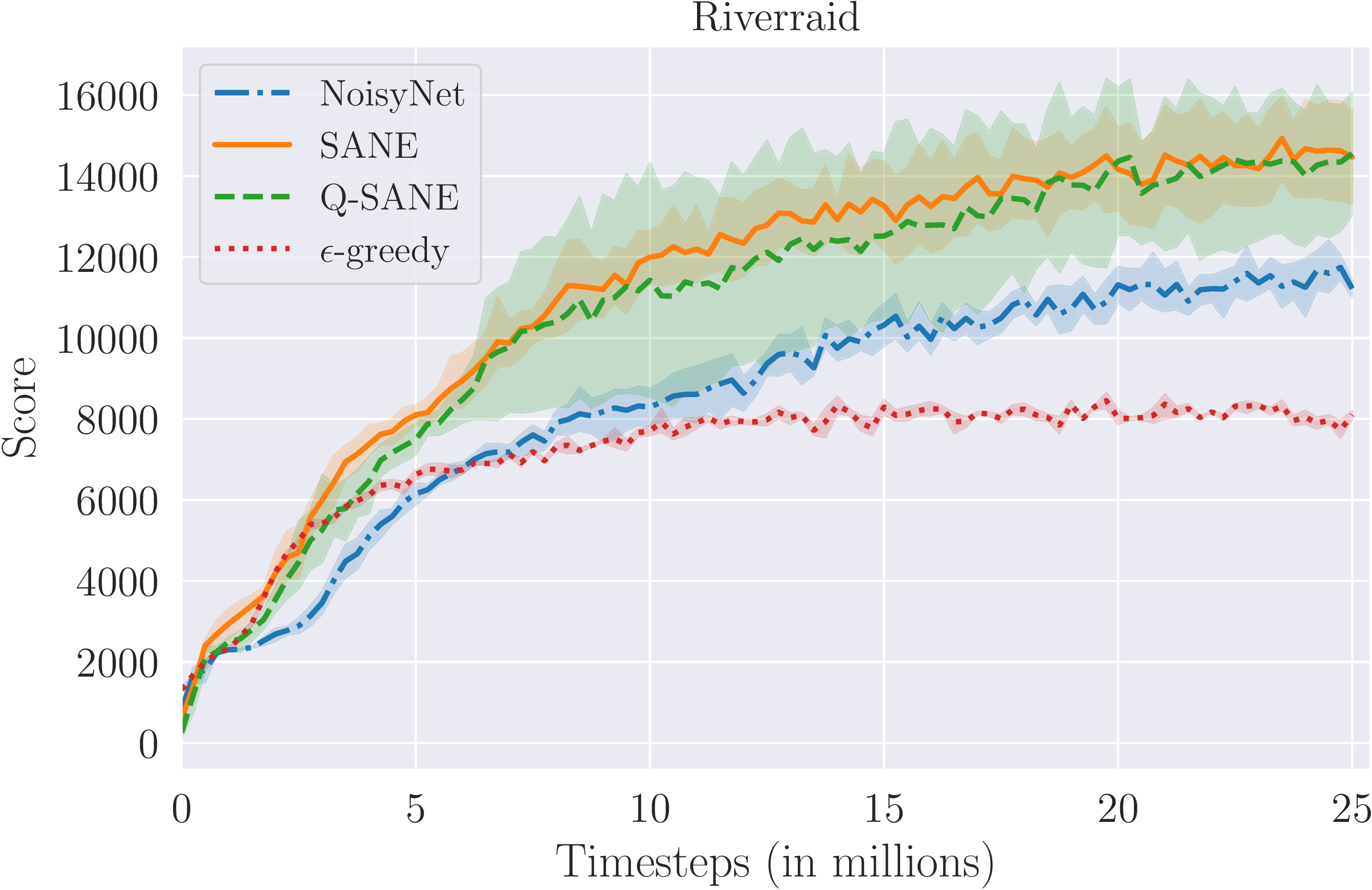}
\end{subfigure}
~
\begin{subfigure}[b]{0.23\textwidth}
\includegraphics[width=\linewidth,height=3cm]{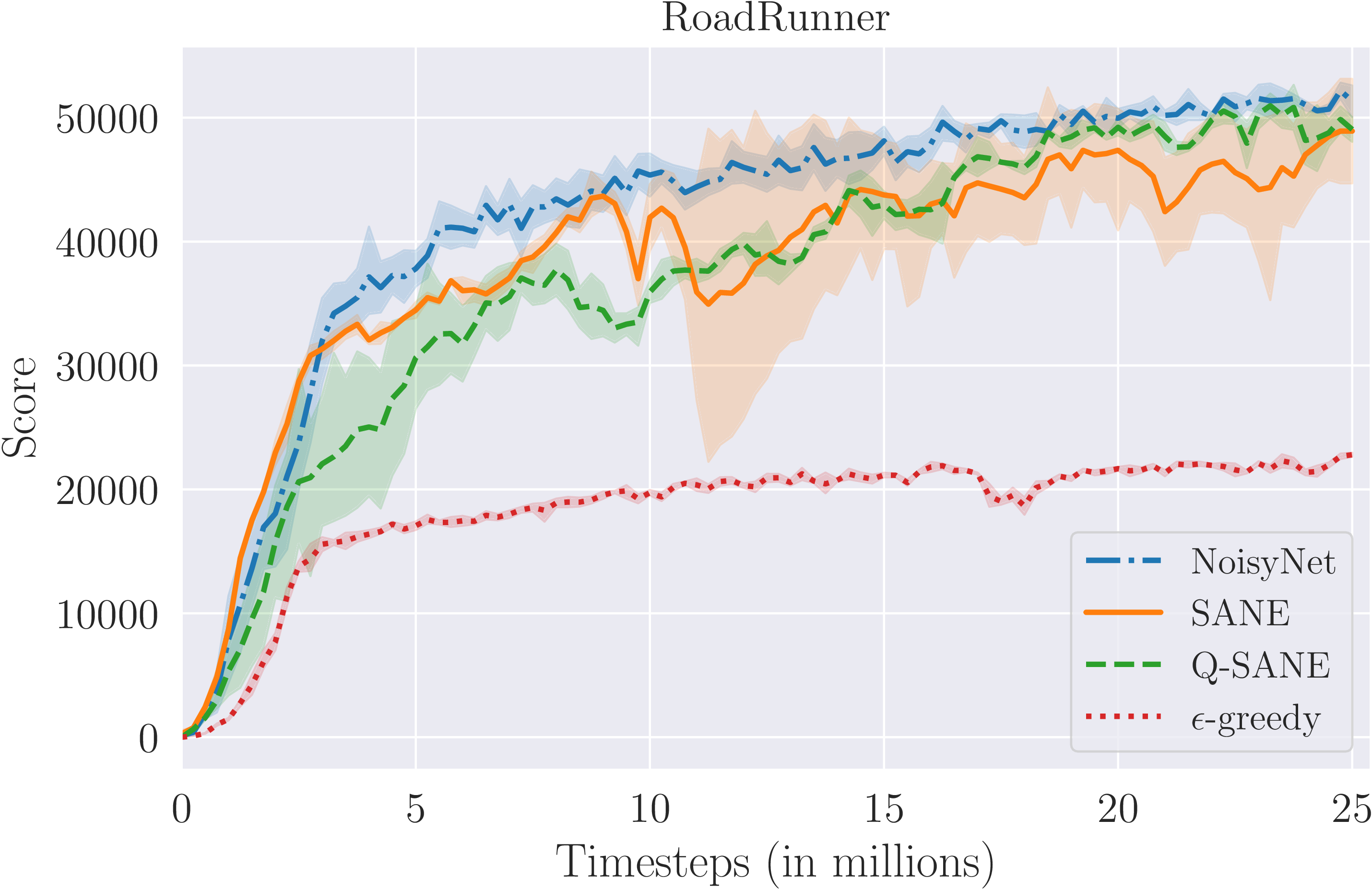}
\end{subfigure}
~
\begin{subfigure}[b]{0.23\textwidth}
\includegraphics[width=\linewidth,height=3cm]{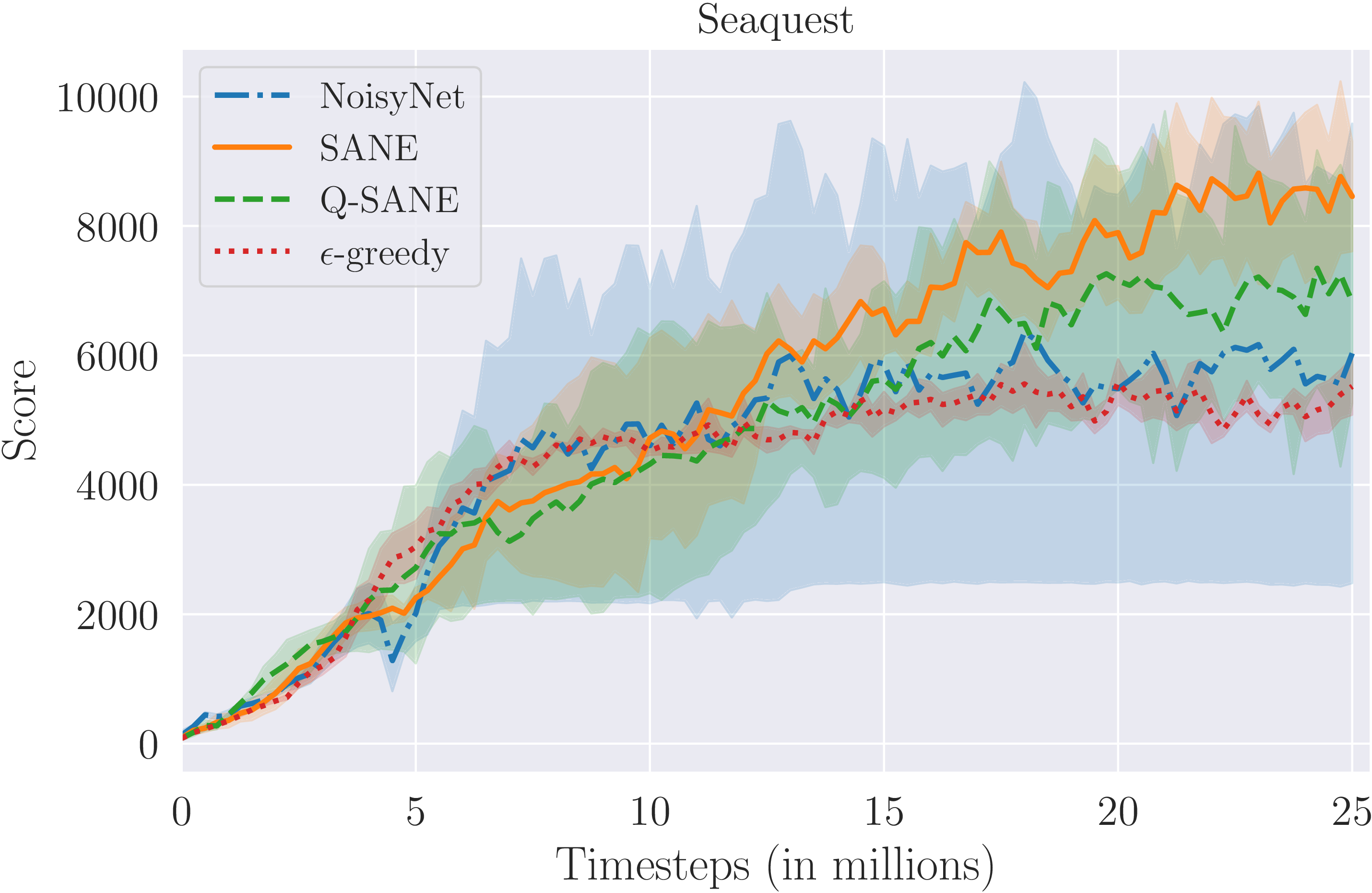}
\end{subfigure}
 \newline 

\hspace{1.5cm}
\begin{subfigure}[b]{0.23\textwidth}
\includegraphics[width=\linewidth,height=3cm]{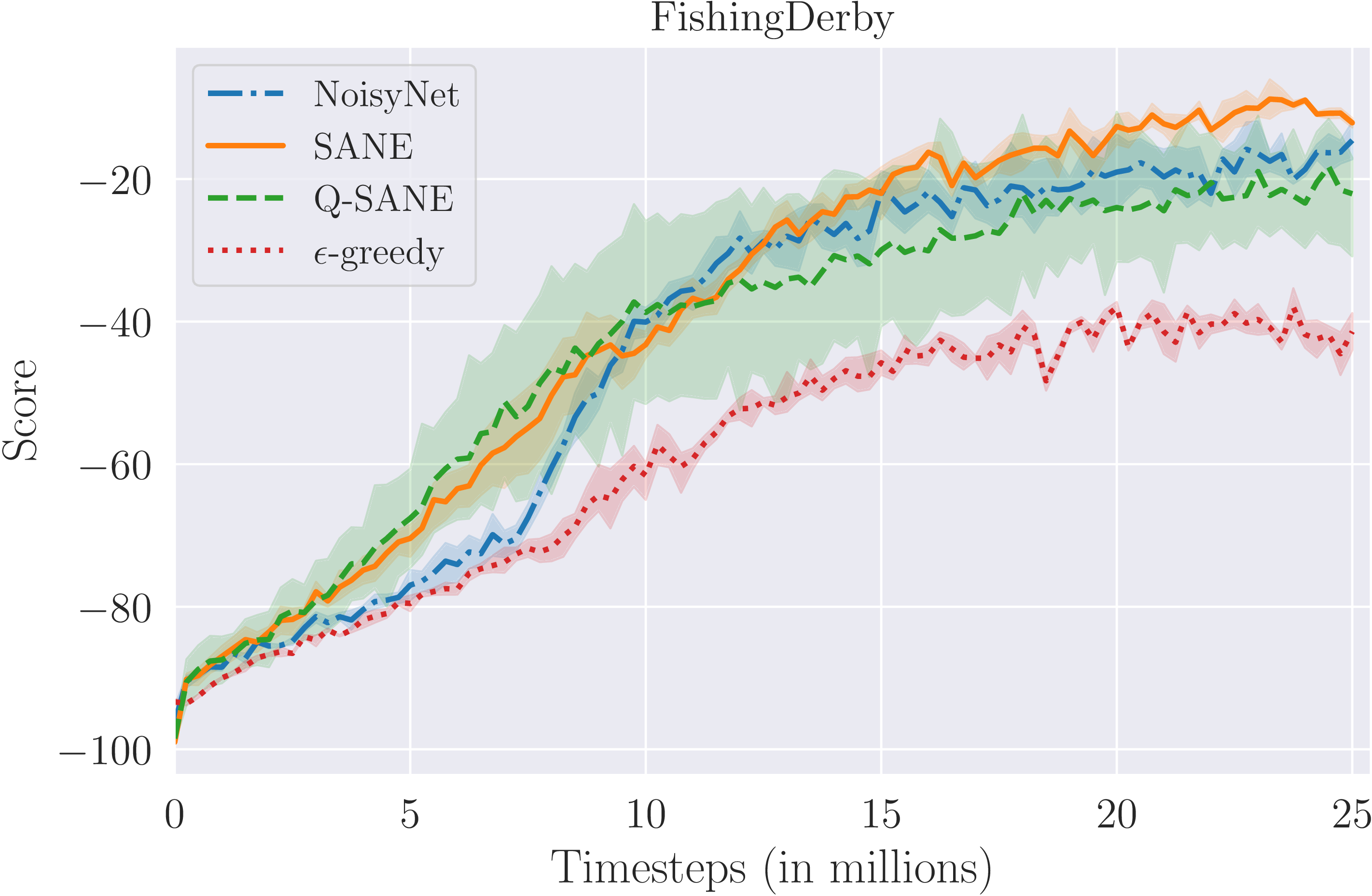}
\end{subfigure}
~
\begin{subfigure}[b]{0.23\textwidth}
\includegraphics[width=\linewidth,height=3cm]{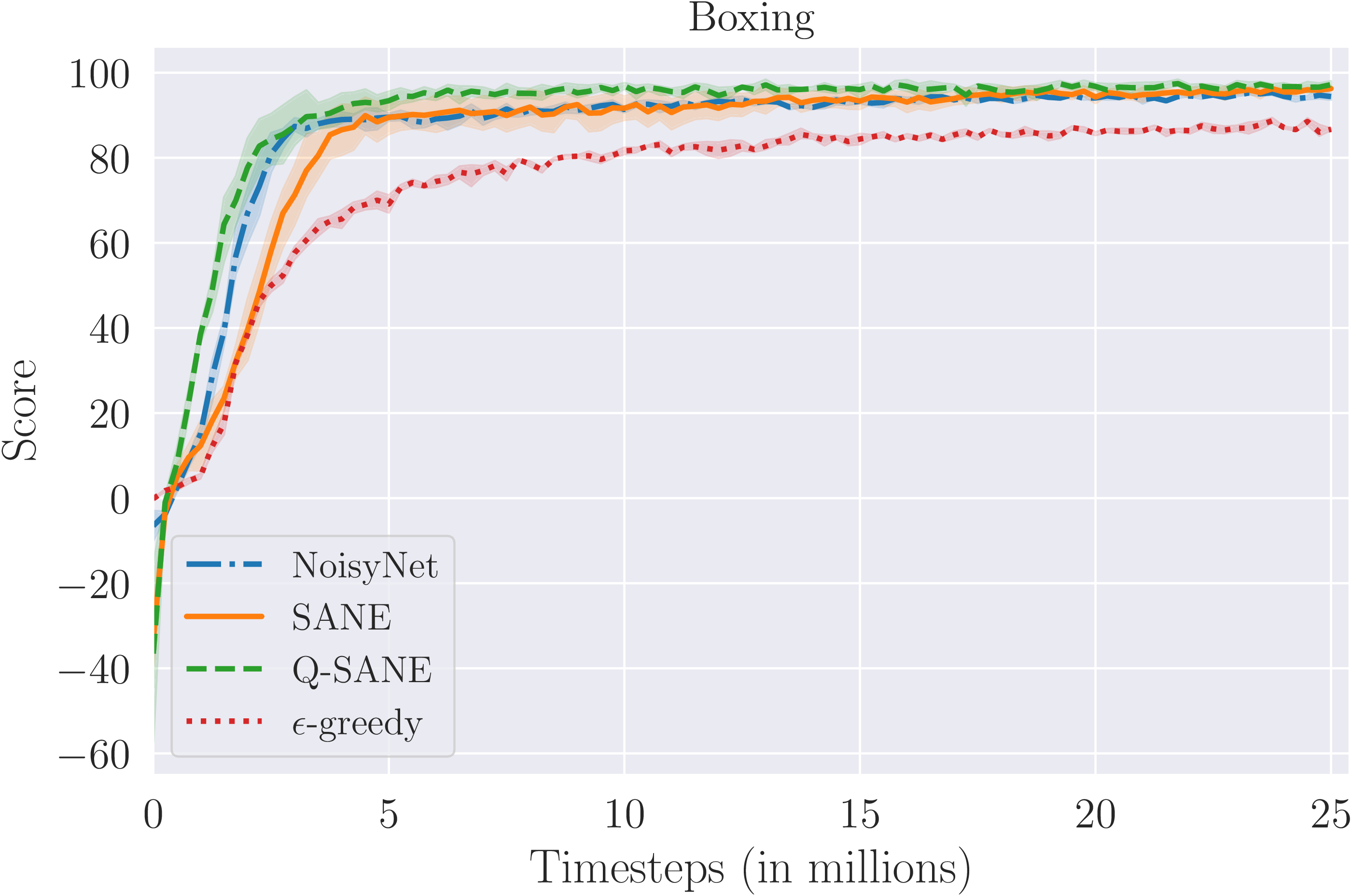}
\end{subfigure}
~
\begin{subfigure}[b]{0.23\textwidth}
\includegraphics[width=\linewidth,height=3cm]{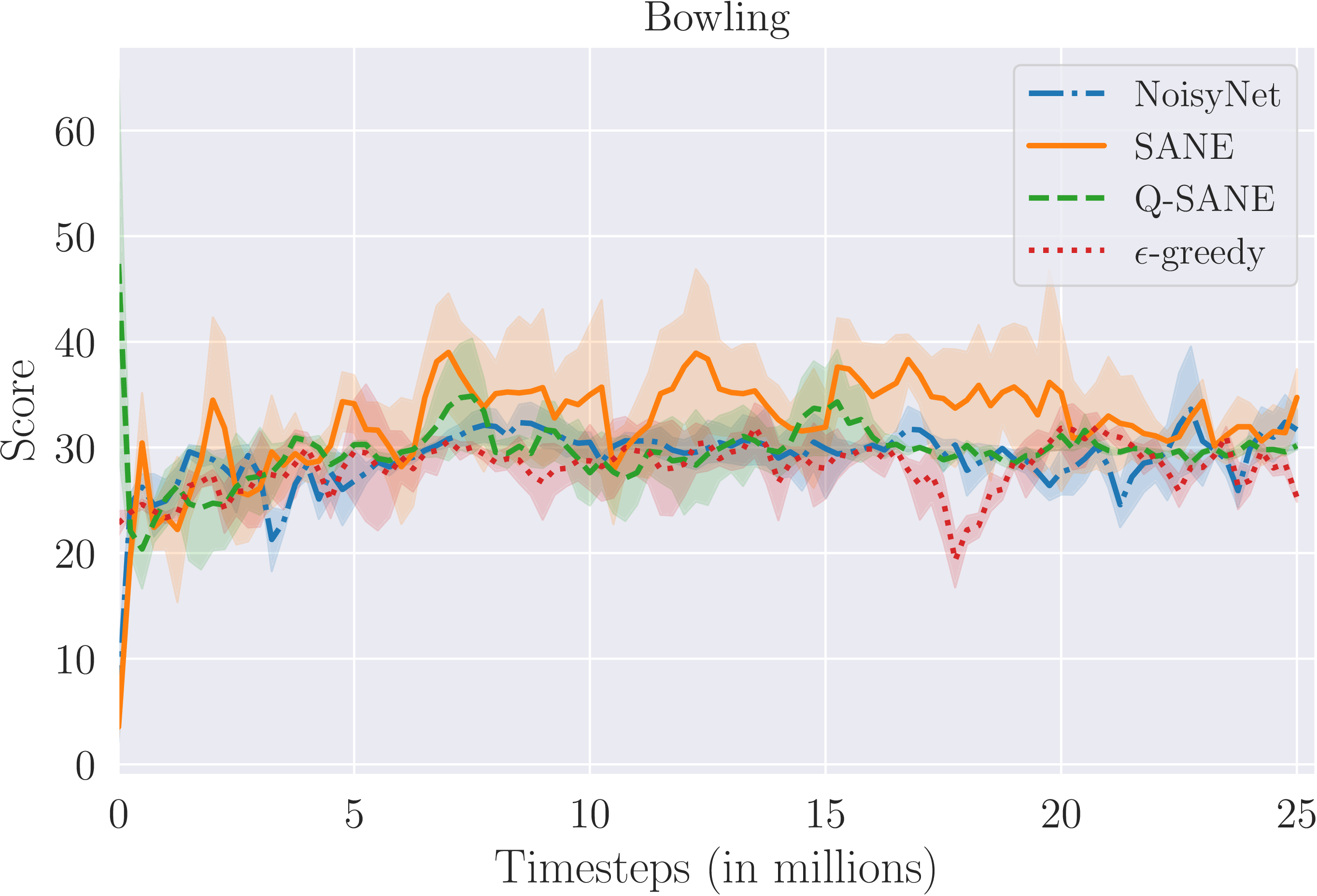}
\end{subfigure}

\caption{Learning curves of SANE DQNs, NoisyNet DQNs and $\epsilon$-greedy DQNs.}
\label{fig:dqn_exp}
\end{figure*} 

We follow the initialization scheme followed by \cite{fortunato2018noisy} to initialize the parameters of NoisyNet DQNs. Every layer of the main network of simple-SANE and Q-SANE DQNS are initialized with the Glorot uniform initialization scheme \cite{glorot2010understanding}, while every layer of the perturbation module of both the SANE DQN variants are initialized with samples from $\mathcal{N}(0,\frac{2}{l})$, where $l$ is number of inputs to the layer.  
\begin{table*}[t]
   \caption{Scores of DQN agents when evaluated without noise injection or exploratory action selection.}
    \label{tab:DQN_scores}

\centering

  \begin{tabular}{ccccc}
    \toprule
    Game & Q-SANE &  simple-SANE & NoisyNets & $\epsilon$-greedy  \\
    \midrule
    Asterix &126213$\pm$20478 &\textbf{133849 $\pm$ 49379} & 110566 $\pm$ 31800 &15777 $\pm$ 3370\\
    Atlantis & 141337 $\pm$ 67719 & \textbf{265144 $\pm$ 151154} &162738 $\pm$ 73271 & 229921 $\pm$ 41143 \\
    Enduro & 2409$\pm$ 321 &\textbf{2798 $\pm$ 311} & 2075 $\pm$ 24 & 1736$\pm$ 197 \\
    IceHockey& 2.99$\pm$1.4  &1.43 $\pm$ 1.76& -2.4 $\pm$ 0.52 &\textbf{3.46 $\pm$ 0.8} \\
    Qbert & \textbf{17358 $\pm$1015} & 15341 $\pm$ 162&15625 $\pm$ 166 &16025 $\pm$ 555 \\
    Riverraid & 14620$\pm$3491 & \textbf{14919 $\pm$ 997}& 11220 $\pm$ 223 &12023 $\pm$ 512\\
    RoadRunner & 49598$\pm$1635 & 45929 $\pm$ 1648 & \textbf{51805 $\pm$ 885} &47570 $\pm$ 1651 \\
    Seaquest & 8368$\pm$3426 & \textbf{8805 $\pm$ 1392} & 6031 $\pm$ 3567&7682 $\pm$ 1648 \\
    \midrule
    FishingDerby & -19.78 $\pm$7.2 & -12.1 $\pm$ 4.2 &\textbf{-11.5 $\pm$ 5.4} &-33.9 $\pm$ 9.1 \\
    Boxing & 95.3 $\pm$ 3 & 93.2 $\pm$ 4.5 & 95.5 $\pm$ 1.7&\textbf{96.6 $\pm$ 0.73} \\
    Bowling & 29$\pm$1 &28.08 $\pm$ 1.2 & \textbf{37.4 $\pm$ 3.8}&20.6 $\pm$ 4.7 \\
    \midrule
    \textbf{Score (8 games)}  & 4.86 & \textbf{5.51} & 4.28 
    & 3.33\\
        \textbf{Score (11 games)}  & 4.1 & \textbf{4.85} & 3.98 & 3.25 \\
    \bottomrule
  \end{tabular}
\end{table*}

\begin{table*}
    \caption{Scores of DQN Agents when evaluated with noise injection.}
    \label{tab:DQN_scores_perturbed}
\centering
  \begin{tabular}{cccc}
    \toprule
    Game & Q-SANE &  simple-SANE & NoisyNets \\
    \midrule
    Asterix & 182797$\pm$51182 &\textbf{194547 $\pm$ 56492} & 134682 $\pm$ 26574\\
    Atlantis & \textbf{281189$\pm$126834} & 230837 $\pm$ 104472 &166512 $\pm$ 93945  \\
    Enduro & 2849 $\pm$ 464 & \textbf{2855 $\pm$ 579} & 1946 $\pm$ 136 \\
    IceHockey& \textbf{2.86 $\pm$ 1.97} &1.9 $\pm$ 3.25& -1.53 $\pm$ 0.45  \\
    Qbert & \textbf{16950$\pm$479} & 15438 $\pm$ 57&13824 $\pm$ 2690  \\
    Riverraid & 15168$\pm$ 2068& \textbf{15434 $\pm$ 891}& 11076 $\pm$ 889 \\
    RoadRunner & 47434$\pm$ 2352& 47578 $\pm$ 3787 & \textbf{51260 $\pm$ 712}  \\
    Seaquest & 7184$\pm$ 2806& \textbf{7844 $\pm$ 1245} & 6087 $\pm$ 3654 \\
    \midrule
    FishingDerby & -15.92 $\pm$ 7.9&  \textbf{-10.83 $\pm$ 2.34} & -14 $\pm$ 2.8  \\
    Boxing & \textbf{96.16$\pm$ 1.73} & 95.1 $\pm$ 2.1 & 93.6 $\pm$ 2.7 \\
    Bowling & 28.8$\pm$ 0.61 &28.13 $\pm$ 1.25 & \textbf{34.2 $\pm$ 2.4}\\
    \midrule
    \textbf{Score (8 games)}  & \textbf{6.43} & 6.2 & 4.64\\
    \textbf{Score (11 games)}  & \textbf{5.54} &5.37 & 4.22 \\
    \bottomrule
  \end{tabular}
\end{table*}

\subsection{Architecture and Hyperparameters}
We use the same network structure to train NoisyNet, simple-SANE, Q-SANE and $\epsilon$-greedy DQNs. This network structure closely follows the architecture suggested in \cite{mnih2015human}. The inputs to all the networks are also pre-processed in a similar way.

The perturbations for NoisyNet, simple-SANE and Q-SANE DQNs are sampled using the Factored Gaussian noise setup \cite{fortunato2018noisy}. The SANE perturbation module used for all games and all SANE agents is a 1-hidden layer fully connected neural network. 
We train simple-SANE DQNs on the games of Asterix and Seaquest to determine the size of the  hidden layer. A hyperparameter search over the set $\{32, 64, 128, 256, 512\}$ revealed that a module with 256 hidden neurons gave the best results on these games. The hidden layer uses ReLU activation, and output layer computes one output which corresponds to the state aware standard deviation $\sigma(h(s;\theta^b);\Theta)$.

We train the DQNs with an Adam optimizer with learning rate, $\alpha = 6.25 \times 10^{-5}$. All other hyperparameters use the same values as used by \cite{mnih2015human}. The agents are trained for a total of 25M environment interactions where each training episode is limited to 100K agent-environment interactions.  Both NoisyNet and SANE DQNs use greedy action selection. Please refer to Sections B and C in the Appendix for more details about the implementation. 

For each game in the test suite, we train three simple-SANE, Q-SANE, NoisyNet and $\epsilon$-greedy DQN agents. 
Figure \ref{fig:dqn_exp} shows the average learning curves of all the learning agents. Each point in the learning curve corresponds to the average reward received by the agent in the last 100 episodes, averaged over 3 independent runs. Table \ref{tab:DQN_scores} shows the mean scores achieved by simple-SANE, Q-SANE, NoisyNet and $\epsilon$-greedy DQNs after being trained for 25M environment interactions on being evaluated for 500K frames with no noise injection. The scores of the best scoring agents in each game have been highlighted. We also evaluate simple-SANE, Q-SANE and NoisyNet DQNs with noise injection. These scores are presented in Table \ref{tab:DQN_scores_perturbed}. Tables \ref{tab:DQN_scores} and \ref{tab:DQN_scores_perturbed} also report the mean human-normalized scores (HNS) \cite{DBLP:journals/corr/abs-1905-12726} achieved by these methods on the 8 games which are likely to benefit from SANE exploration and on the whole 11 game suite. We also present some high-risk and low-risk states identified by Q-SANE agents in Figures \ref{fig:all_states_high_risk} and \ref{fig:all_states_low_risk}.

We observe that when evaluated without noise injection, both Q-SANE and simple-SANE outperform NoisyNets in 6 of the 8 games in the sub-suite.  NoisyNets achieve higher scores than both SANE variants in RoadRunner. 
In the three games not in the sub-suite, NoisyNets achieve higher but similar scores in FishingDerby and Boxing while performing much better in Bowling compared to the other agents. Evaluating the agents with noise injection proves beneficial for both SANE and NoisyNet agents, all of them achieving higher mean HNS in the 8 game sub-suite and the whole test suite. However, simple-SANE and Q-SANE agents achieve greater gains as they score higher than NoisyNets in 7 games in the sub-suite. SANE agents also score better in the remaining three games but do not manage to score better than NoisyNets in Bowling. Q-SANE and simple-SANE achieve the highest mean HNS on both the 8 game sub-suite and the whole test suite when evaluated with and without noise injection respectively.
\balance

\section{Conclusion}
\label{sec:conclusion}
In this paper, we derive a variational Thompson sampling approximation for DQNs, which uses the distribution over the network parameters as a posterior approximation. We interpret NoisyNet DQNs as an approximation to this variational Thompson Sampling method where the posterior is approximated using a state uniform Gaussian distribution. Using a more general posterior approximation, we propose State Aware Noisy Exploration, a novel exploration strategy that enables state dependent exploration in deep reinforcement learning algorithms by perturbing the model parameters. The perturbations are injected into the network with the help of an augmented SANE  module, which draws noise samples from a Gaussian distribution whose variance is conditioned on the current state of the agent. We hypothesize that such state aware perturbations are useful to direct exploration in tasks that go through a combination of high risk and low risk situations, an issue not considered by other methods that rely on noise injection. 

We test this hypothesis by evaluating two SANE DQN variants, namely simple-SANE and Q-SANE DQNs, on a suite of 11 Atari games containing a selection of games, most of which fit the above criterion and some that do not.
We observe that both simple-SANE and Q-SANE perform better than NoisyNet agents in most of the games in the suite, achieving better mean human-normalized scores. 

An additional benefit of SANE noise injection mechanism is its flexibility of design. SANE effects exploration via a separate perturbation module, the size or architecture of which is not tied to the model being perturbed and hence is flexible to user design and can be tailored to the task. As a consequence, this exploration method might scale better to larger network models. Hence, SANE presents a computationally inexpensive way to incorporate state information into exploration strategies and is a step towards more effective, efficient  and scalable exploration.

\section*{Acknowledgements}
This work was supported by the National Research Foundation Singapore
under its AI Singapore Program (Award Number: AISGRP-2018-006).

\bibliographystyle{ACM-Reference-Format}
\bibliography{citations}
\newpage
 \appendix
 \section{Deriving the ELBO}
In Section \ref{sec:ts}, we mentioned that minimizing the $KL(q(\theta), p(\theta|\mathcal{D})$ is equivalent to maximizing the ELBO. Here we provide a derivation of that claim.

\begin{align*}
KL(q(\theta), p(\theta|D)) = \int q(\theta) \log \frac{q(\theta)}{p(\theta|D)} d\theta
\end{align*}

Now, 
\begin{align*}
p(\theta|D) =& p(\theta|X,Y) = \frac{p(\theta)p(X,Y|\theta)}{p(X,Y)} = \frac{p(\theta)p(Y|X,\theta)p(X|\theta)}{p(X,Y)} \\
=& \frac{p(\theta)p(Y|X,\theta)p(X)}{p(X,Y)};  
\end{align*}

Substituting the above value, 
\begin{align*}
&KL(q(\theta), p(\theta|D)) = \int q(\theta) \left[ \log \frac{q(\theta)p(X,Y)}{p(\theta)p(Y|X,\theta)p(X)} \right] d\theta \\
&= \int q(\theta) \log \frac{q(\theta)}{p(\theta)p(Y|X,\theta)}d\theta + \int q(\theta) \log \frac{  P(X,Y)}{P(X)} d\theta \\
&= \int q(\theta) \log \frac{q(\theta)}{p(\theta)p(Y|X,\theta)} d\theta +  \log \frac{  P(X,Y)}{P(X)} \\
&= \int q(\theta) \log \frac{q(\theta)}{p(\theta)}d\theta - \int q(\theta) \log p(Y|X,\theta)d\theta +  \log \frac{  P(X,Y)}{P(X)} \\
&= KL( q(\theta), p(\theta)) - \int q(\theta) \log p(Y|X,\theta)d\theta +  \log \frac{  P(X,Y)}{P(X)}
\end{align*}

Now, $P(X,Y)$ and $P(X)$ are fixed with respect to $\theta$. So, minimizing $KL(q(\theta), p(\theta|D))$ is equivalent to maximizing the ELBO (Equation \ref{eq:KL1}).

 \section{The SANE-DQN Algorithm}
Algorithm \ref{alg:sae_dqn} describes the implementation of a SANE DQN. The Q network and the target network in SANE DQNs have their own copies of the network parameters. These are denoted by $\theta$ and $\hat{\theta}$ respectively. The Q and target networks also maintain different copies of the parameters of the perturbation module, $\Theta$ and $\hat{\Theta}$ respectively. Further, $\theta$ is partitioned into two sets, $\theta^b, \theta^p$, where $\theta^b$ is the set of \textit{base} parameters that help us compute the hidden state representation $h(s; \theta^b)$ and $\theta^p$ is the set of network parameters that are to be perturbed with state aware perturbations. We define similar counterparts $\hat{\theta^b}$ and $\hat{\theta^p}$ in the target network. 

With every forward pass, the network first calculates $h(s,\theta^b)$, which is then passed to the perturbation module. Factored Gaussian noise samples are procured and multiplied with $\sigma(h(s; \theta^b); \Theta)$, to get perturbations equivalent to those directly sampled from $\mathcal{N}(0,\sigma^2(h(s; \theta^b); \Theta))$ (Equation \ref{eq:reparameterization}). These perturbations are added to the parameters $\theta^p$. The agent then selects the action greedily with respect to the action values computed by the perturbed Q-network. While computing the batch-loss, the Q network and target network parameters $\theta^p$ and $\hat{\theta^p}$ are perturbed with state aware perturbations sampled from $\mathcal{N}(0,\sigma^2(h(s; \theta^b); \Theta))$ and $\mathcal{N}(0,\sigma^2(h(s; \hat{\theta^b}); \hat{\Theta}))$ respectively (Lines 21-25 in Algorithm \ref{alg:sae_dqn}). This loss is then backpropagated to train $\theta$ and $\Theta$.

\begin{algorithm}[ht]
\caption{SANE Deep Q Learning}
\label{alg:sae_dqn}
\begin{algorithmic}[1]
\State Initialize the target network parameters $\hat{\theta},\hat{\Theta}$ and the Q network  parameters $\theta, \Theta$ randomly. 
\State Initialize an empty experience replay buffer;
\State Initialize noise sampling method $N$.
\State steps $\gets$ 0
\While {steps $\leq$ max\_steps}
\State $t \gets 0$
\State Observe $s_0$
\While {$s_t$ not terminal}
\State Compute $h(s_t;\theta^b)$.
\State Sample perturbations $\epsilon \sim N(0,1)$ to perturb $\theta^p$ 
\State $\theta^{p'} \gets \theta^{p} +  \sigma(h(s_t;\theta^{b}); \Theta)\epsilon$ 
\State $a_t \gets \max\limits_{a} Q(s_t, a; \theta^{p'}, \theta^b) $
\State Take $a_t$, observe next state $s'$, reward $r_t$
\State Add transition $(s_t, a_t, r_t, s')$ to the replay buffer; 
\If {buffer\_size $\geq$ max\_buffer\_size}
\State Remove oldest buffer entry; 
\EndIf
\If {steps $\bmod$ update\_frequency = 0}
\State Sample $b$ transitions $\{(s_{i}, a_{i}, r_{i}, s'_{i}), 1 \leq i \leq b\}$
 uniformly from the replay buffer.
\State $L \gets 0$
\For {$i \in \{1 \cdots b\}$}
\State Compute  $h(s_t';\theta^b)$ and  $h(s_t';\hat{\theta^b})$
\State Sample $\epsilon, \hat{\epsilon} \sim N(0,1)$ to perturb $\theta^p, \hat{\theta^p}$ 
\State $\theta^{p'} \gets \theta^{p} +  \sigma(h(s_t;\theta^{b}); \Theta)\epsilon$ 
\State $\hat{\theta^{p'}} \gets \hat{\theta^{p}} +  \sigma(h(s_t;\hat{\theta^{b}}); \hat{\Theta})\hat{\epsilon}$
\State $T = (r_{i} + \gamma\max\limits_{a} Q(s'_{i}, a; \hat{\theta^{p'}},\hat{\theta^b}))$
\State $L= L+ \frac{1}{b}\left[Q(s_{i}, a_{i}; \theta^{p'},\theta^b) -  T \right]^{2}$
\EndFor
\State Backpropagate to minimize the batch loss  L
\EndIf

\If {steps $\bmod$ copy\_frequency = 0}
\State $\hat{\theta} \gets \theta$ ; $\hat{\Theta} \gets \Theta$ 
\EndIf
\State $t \gets t + 1$; steps $\gets$ steps+1
\EndWhile
\EndWhile
\end{algorithmic}
\end{algorithm}

\section{DQN Implementation Details}
The network structure and input pre-processing closely follows the architecture and method suggested in \cite{mnih2015human}. The DQN consists of three convolutional layers followed by two linear layers. The first convolutional layer has 32 filters of size $8 \times 8$. This layer is followed by a convolutional layer with 64 filters of size $4 \times  4$. The last convolutional layer has 64 filters of size $3  \times 3$. The convolutional layers use strides of 4,2 and 1 respectively. The convolutional layers are followed by 2 fully connected layers, a hidden layer with 512 neurons and an output layer with the number of outputs being equal to the number of actions available to the agent for the task. With the exception of the output layer, a ReLU activation function follows every layer.

The perturbations for both the networks are sampled using the Factored Gaussian noise setup \cite{fortunato2018noisy}. The state aware perturbation module used for all games is a 1-hidden layer fully connected neural network. The hidden layer consists of 256 neurons and uses ReLU activation. The output layer computes one output which corresponds to the state aware standard deviation $\sigma(h(s;\theta^b);\Theta)$.

We train the DQNs with an Adam optimizer with learning rate, $\alpha = 6.25 \times 10^{-5}$ and $\epsilon = 1.5 \times 10^{-4}$. We use a replay buffer that can hold a maximum of 1M transitions. We populate the replay buffer with 50K transitions that we obtain by performing random actions for $\epsilon$-greedy agents and by following the policy suggested by the network for NoisyNet and SANE agents. Thereafter, we train the Q network once every 4 actions, with a batch of 32 transitions sampled uniformly from the replay buffer. We copy over the parameters of the Q network to the target network after every 10K transitions. We use a discount factor of $\gamma=0.99$ for all games. Additionally, the rewards received by the agent are clipped in the range $[-1,1]$. 

The agent is trained for a total of 25M agent-environment interactions. The input to the network is a concatenation of 4 consecutive frames, and we take a random number of no-op actions (upto 30) at the start of each episode, so that the agent is given a random start.

The codebase for our SANE, Q-SANE, NoisyNet and $\epsilon$-greedy DQN agents are available at \cite{gitSANE}.

\section{Additional Experimental Details}

\subsection{Human Normalized Scores}
The human normalized score for any agent is calculated as follows
\begin{equation}
    \text{HNS} = \frac{\text{Score}_\text{agent} - \text{Score}_\text{random}} {\text{Score}_\text{human} - \text{Score}_\text{random}}
\end{equation}
We use the same random and human baseline scores as used in \cite{DBLP:journals/corr/abs-1905-12726}. We list these baseline scores in Table \ref{tab:baseline_scores} for easy access. 
\begin{table}[h]
\centering
    \caption{Baseline human and random values used to calculate Human Normalized Scores}
  \begin{tabular}{ccc}
    \toprule
    Game & Human Score& Random Score\\
    \midrule
    Asterix & 8503&210 \\
    Atlantis & 29028&12850 \\
    Boxing & 12.1&0.1 \\
    Bowling & 160.7&23.1 \\
    Enduro & 860.5& 0\\
    FishingDerby & -38.7& -91.7\\
    IceHockey & 0.9& -11.2\\
    Qbert & 13455   &163.9\\
    Riverraid & 17118&1338.5 \\
    RoadRunner & 7845&11.5 \\
    Seaquest & 42054&68.4 \\
    \bottomrule
  \end{tabular}
    \label{tab:baseline_scores}
\end{table}
\subsection{Visualizations}

\begin{figure*}[t!]
\begin{subfigure}[b]{0.15\textwidth}
\includegraphics[width=\linewidth]{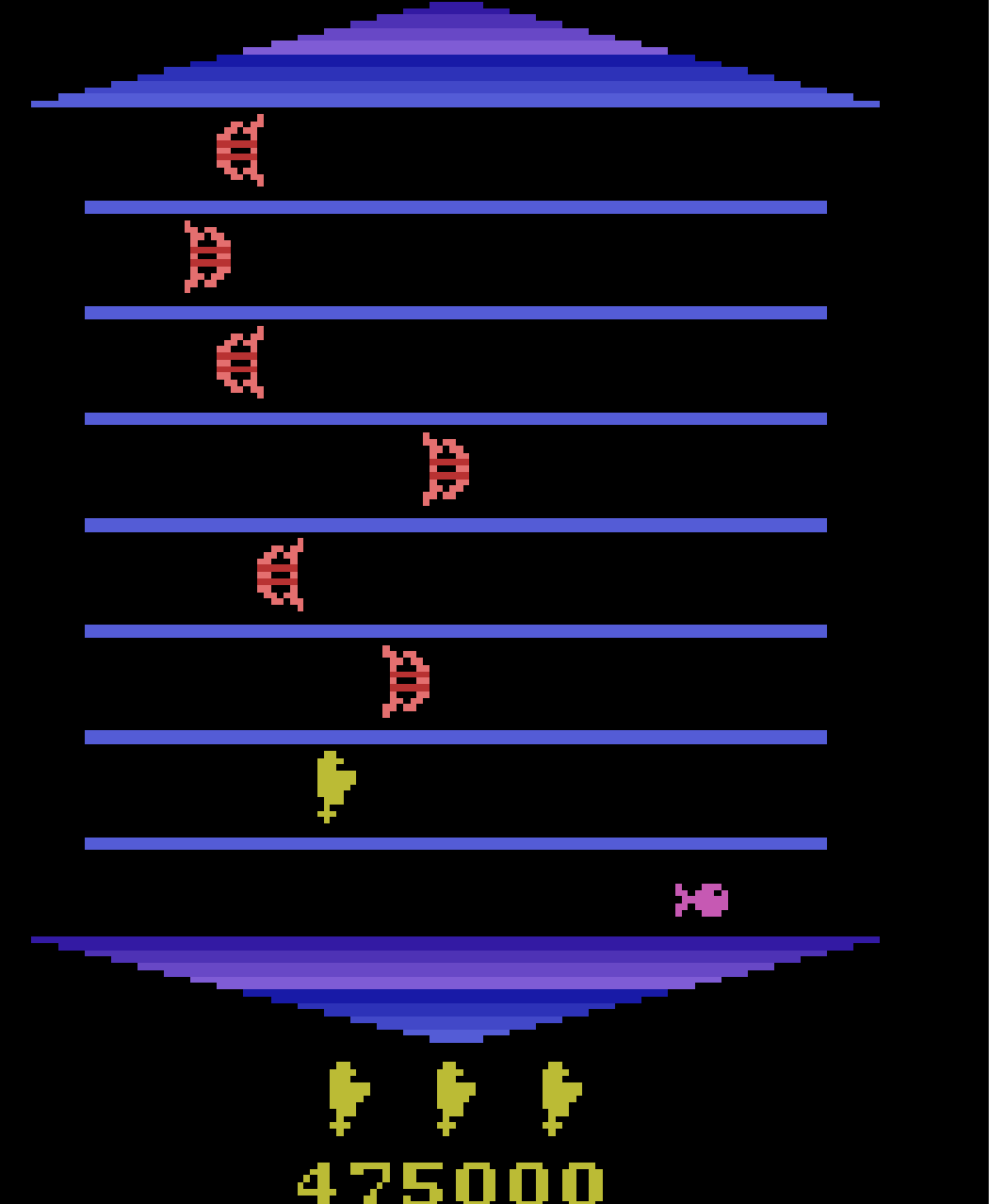}
\caption{$\sigma=3.62\times 10^{-8}$}
\label{fig:asterix_h_app}
\end{subfigure}
~
\begin{subfigure}[b]{0.15\textwidth}
\includegraphics[width=\linewidth]{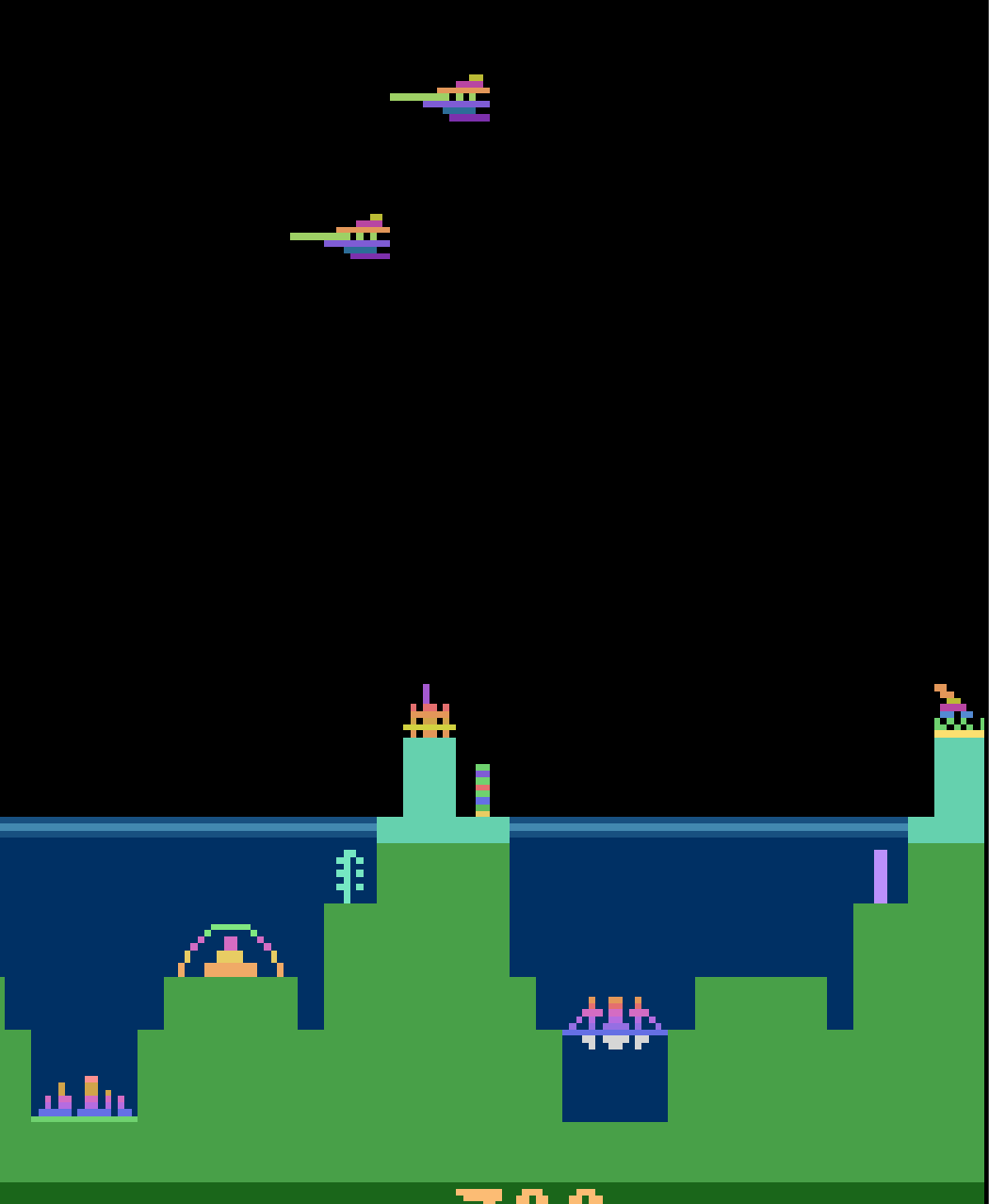}
\caption{$\sigma=7.84\times 10^{-7}$}
\label{fig:atlantis_h_app}
\end{subfigure}
~
\begin{subfigure}[b]{0.15\textwidth}
\includegraphics[width=\linewidth]{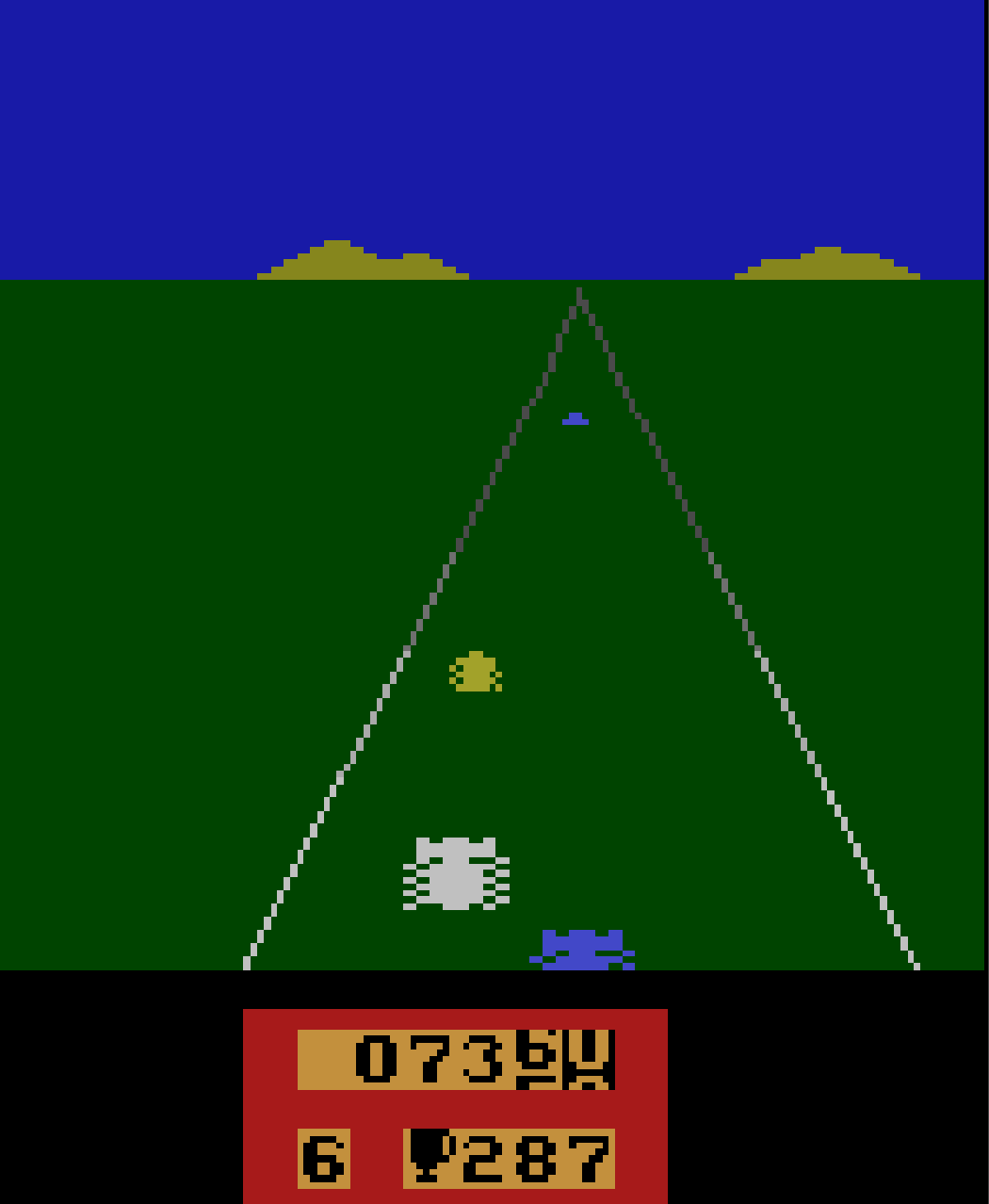}
\caption{$\sigma=2.2\times 10^{-7}$}
\label{fig:enduro_h_app}
\end{subfigure}
~
\begin{subfigure}[b]{0.15\textwidth}
\includegraphics[width=\linewidth]{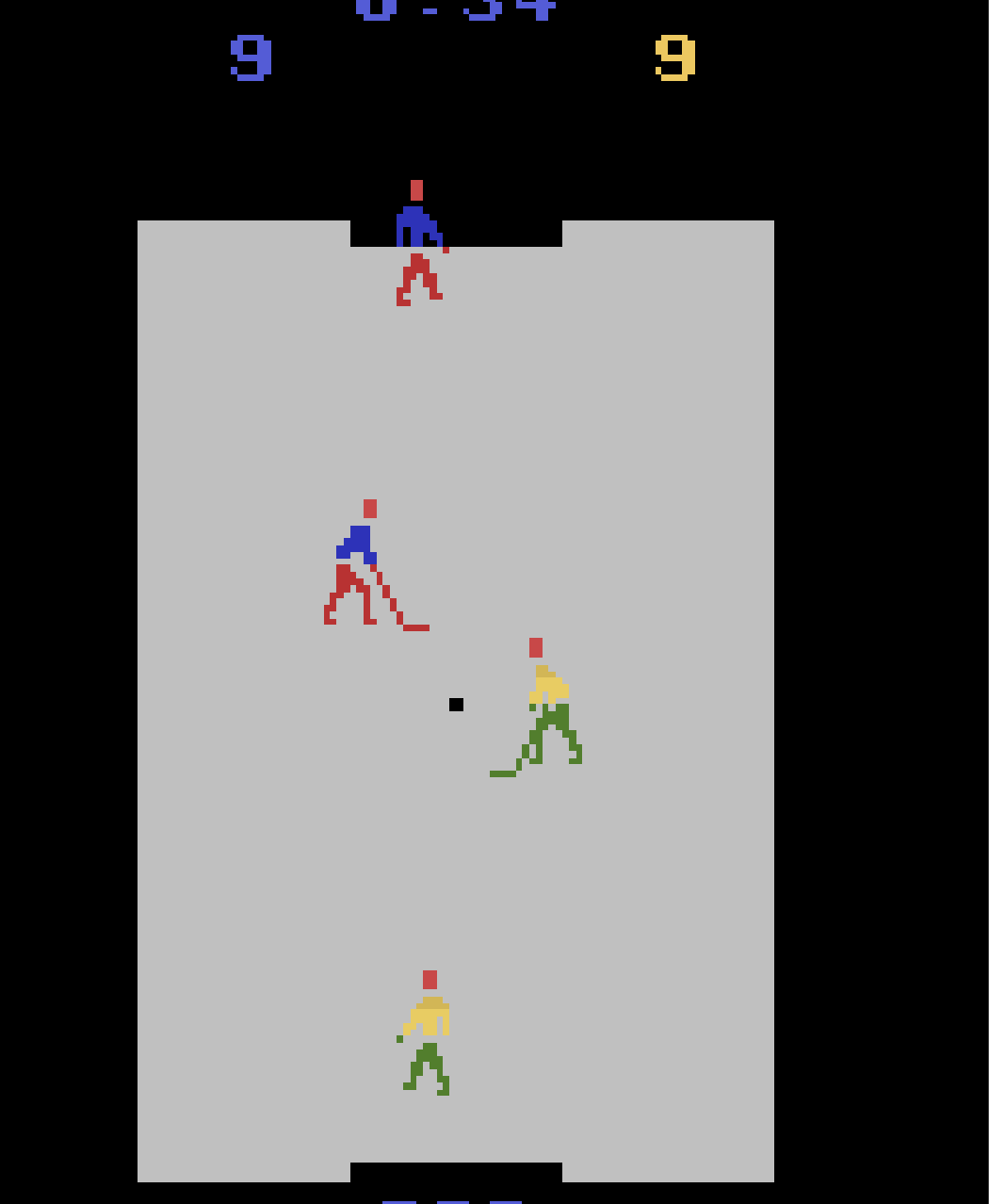}
\caption{$\sigma=3.62\times 10^{-6}$}
\label{fig:icehockey_h_app}
\end{subfigure}
 
\begin{subfigure}[b]{0.15\textwidth}
\includegraphics[width=\linewidth]{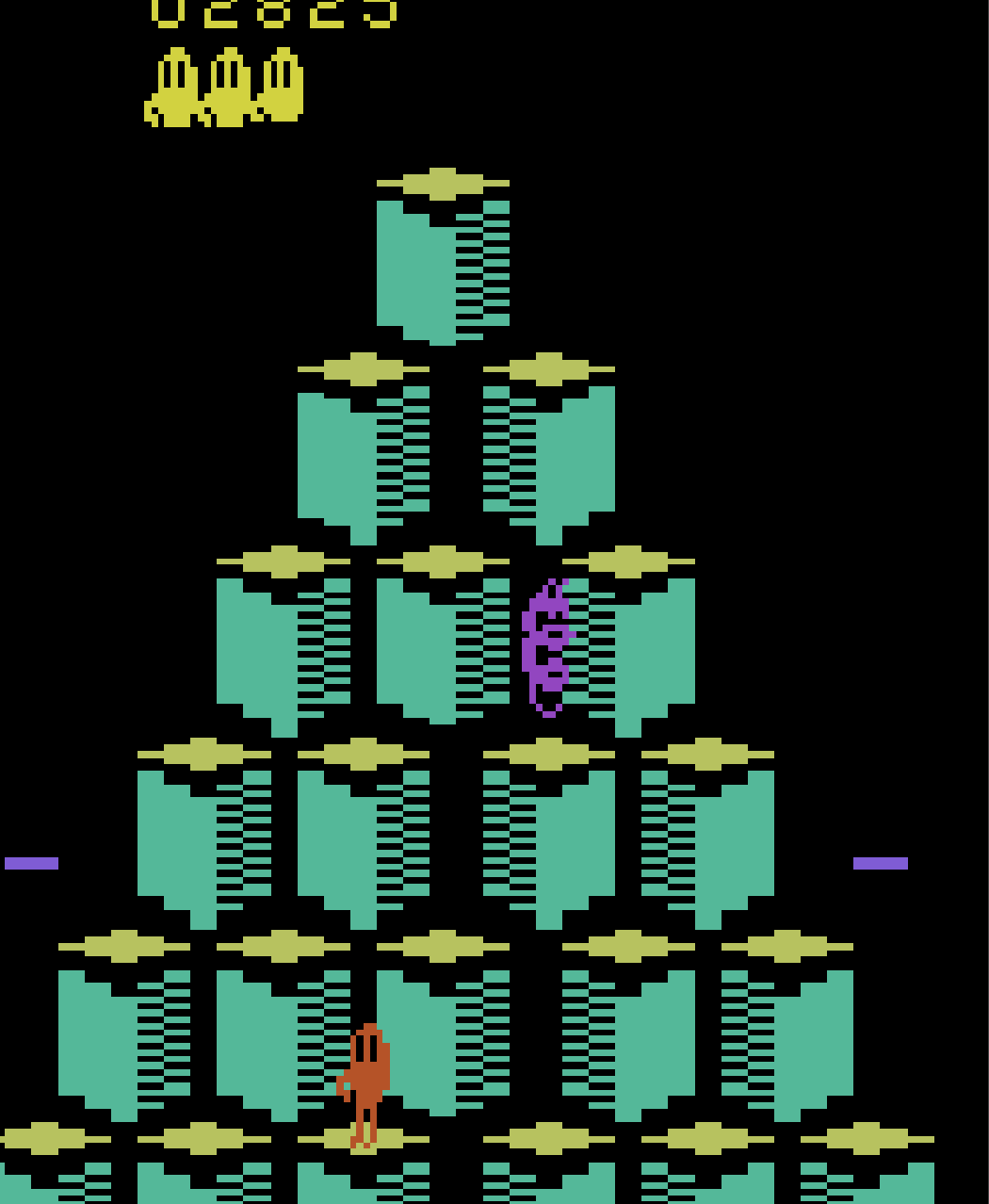}
\caption{$\sigma=8.55\times 10^{-6}$}
\label{fig:qbert_h_app}
\end{subfigure}
~
\begin{subfigure}[b]{0.15\textwidth}
\includegraphics[width=\linewidth]{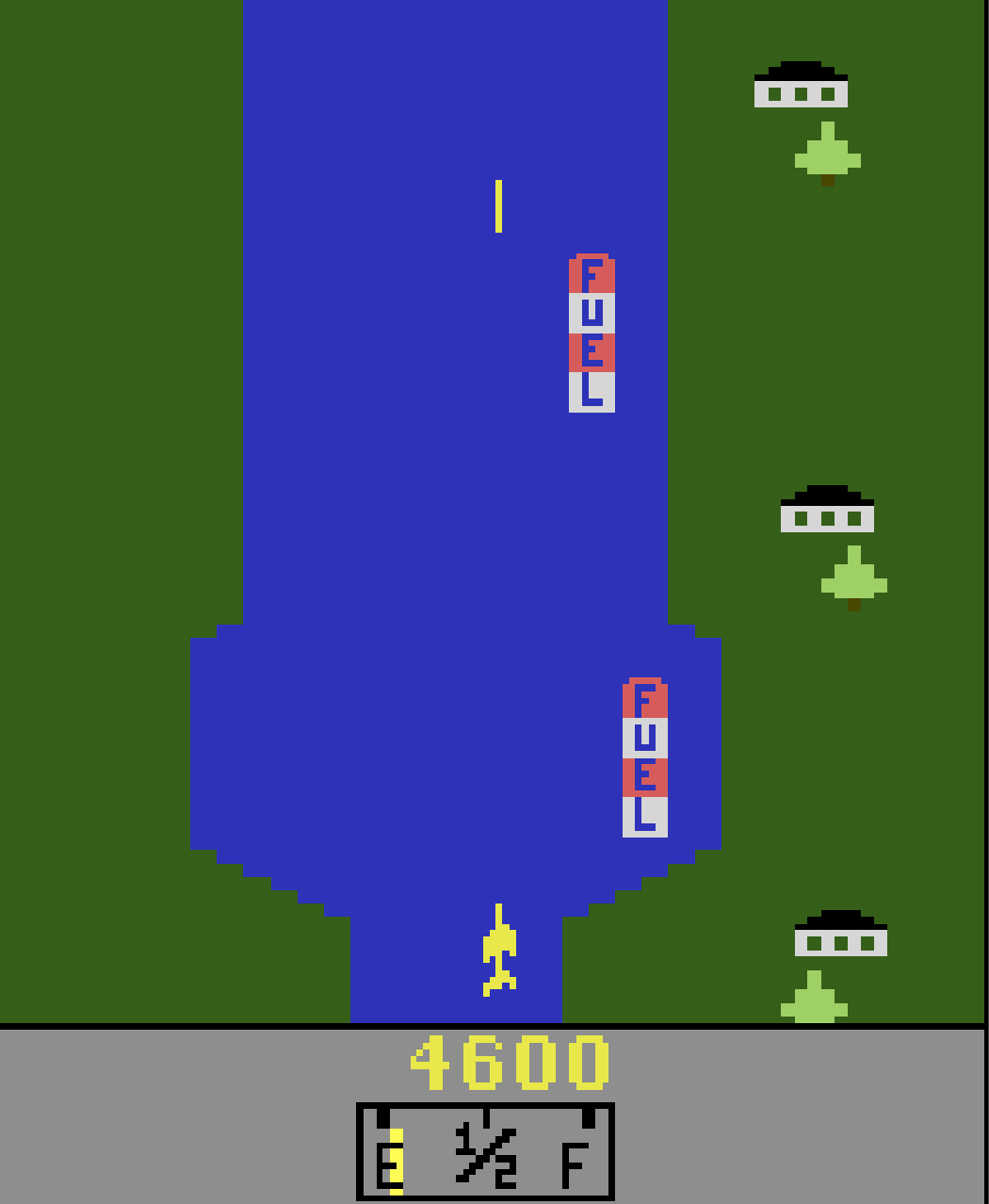}
\caption{$\sigma=1.52\times 10^{-6}$}
\label{fig:riverraid_h_app}
\end{subfigure}
~
\begin{subfigure}[b]{0.15\textwidth}
\includegraphics[width=\linewidth]{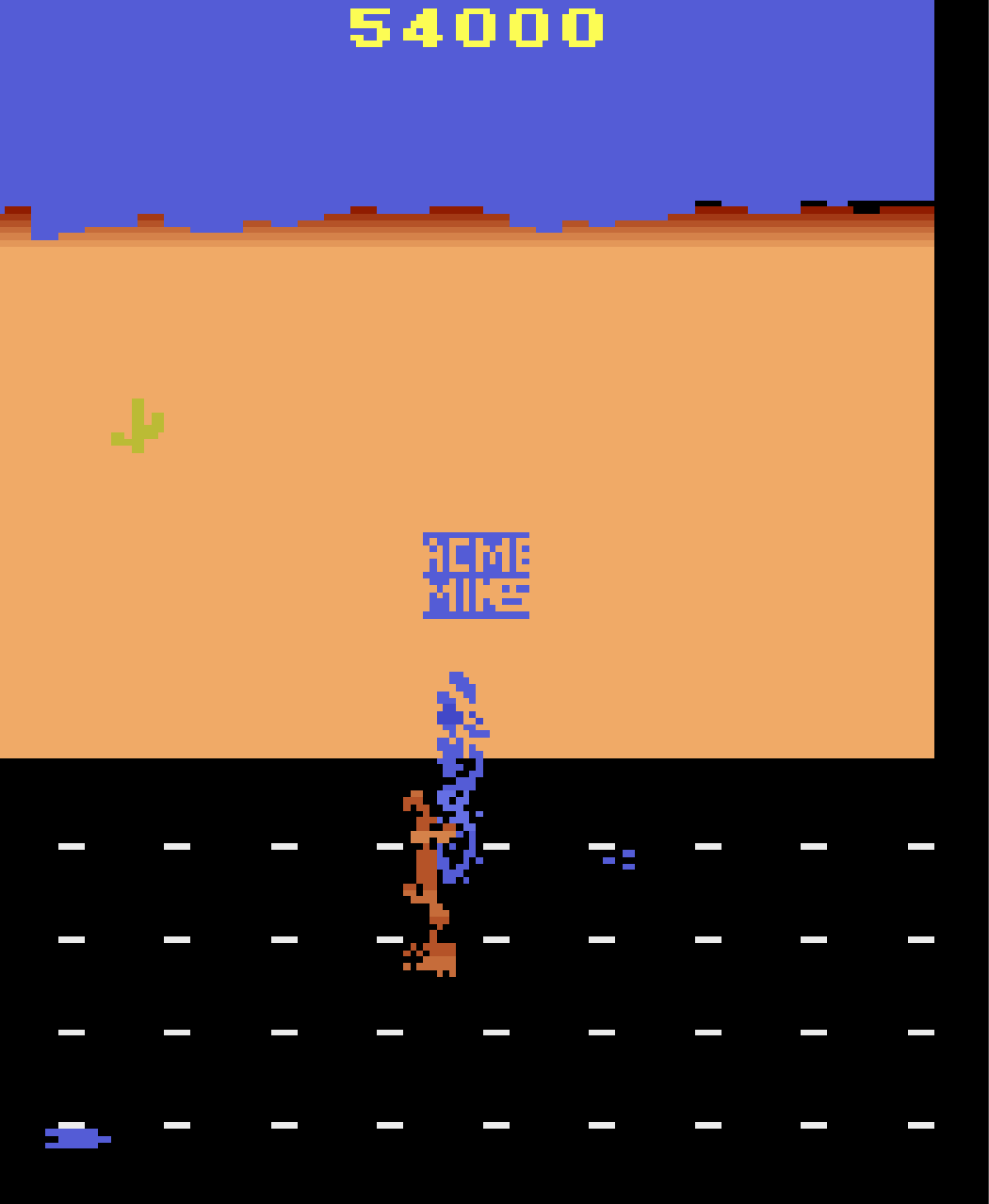}
\caption{$\sigma=1.03\times 10^{-6}$}
\label{fig:roadrunner_h_app}
\end{subfigure}
~
\begin{subfigure}[b]{0.15\textwidth}
\includegraphics[width=\linewidth]{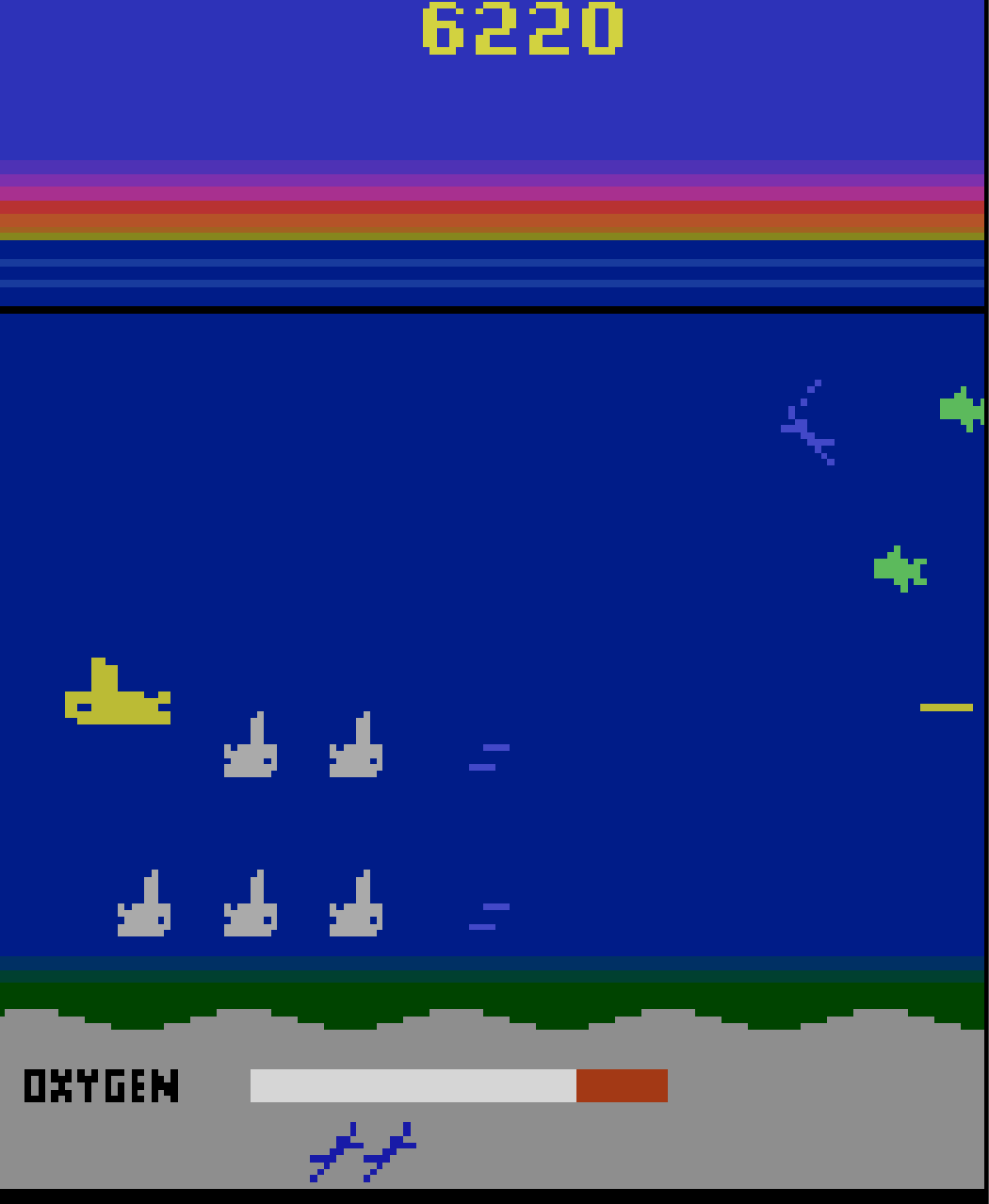}
\caption{$\sigma=1.13\times 10^{-7}$}
\label{fig:seaquest_h_app}
\end{subfigure}
\caption{High risk states learnt by Q-SANE in the 8 game sub-suite. The captions mention the standard deviation predicted for each state.}
\label{fig:all_states_high_risk_app}
\end{figure*}

\begin{figure*}[t!]
\begin{subfigure}[b]{0.15\textwidth}
\includegraphics[width=\linewidth]{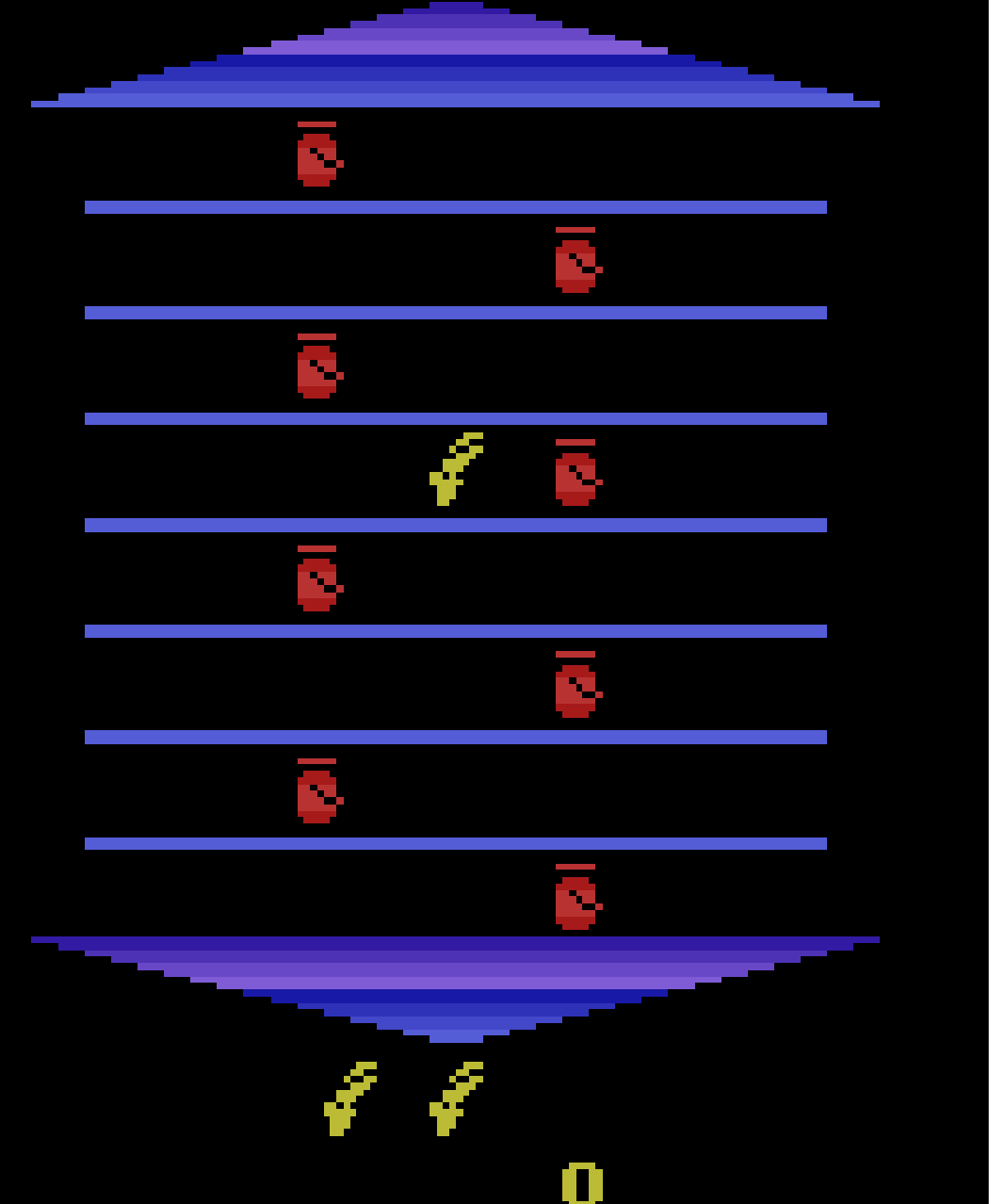}
\caption{$\sigma=3.18\times 10^{-4}$}
\label{fig:asterix_l_app}
\end{subfigure}
~
\begin{subfigure}[b]{0.15\textwidth}
\includegraphics[width=\linewidth]{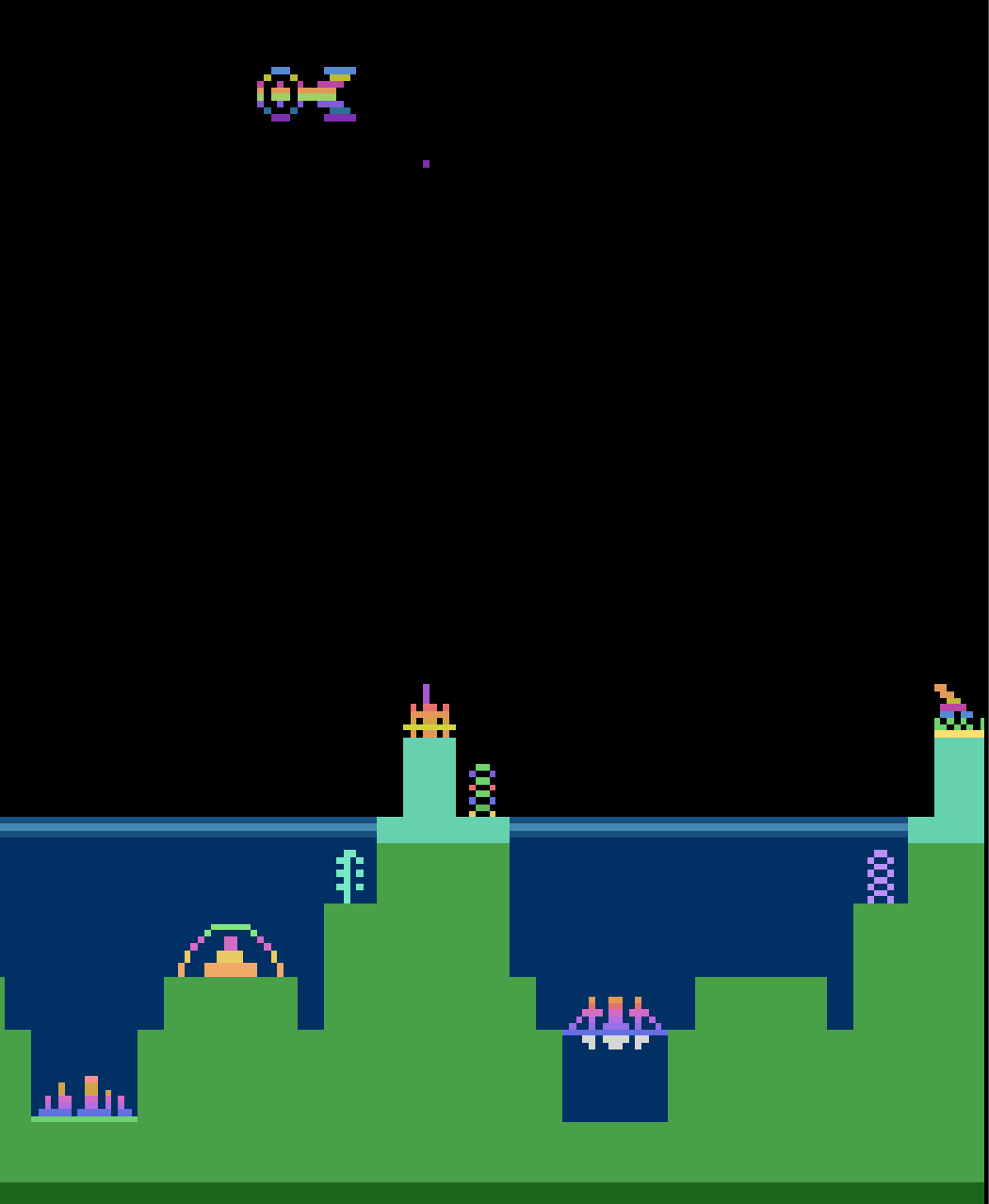}
\caption{$\sigma=3.29\times 10^{-4}$}
\label{fig:atlantis_l_app}

\end{subfigure}
~
\begin{subfigure}[b]{0.15\textwidth}
\includegraphics[width=\linewidth]{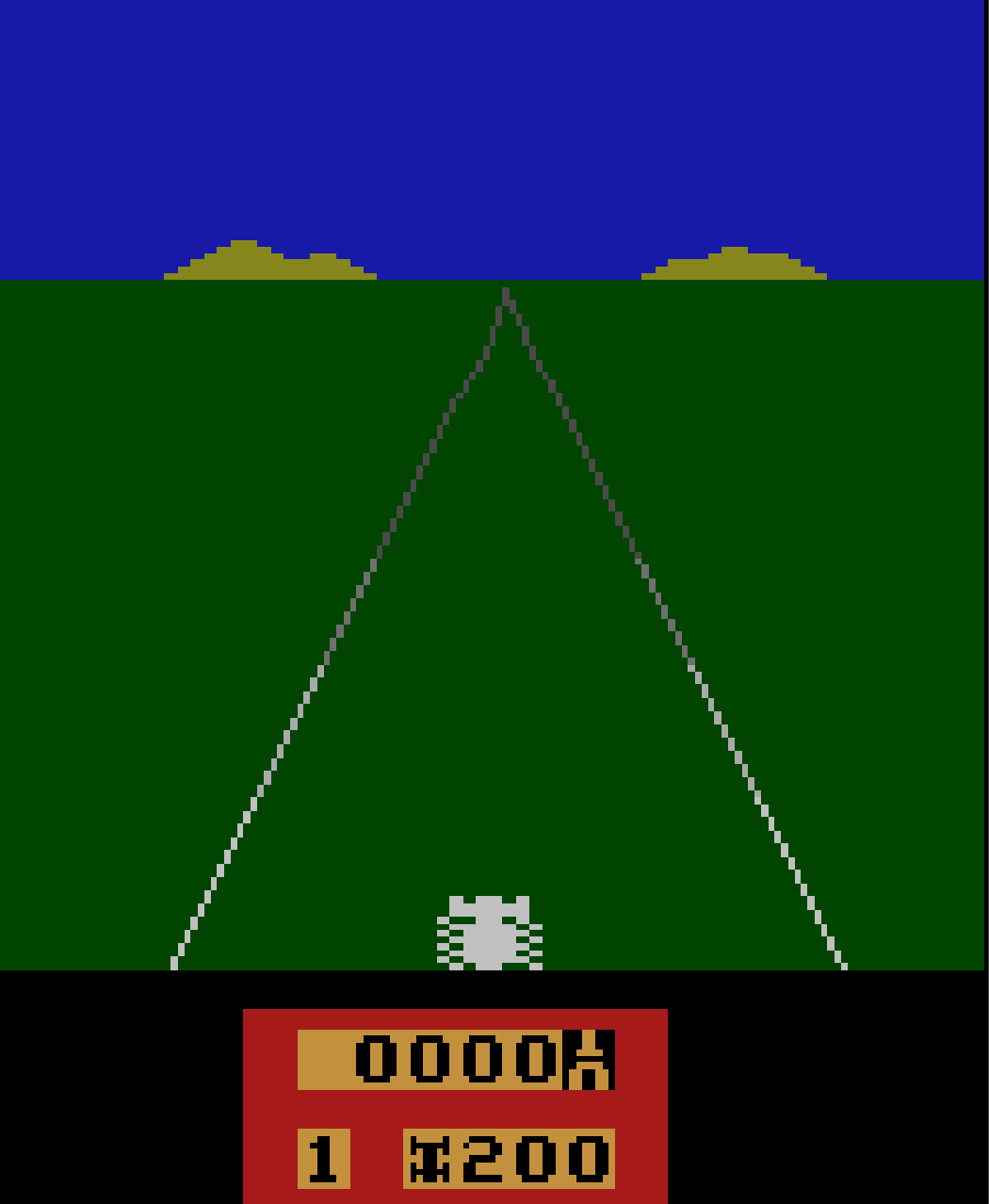}
\caption{$\sigma=2.6\times 10^{-4}$}
\label{fig:enduro_l_app}

\end{subfigure}
~
\begin{subfigure}[b]{0.15\textwidth}
\includegraphics[width=\linewidth]{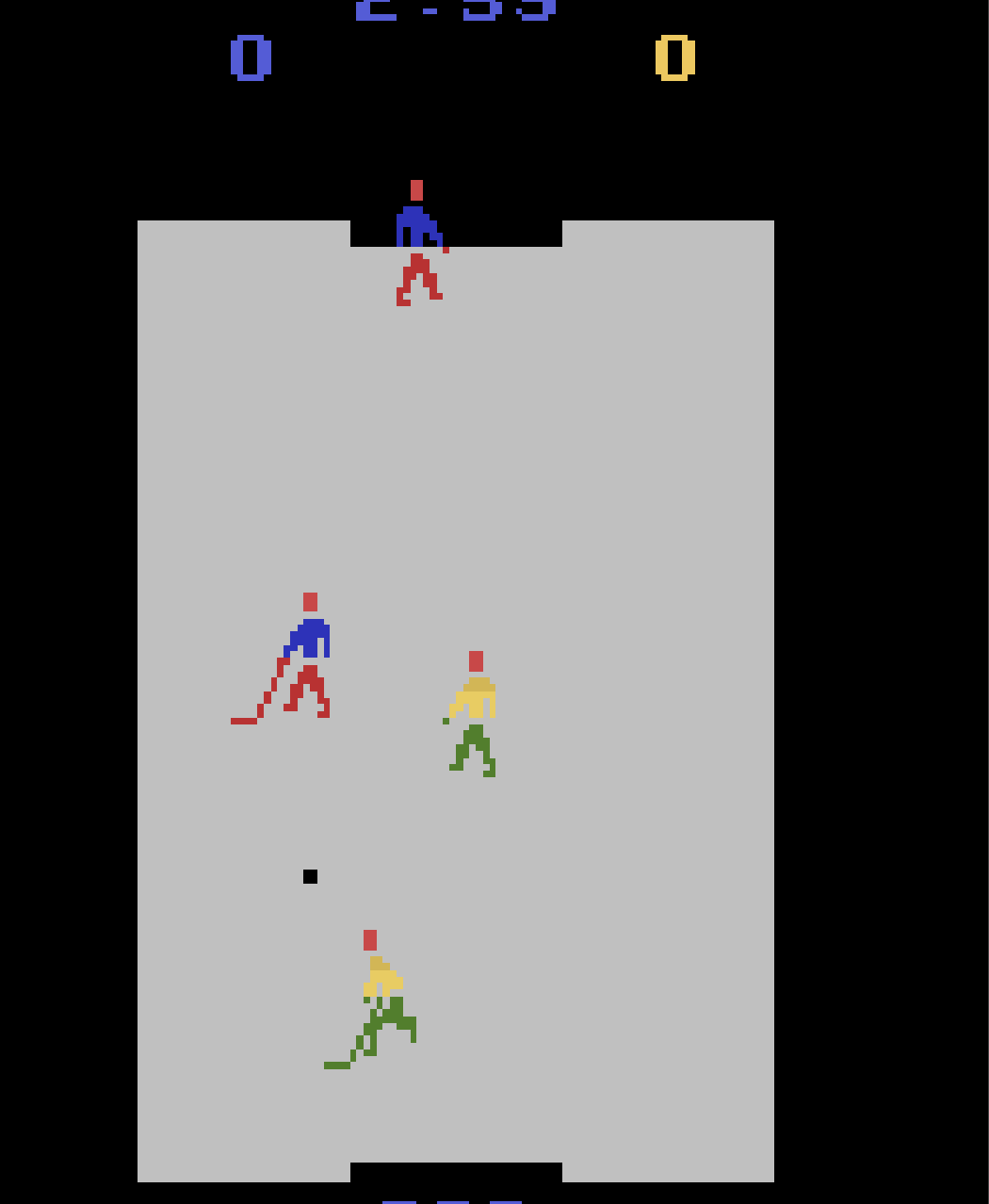}
\caption{$\sigma=7.97\times 10^{-4}$}
\label{fig:icehockey_l_app}

\end{subfigure}
 
\begin{subfigure}[b]{0.15\textwidth}
\includegraphics[width=\linewidth]{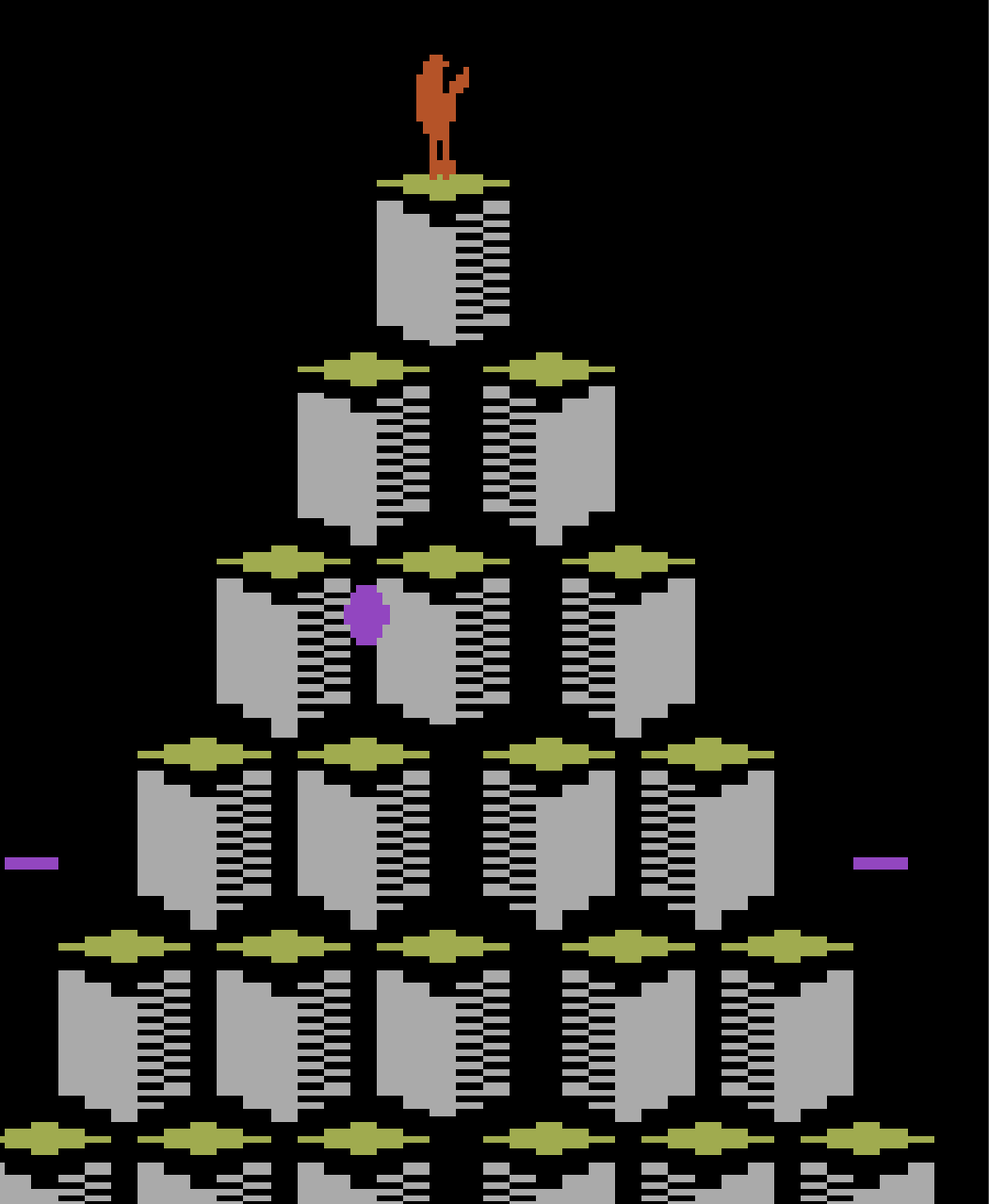}
\caption{$\sigma=4.07\times 10^{-4}$}
\label{fig:qbert_l_app}

\end{subfigure}
~
\begin{subfigure}[b]{0.15\textwidth}
\includegraphics[width=\linewidth]{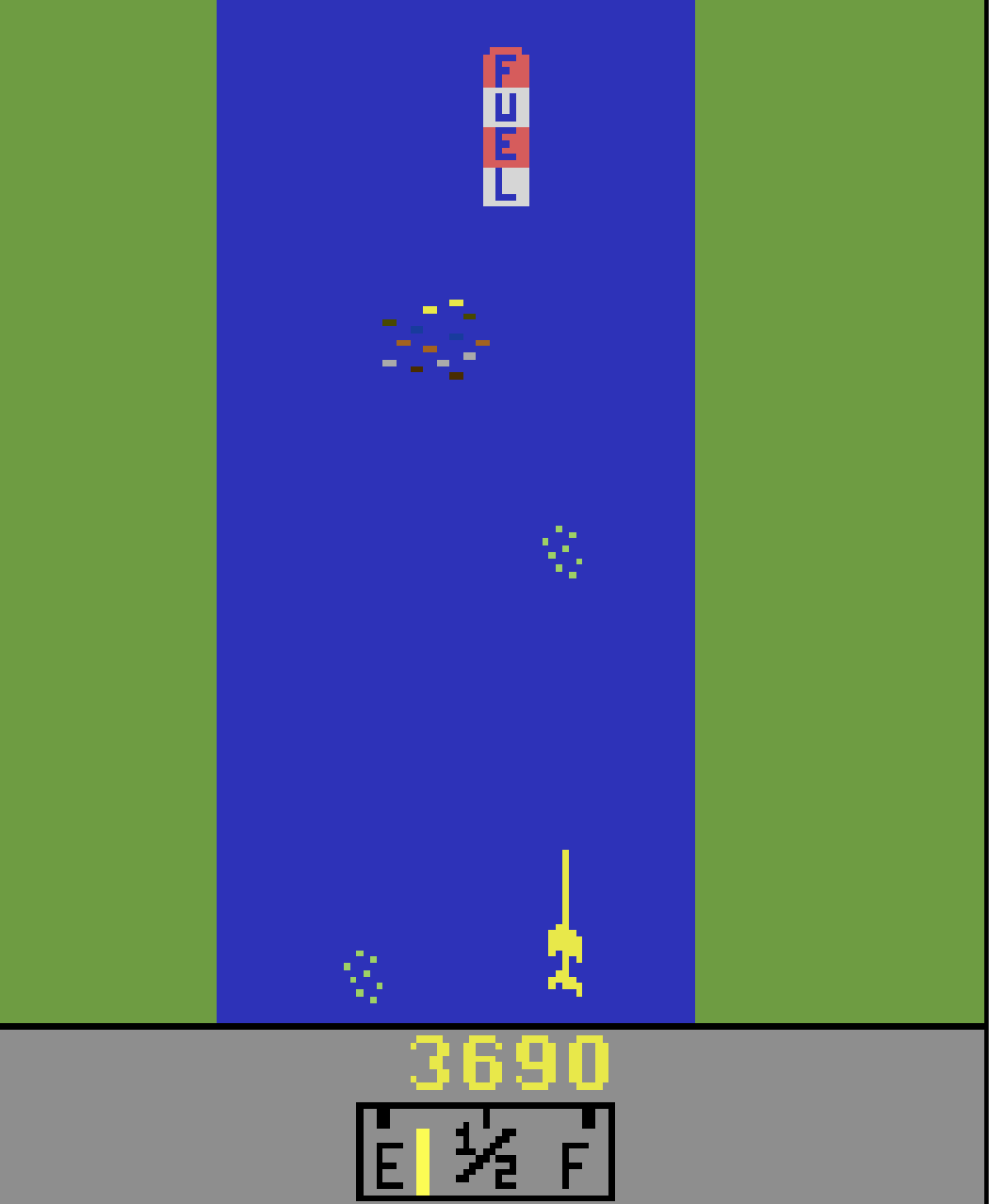}
\caption{$\sigma=3.62\times 10^{-4}$}
\label{fig:riverraid_l_app}

\end{subfigure}
~
\begin{subfigure}[b]{0.15\textwidth}
\includegraphics[width=\linewidth]{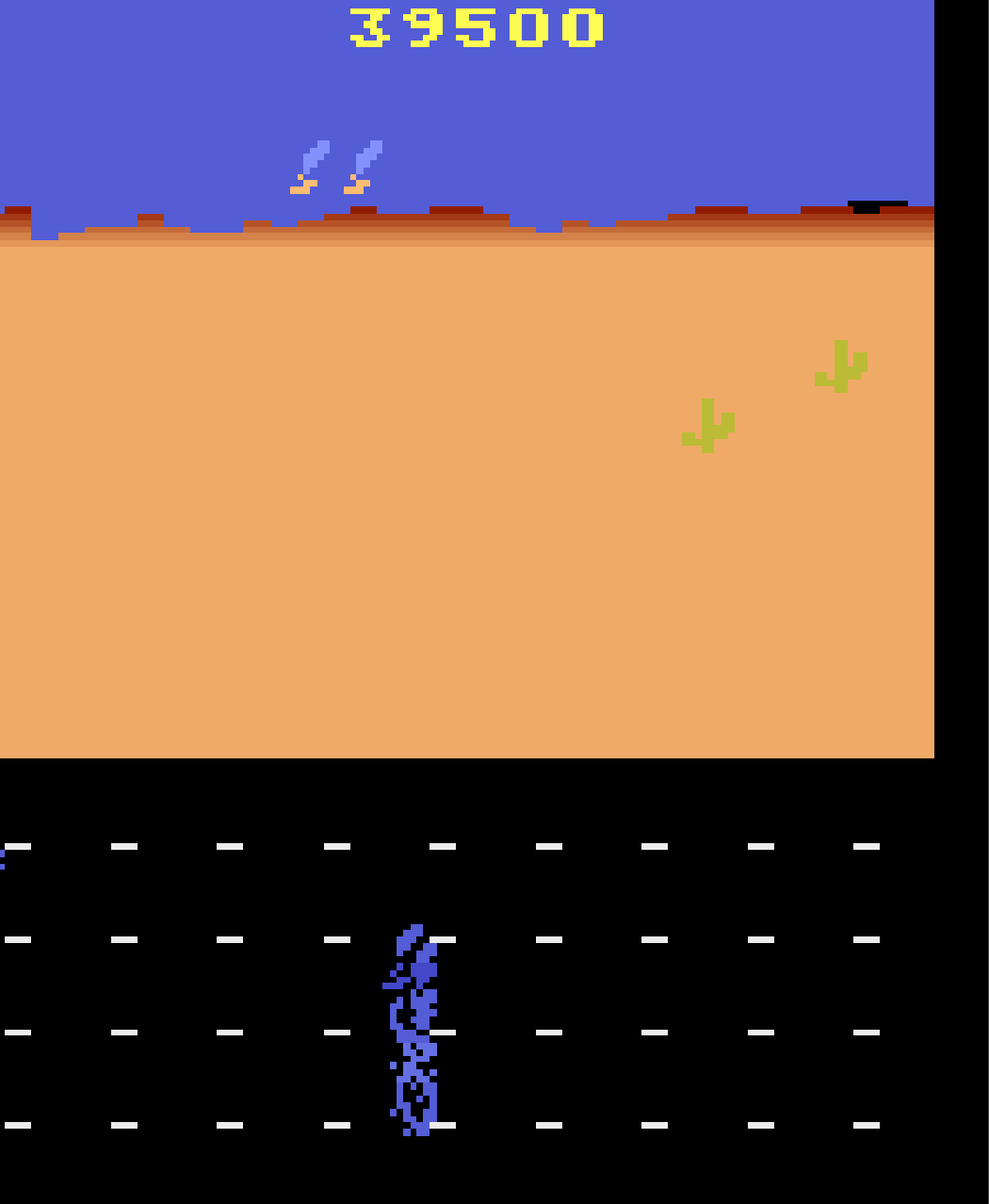}
\caption{$\sigma=7.82\times 10^{-5}$}
\label{fig:roadrunner_l_app}

\end{subfigure}
~
\begin{subfigure}[b]{0.15\textwidth}
\includegraphics[width=\linewidth]{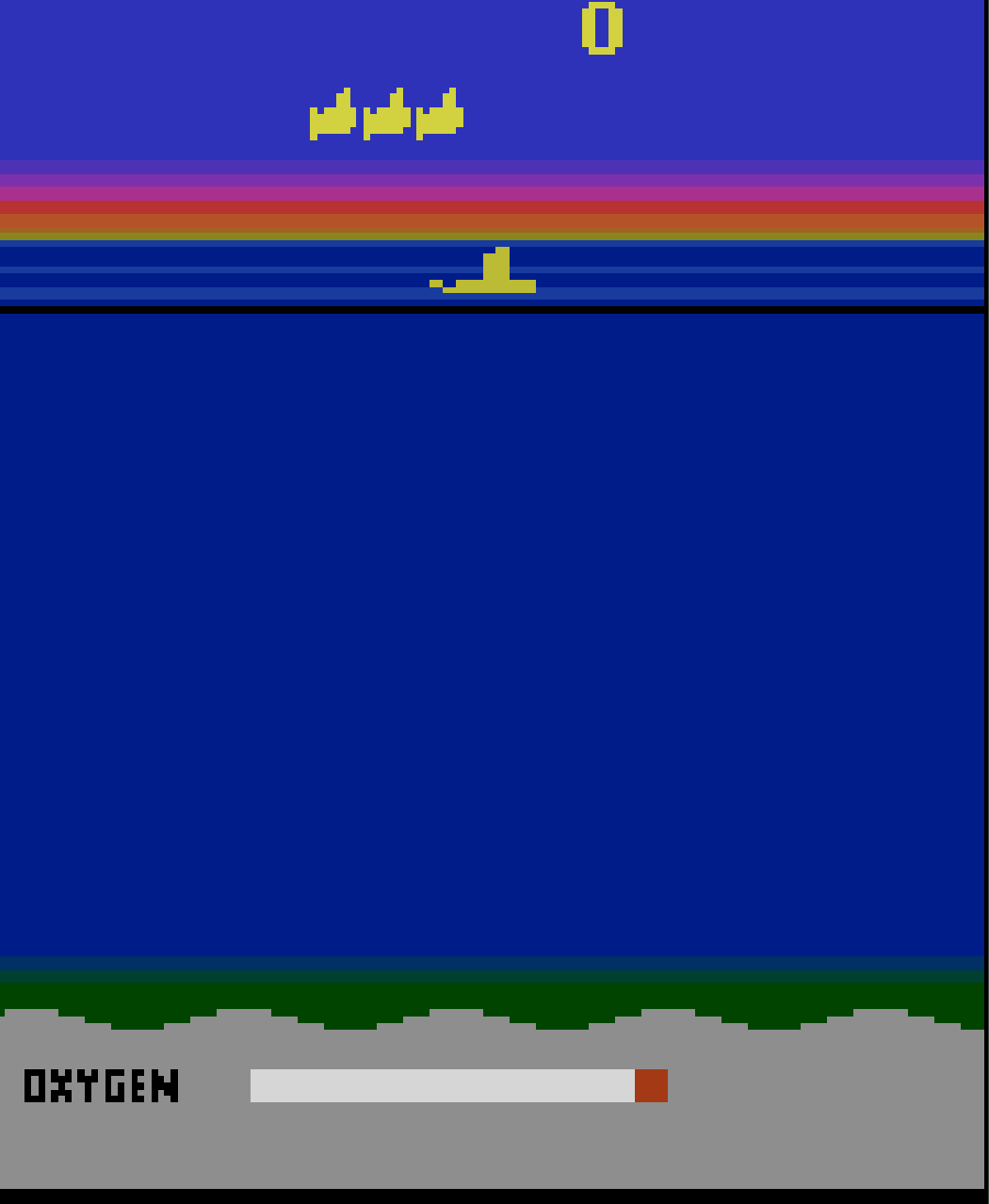}
\caption{$\sigma=1.3\times 10^{-4}$}
\label{fig:seaquest_l_app}

\end{subfigure}
 
\caption{Low risk states learnt by Q-SANE in the 8 game sub-suite. The captions mention the standard deviation predicted for each state.}
\label{fig:all_states_low_risk_app}
\end{figure*}

We present the high risk and low risk states (from Figures \ref{fig:all_states_high_risk} and \ref{fig:all_states_low_risk}) along with the state aware standard deviation predicted by the Q-SANE agents in Figures \ref{fig:all_states_high_risk_app} and \ref{fig:all_states_low_risk_app} respectively.

\end{document}